\documentclass{article}

\PassOptionsToPackage{numbers, compress}{natbib}

\usepackage[preprint]{neurips_2023}

\usepackage[utf8]{inputenc} 
\usepackage[T1]{fontenc}    
\usepackage[hidelinks]{hyperref}       
\usepackage{url}            
\usepackage{booktabs}       
\usepackage{amsfonts}       
\usepackage{nicefrac}       
\usepackage{xcolor}         
\usepackage{graphicx}
\usepackage{multirow}
\usepackage{tikz}
\usepackage[edges]{forest}
\usepackage{natbib}
\definecolor{hidden-draw}{RGB}{20,68,106}
\definecolor{hidden-pink}{RGB}{255,245,247}
\usepackage{amsmath}
\usepackage{setspace}
\usepackage{cleveref} 
\usepackage{makecell}
\usepackage{colortbl}

\title{Efficient Multimodal Large Language Models:\\ A Survey}

\author{%
      Yizhang Jin\textsuperscript{1,2,*},
      Jian Li\textsuperscript{1,*},
      Yexin Liu\textsuperscript{3},
      Tianjun Gu\textsuperscript{4},
      Kai Wu\textsuperscript{1},
      Zhengkai Jiang\textsuperscript{1},
  \\ 
  \textbf{
      Muyang He\textsuperscript{3}, 
      Bo Zhao\textsuperscript{3},
      Xin Tan\textsuperscript{4},
      Zhenye Gan\textsuperscript{1},
      Yabiao Wang\textsuperscript{1},
      Chengjie Wang\textsuperscript{1},
  }
  \\
  \textbf{
      Lizhuang Ma\textsuperscript{2}
  } 
  \\
  \\
  \textsuperscript{1}Youtu Lab, Tencent,
  \textsuperscript{2}SJTU, 
  \textsuperscript{3}BAAI, 
  \textsuperscript{4}ECNU 
  \\
}

\begin{document}

\maketitle

\begin{abstract}
In the past year, Multimodal Large Language Models (MLLMs) have demonstrated remarkable performance in tasks such as visual question answering, visual understanding and reasoning. However, the extensive model size and high training and inference costs have hindered the widespread application of MLLMs in academia and industry. Thus, studying efficient and lightweight MLLMs has enormous potential, especially in edge computing scenarios. In this survey, we provide a comprehensive and systematic review of the current state of efficient MLLMs. Specifically, we summarize the timeline of representative efficient MLLMs, research state of efficient structures and strategies, and the applications. Finally, we discuss the limitations of current efficient MLLM research and promising future directions. Please refer to our GitHub repository for more details: \href{https://github.com/lijiannuist/Efficient-Multimodal-LLMs-Survey}{https://github.com/lijiannuist/Efficient-Multimodal-LLMs-Survey}.



\end{abstract}



\footnotetext[1]{* Equal contribution.}
\footnotetext[2]{\ \ Yizhang Jin is an intern in Tencent, and Jian Li is the project leader.}

\section{Introduction}

Large-scale pretraining, a leading approach in Artificial Intelligence(AI), has seen general-purpose models like large language and multimodal models outperform specialized deep learning models across many tasks. The remarkable abilities of Large Language Models (LLM) have inspired efforts to merge them with other modality-based models to enhance multimodal competencies. This concept is further supported by the remarkable success of proprietary models like OpenAI's GPT-4V~\cite{achiam2023gpt} and Google's Gemini\cite{team2023gemini}.
As a result, Multimodal Large Language Models (MLLMs) have emerged, including the mPLUG-Owl series\cite{ye2023mplugowl, ye2023mplugowl2}, InternVL~\cite{chen2023internvl}, EMU~\cite{sun2023generative}, LLaVA~\cite{liu2023llava}, InstructBLIP~\cite{dai2024instructblip}, MiniGPT-v2~\cite{chen2023minigpt-v2}, and MiniGPT-4\cite{zhu2023minigpt4}. These models circumvent the computational cost of training from scratch by effectively leveraging the pre-training knowledge of each modality. MLLMs inherit the cognitive capabilities of LLMs, showcasing numerous remarkable features such as robust language generation and transfer learning abilities. Moreover, by establishing strong representational connections and alignments with other modality-based models, MLLMs can process inputs from multiple modalities, significantly broadening their application scope.

\begin{figure}[!t]
\centering
\includegraphics[width=\linewidth]{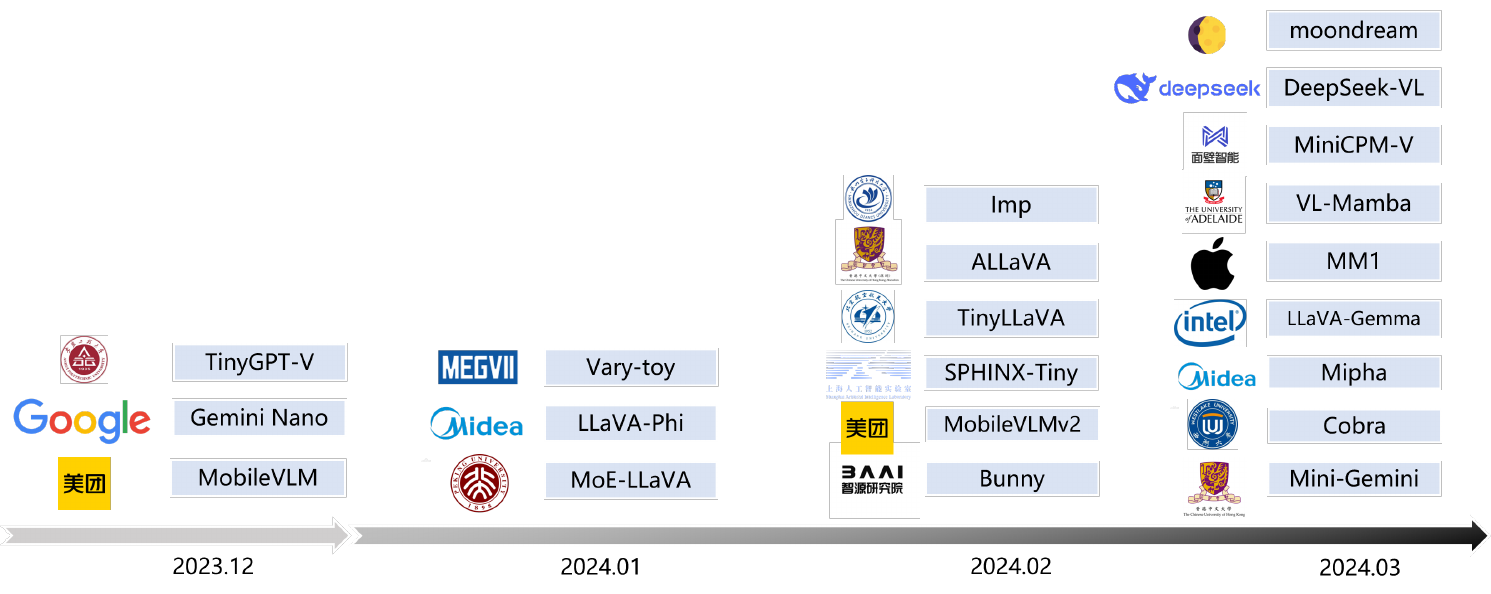}
\caption{The timeline of efficient MLLMs.}
\label{fig_1}
\end{figure}

The success of MLLMs is largely attributed to the scaling law: the performance of an AI model improves as more resources, such as data, computational power, or model size, are invested into it. However, scalability comes at the cost of high resource demands, which hinders the development and deployment of large models. For example, the training of MiniGPT-v2 necessitates a total of over 800 GPU hours, as calculated based on NVIDIA A100 GPUs~\cite{chen2023minigpt-v2}. This imposes a substantial expense that is difficult for researchers outside of major enterprises to bear. Aside from training, inference constitutes the major portion of resource consumption in mllm. Consider a typical scenario where the model input consists of an image with dimensions of \(336 \times 336\) pixels and a text prompt with a length of 40 tokens, performing inference with LLaVA-1.5 and a Vicuna-13B LLM backbone requires 18.2T FLOPS and 41.6G of memory usage. 
The resource-intensive nature of large-scale models has also sparked concerns about democratization and privacy protection, considering that the current mainstream MLLMs, represented by GPT-4V and Gemini, are controlled by a few dominant corporations and operate in the cloud. As demonstrated in the aforementioned experiments, even for open-source MLLMs, high requirements for computation resources make it challenging to run them on edge devices. This further exacerbates the challenges associated with ensuring equitable access and preserving user privacy.

\tikzstyle{my-box}=[
    rectangle,
    draw=hidden-draw,
    rounded corners,
    text opacity=1,
    minimum height=1.5em,
    minimum width=5em,
    inner sep=2pt,
    align=center,
    fill opacity=.5,
    line width=0.8pt,
]
\tikzstyle{leaf}=[my-box, minimum height=1.5em,
    fill=hidden-pink!80, text=black, align=left,font=\normalsize,
    inner xsep=2pt,
    inner ysep=4pt,
    line width=0.8pt,
]

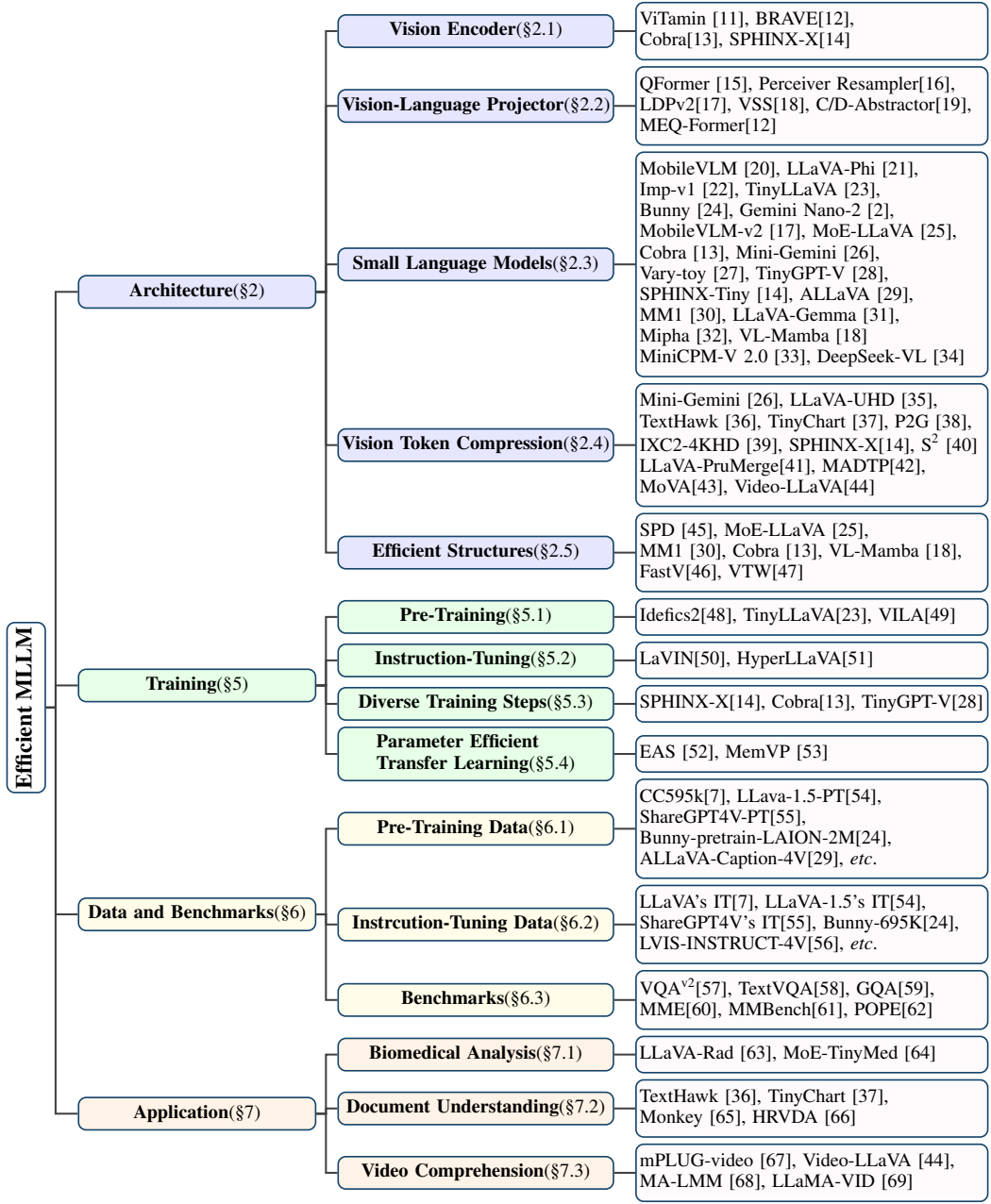
\begin{figure*}[ht]
    \centering
    \resizebox{\textwidth}{!}{
        \begin{forest}
            forked edges,
            for tree={
                grow=east,
                reversed=true,
                anchor=base west,
                parent anchor=east,
                child anchor=west,
                base=center,
                font=\large,
                rectangle,
                draw=hidden-draw,
                rounded corners,
                align=left,
                text centered,
                minimum width=4em,
                edge+={darkgray, line width=1pt},
                s sep=3pt,
                inner xsep=2pt,
                inner ysep=3pt,
                line width=0.8pt,
                ver/.style={rotate=90, child anchor=north, parent anchor=south, anchor=center},
            },
            where level=1{text width=12em,font=\normalsize,}{},
            where level=2{text width=14em,font=\normalsize,}{},
            where level=3{text width=16em,font=\normalsize,}{},
            where level=4{text width=18em,font=\normalsize,}{},
            where level=5{text width=18em,font=\normalsize,}{},
            [
                \textbf{Efficient MLLM}, ver
                [
                        \textbf{Architecture}(\S\ref{sec:arch}), fill=blue!10    
                        [
                        \textbf{Vision Encoder}(\S\ref{subsec:ve}), fill=blue!10
                        [
                            ViTamin~\citep{chen2024vitamin}{,} 
                            BRAVE\cite{kar2024brave}{,}
                            \\Cobra\cite{zhao2024cobra}{,}
                            SPHINX-X\cite{gao2024sphinx}, leaf, text width=18em
                        ]
                        ]
                        [
                            \textbf{Vision-Language Projector}(\S\ref{subsec:vlp}), fill=blue!10
                            [
                                QFormer~\citep{li2023blip2}{,}
                                Perceiver Resampler\cite{alayrac2022flamingo}{,}
                                \\ LDPv2\cite{chu2024mobilevlmv2}{,}
                                VSS\cite{qiao2024vlmamba}{,}
                                C/D-Abstractor\cite{cha2023honeybee}{,}
                                \\MEQ-Former\cite{kar2024brave}, leaf, text width=18em
                            ]
                        ]
                        [
                         \textbf{Small Language Models}(\S\ref{subsec:slm}), fill=blue!10
                          [
                            MobileVLM~\cite{chu2023mobilevlmv1}{,}
                            LLaVA-Phi~\cite{zhu2024llava-phi}{,}
                            \\Imp-v1~\cite{imp2024}{,}
                            TinyLLaVA~\cite{zhou2024tinyllava}{,}
                            \\Bunny~\cite{he2024bunny}{,} Gemini Nano-2~\cite{team2023gemini}{,}
                            \\MobileVLM-v2~\cite{chu2024mobilevlmv2}{,}
                            MoE-LLaVA~\cite{lin2024moe-llava}{,}
                            \\Cobra~\cite{zhao2024cobra}{,}
                            Mini-Gemini~\cite{li2024mini-gemini}{,}
                            \\Vary-toy~\cite{wei2024vary-toy}{,}
                            TinyGPT-V~\cite{yuan2023tinygpt-v}{,}
                            \\SPHINX-Tiny~\cite{gao2024sphinx}{,}
                            ALLaVA~\cite{chen2024allava}{,}
                            \\MM1~\cite{mckinzie2024mm1}{,}
                            LLaVA-Gemma~\cite{hinck2024llava-gemma}{,}
                            \\Mipha~\cite{zhu2024mipha}{,}
                            VL-Mamba~\cite{qiao2024vlmamba}
                            \\MiniCPM-V 2.0~\cite{minicpm-v}{,}
                            DeepSeek-VL~\cite{lu2024deepseekvl}, leaf, text width=18em
                         ]
                        ]
                        [
                            \textbf{Vision Token Compression}(\S\ref{subsec:vtc}), fill=blue!10
                            [
                                Mini-Gemini~\cite{li2024mini-gemini}{,}
                                LLaVA-UHD~\cite{xu2024llavauhd}{,}
                                \\TextHawk~\cite{yu2024texthawk}{,}
                                TinyChart~\cite{zhang2024tinychart}{,}
                                P2G~\cite{chen2024p2g}{,}
                                \\IXC2-4KHD~\cite{internlmxcomposer2_4khd}{,}
                                SPHINX-X\cite{gao2024sphinx}{,}
                                $\text{S}^\text{2}$~\citep{shi2024S2}
                                \\LLaVA-PruMerge\cite{shang2024llavaprumerge}{,}
                                MADTP\cite{cao2024madtp}{,}
                                \\MoVA\cite{zong2024mova}{,}
                                Video-LLaVA\cite{lin2023videollava}, leaf, text width=18em
                            ]
                        ]
                        [
                         \textbf{Efficient Structures}(\S\ref{subsec:es}), fill=blue!10
                            [
                                SPD~\cite{gagrani2024speculative}{,}
                                 MoE-LLaVA~\citep{lin2024moe-llava}{,}
                                \\MM1~\cite{mckinzie2024mm1}{,}
                                Cobra~\citep{zhao2024cobra}{,}
                                VL-Mamba~\cite{qiao2024vlmamba}{,}
                                \\FastV\cite{chen2024fastv}{,}
                                VTW\cite{lin2024vtw}, leaf, text width=18em
                            ]
                    ]
                    ]                  
                [
                        \textbf{Training}(\S\ref{sec:train}), fill=green!10
                        [
                            \textbf{Pre-Training}(\S\ref{subsec:ptt}), fill=green!10
                            [
                            Idefics2\cite{laurençon2024idefics2}{,}
                            TinyLLaVA\cite{zhou2024tinyllava}{,}
                            VILA\cite{lin2023vila}, leaf, text width=18em
                            ]
                        ]
                        [
                            \textbf{Instruction-Tuning}(\S\ref{subsec:itt}), fill=green!10
                              [
                              LaVIN\cite{luo2024lavin}{,}
                               HyperLLaVA\cite{zhang2024hyperllava}, leaf, text width=18em
                              ]
                        ]
                        [
                            \textbf{Diverse Training Steps}(\S\ref{subsec:div}), fill=green!10
                              [
                              SPHINX-X\cite{gao2024sphinx}{,}
                              Cobra\cite{zhao2024cobra}{,}
                              TinyGPT-V\cite{yuan2023tinygpt-v}, leaf, text width=18em
                              ]
                        ]
                        [
                            \textbf{Parameter Efficient}
                            \\\textbf{Transfer Learning}(\S\ref{subsec:pe}), fill=green!10
                            [
                            EAS~\citep{wu2024EAS}{,}
                            MemVP~\cite{jie2024MemVP}, leaf, text width=18em
                            ]
                        ]
                    ]
                     [
                        \textbf{Data and Benchmarks}(\S\ref{datasets}), fill=yellow!10
                        [
                            \textbf{Pre-Training Data}(\S\ref{subsec:pt}), fill=yellow!10
                            [
                            CC595k\cite{liu2023llava}{,}
                            LLava-1.5-PT\cite{liu2023llava1.5}{,}
                            \\ShareGPT4V-PT\cite{chen2023sharegpt4v}{,}
                            \\Bunny-pretrain-LAION-2M\cite{he2024bunny}{,}
                            \\ALLaVA-Caption-4V\cite{chen2024allava}{, \textit{etc}.}, leaf, text width=18em
                            ]
                        ]
                        [
                            \textbf{Instrcution-Tuning Data}(\S\ref{subsec:it}), fill=yellow!10
                            [
                            LLaVA’s IT\cite{liu2023llava}{,}
                            LLaVA-1.5’s IT\cite{liu2023llava1.5}{,}
                            \\ShareGPT4V’s IT\cite{chen2023sharegpt4v}{,}
                            Bunny-695K\cite{he2024bunny}{,}
                            \\LVIS-INSTRUCT-4V\cite{wang2023see}{, \textit{etc}.}, leaf, text width=18em
                            ]
                        ][
                            \textbf{Benchmarks}(\S\ref{subsec:bench}), fill=yellow!10
                            [
                            VQA$^\text{v2}$\cite{goyal2017vqav2}{,}
                            TextVQA\cite{singh2019vqat}{,}
                            GQA\cite{hudson2019gqa}{,}
                            \\MME\cite{fu2024mme}{,}
                            MMBench\cite{liu2023mmbench}{,}
                            POPE\cite{li2023pope}, leaf, text width=18em
                            ]
                        ]
                    ]
                [
                    \textbf{Application}(\S\ref{sec:app}), fill=orange!10
                    [
                        \textbf{Biomedical Analysis}(\S\ref{subsec:bio}),  fill=orange!10
                        [
                        LLaVA-Rad~\citep{chaves2024llava-rad}{,}
                        MoE-TinyMed~\citep{jiang2024moetinymed}, leaf, text width=18em
                        ]
                    ]
                    [
                        \textbf{Document Understanding}(\S\ref{subsec:doc}),  fill=orange!10
                         [
                        TextHawk~\citep{yu2024texthawk}{,}
                        TinyChart~\citep{zhang2024tinychart}{,}
                        \\Monkey~\citep{li2024monkey}{,}
                        HRVDA~\citep{liu2024hrvda}, leaf, text width=18em
                        ]
                    ]
                    [
                        \textbf{Video Comprehension}(\S\ref{subsec:video}),  fill=orange!10
                         [
                        mPLUG-video~\citep{xu2023mplug2}{,}
                        Video-LLaVA~\citep{lin2023videollava}{,}
                        \\MA-LMM~\citep{he2024malmm}{,}
                        LLaMA-VID~\citep{li2023llamavid}, leaf, text width=18em
                        ]
                    ]
                ]
            ]
        \end{forest}
        }
    \vspace{-0mm}
    \caption{Organization of efficient multimodal large language models advancements.}
    \label{fig:efficient MLLMs structure}
    \vspace{0mm}
\end{figure*}

In light of these challenges, there has been growing attention on the study of efficient MLLMs. The primary objective of these endeavors is to decrease the resource consumption of MLLMs and broaden their applicability while minimizing performance degradation. Research on efficient MLLMs began with replacing large language models with lightweight counterparts and performing typical visual instruction tuning. Subsequent studies further enhanced capabilities and expanded use cases in the following ways: (1) lighter architectures were introduced with an emphasis on efficiency, aiming to reduce the number of parameters or computational complexity\cite{lin2024moe-llava, zhao2024cobra, qiao2024vlmamba}; (2) more specialized components were developed, focusing on efficiency optimizations tailored to advanced architectures or imbuing specific properties, such as locality\cite{cha2023honeybee,chu2024mobilevlmv2,kar2024brave}; and (3) support for resource-sensitive tasks was provided, with some works employing visual token compression to boost efficiency, enabling the transfer of MLLM capabilities to resource-intensive tasks such as high-resolution image and video understanding\cite{xu2024llavauhd, internlmxcomposer2_4khd, gao2024sphinx, shi2024S2}.

In this survey, we aim to present an exhaustive organization of the recent advancements in the rapidly evolving field of efficient MLLMs, as depicted in Figure.\ref{fig:efficient MLLMs structure}. We organize the literature in a taxonomy consisting of six primary categories, encompassing various aspects of efficient MLLMs, including \textbf{architecture}, \textbf{efficient vision}, \textbf{efficient LLMs}, \textbf{training}, \textbf{data and benchmarks}, and \textbf{applications}.

\begin{itemize}
    \item[\textbf{•}] \textbf{Architecture} focuses on the MLLM framework developed by efficient techniques to reduce the computational cost. The architecture is composed of multiple modality-based fundamental models, exhibits characteristics distinct from single-modal models, thus promoting the development of novel technologies. 
    \item[\textbf{•}] \textbf{Efficient Vision} explores optimizing efficient visual fracture extraction strategies, emphasizing methods that boost efficiency while maintaining accuracy. It addresses integrating high-quality visual data for effective cross-modal understanding.
    
    \item[\textbf{•}] \textbf{Efficient LLMs} explores these strategies of improving the computational efficiency and scalability of language models. It examines the trade-offs between model complexity and performance while suggesting promising avenues for balancing these competing factors. 
    \item[\textbf{•}] \textbf{Training} surveys the landscape of training methodologies that are pivotal in the development of efficient MLLMs. It addresses the challenges associated with the pre-training stage, instruction-tuning stage, and the overall training strategy for state-of-the-art results.
    \item[\textbf{•}] \textbf{Data and Benchmarks} evaluates the efficiency of datasets and benchmarks used in the evaluation of multimodal language models. It assesses the trade-offs between dataset size, complexity, and computational cost, while advocating for the development of benchmarks that prioritize efficiency and relevance to real-world applications.
    \item[\textbf{•}] \textbf{Application} investigates the practical implications of efficient MLLMs in various domains, emphasizing the balance between performance and computational cost. By addressing resource-intensive tasks such as high-resolution image understanding and medical question-answering, this section highlights the potential of efficient MLLMs to broaden their application scope and contribute to real-world problem-solving.
\end{itemize}

In summary, this survey delves into these research endeavors, exploring various strategies for making MLLMs more resource-efficient. We review the development history of efficient MLLMs, provide a taxonomy of the strategies for efficient MLLMs, and comprehensively compare the performance of existing efficient MLLMs.Through this exploration, we aspire to provide a comprehensive understanding of the current state-of-the-art, thereby illuminating the intricate nuances of this emerging field. Furthermore, this survey serves as a roadmap, highlighting potential avenues for future research, and fostering a deeper comprehension of the challenges and opportunities that lie ahead in the domain of efficient MLLMs.
In addition to the survey, we have established a GitHub repository where we compile the papers featured in the survey, organizing them with the same taxonomy at \href{https://github.com/lijiannuist/Efficient-Multimodal-LLMs-Survey}{https://github.com/lijiannuist/Efficient-Multimodal-LLMs-Survey}. We will actively maintain it and incorporate new research as it emerges.

\section{Architecture}
\label{sec:arch}

\begin{figure}[!t]
\centering
\includegraphics[width=0.95\linewidth]{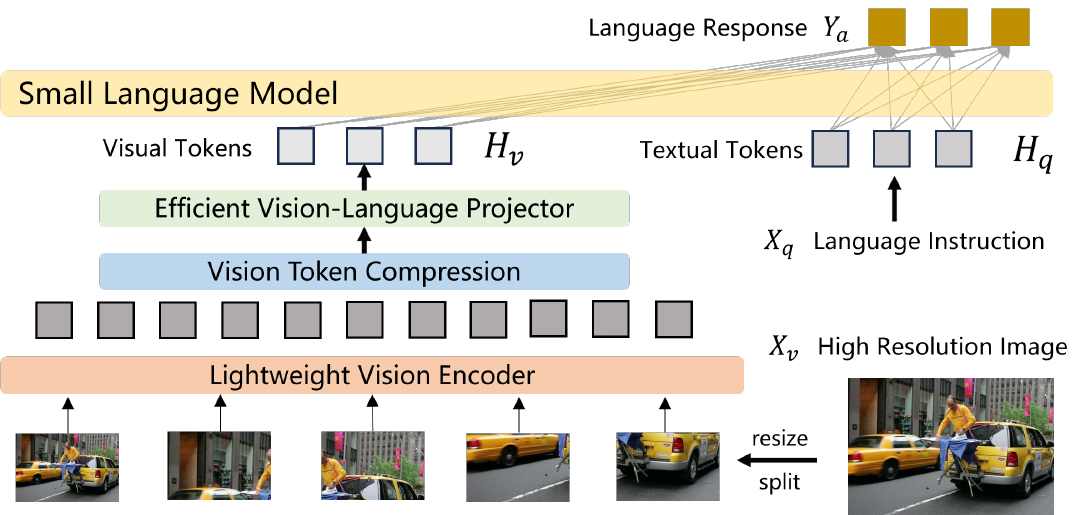}
\caption{The architectures of efficient MLLMs.}
\label{fig_2}
\end{figure}

Following the standard MLLM framework, efficient MLLMs can be divided into three main modules: a visual encoder $g$ tasked with receiving and processing visual inputs, a pre-trained language model that manages the received multimodal signals and performs reasoning, and a visual-language projector $P$ which functions as a bridge to align the two modalities. To enhance the efficiency of the general MLLMs, the primary optimization lies in handling high-resolution images, compressing vision tokens, implementing efficient structures, and utilizing compact language models, among other strategies. A diagram of the architecture is illustrated in Figure.~\ref{fig_2}. Table.~\ref{summarytable} surveys a summary of the efficient MLLMs, which outlines the base LLM, the vision encoder, image resolution, and the projector used to connect vision and language. These efficient MLLMs include: MobileVLM~\cite{chu2023mobilevlmv1}, LLaVA-Phi~\cite{zhu2024llava-phi}, Imp-v1~\cite{imp2024}, TinyLLaVA~\cite{zhou2024tinyllava}, Bunny~\cite{he2024bunny}, Gemini Nano-2~\cite{team2023gemini}, MobileVLM-v2~\cite{chu2024mobilevlmv2}, MoE-LLaVA-3.6B~\cite{lin2024moe-llava}, Cobra~\cite{zhao2024cobra}, Mini-Gemini~\cite{li2024mini-gemini}, Vary-toy~\cite{wei2024vary-toy}, TinyGPT-V~\cite{yuan2023tinygpt-v}, SPHINX-Tiny~\cite{gao2024sphinx}, ALLaVA~\cite{chen2024allava}, MM1-3B~\cite{mckinzie2024mm1}, LLaVA-Gemma~\cite{hinck2024llava-gemma}, Mipha-3B~\cite{zhu2024mipha}, VL-Mamba\cite{qiao2024vlmamba}, MiniCPM-V2.0~\cite{hu2024minicpm}, DeepSeek-VL~\cite{lu2024deepseekvl}, KarmaVLM~\cite{karmavlm}, moondream2~\cite{moondream}. In this section, we sequentially present a comprehensive overview of these three modules, along with other efficient components.

\begin{table}[ht]
\centering
\resizebox{\textwidth}{!}{
\begin{tabular}{lcccccc}
\toprule
\multirow{2}{*}{\textbf{Model}} & \multicolumn{3}{c}{\textbf{Vision Encoder}}                   & \multicolumn{2}{c}{LLM}            & \multirow{2}{*}{\textbf{Vision-LLM Projector}} \\
                       & Variants               & Resolution & Parameter Size & Variants          & Parameter Size &                                              \\ \midrule
MobileVLM~\cite{chu2023mobilevlmv1}              & CLIP ViT-L/14~\cite{radford2021clip}          & 336        & 0.3B           & MobileLLaMA\cite{chu2023mobilevlmv1}       & 2.7B           & LDP\cite{chu2023mobilevlmv1}                                          \\
LLaVA-Phi~\cite{zhu2024llava-phi}              & CLIP ViT-L/14~\cite{radford2021clip}          & 336        & 0.3B           & Phi-2\cite{javaheripi2023phi}             & 2.7B           & MLP                                          \\
Imp-v1~\cite{imp2024}                 & SigLIP~\cite{zhai2023siglip}                 & 384        & 0.4B           & Phi-2\cite{javaheripi2023phi}             & 2.7B           & -                                            \\
TinyLLaVA~\cite{zhou2024tinyllava}              & SigLIP-SO~\cite{zhai2023siglip}              & 384        & 0.4B           & Phi-2\cite{javaheripi2023phi}             & 2.7B           & MLP                                          \\
Bunny~\cite{he2024bunny}                  & SigLIP-SO~\cite{zhai2023siglip}              & 384        & 0.4B           & Phi-2\cite{javaheripi2023phi}             & 2.7B           & MLP                                          \\
MobileVLM-v2-3B~\cite{chu2024mobilevlmv2}        & CLIP ViT-L/14~\cite{radford2021clip}          & 336        & 0.3B           & MobileLLaMA\cite{chu2024mobilevlmv2}       & 2.7B           & LDPv2\cite{chu2024mobilevlmv2}                                        \\
MoE-LLaVA-3.6B~\cite{lin2024moe-llava}         & CLIP-Large~\cite{radford2021clip}             & 384        & -              & Phi-2\cite{javaheripi2023phi}             & 2.7B           & MLP                                          \\
Cobra~\cite{zhao2024cobra}                  & \makecell[c]{DINOv2~\cite{oquab2023dinov2} \\SigLIP-SO~\cite{zhai2023siglip}}     & 384        & 0.3B+0.4B      & Mamba-2.8b-Zephyr\cite{gu2023mamba} & 2.8B           & MLP                                          \\
Mini-Gemini~\cite{li2024mini-gemini}            & CLIP-Large~\cite{radford2021clip}             & 336        & -              & Gemma\cite{gemmateam2024gemma}             & 2B             & MLP                                          \\
Vary-toy~\cite{wei2024vary-toy}               & CLIP~\cite{radford2021clip}                   & 224        & -              & Qwen\cite{bai2023qwen}              & 1.8B           & -                                            \\
TinyGPT-V~\cite{yuan2023tinygpt-v}               & EVA~\cite{fang2023eva}                    & 224/448    & -              & Phi-2\cite{javaheripi2023phi}             & 2.7B           & Q-Former~\cite{li2023blip2}                                     \\
SPHINX-Tiny~\cite{gao2024sphinx}            & \makecell[c]{DINOv2~\cite{oquab2023dinov2} \\CLIP-ConvNeXt~\cite{liu2022convnet}} & 448        & -              & TinyLlama\cite{zhang2024tinyllama}         & 1.1B           & -                                            \\
ALLaVA-Longer~\cite{chen2024allava}          & CLIP-ViT-L/14~\cite{radford2021clip}          & 336        & 0.3B           & Phi-2\cite{javaheripi2023phi}             & 2.7B           & -                                            \\
MM1-3B-MoE-Chat~\cite{mckinzie2024mm1}        & CLIP$_\text{DFN}$-ViT-H~\cite{fang2023data}          & 378        & -              & -                  & 3B$^*$             & C-Abstractor~\cite{cha2023honeybee}                                 \\
LLaVA-Gemma~\cite{hinck2024llava-gemma}            & DinoV2~\cite{oquab2023dinov2}                 & -          & -              & Gemma-2b-it\cite{gemmateam2024gemma}       & 2B             & -                                            \\
Mipha-3B~\cite{zhu2024mipha}               & SigLIP~\cite{zhai2023siglip}                 & 384        & -              & Phi-2\cite{javaheripi2023phi}             & 2.7B           & -                                            \\
VL-Mamba~\cite{qiao2024vlmamba}               & SigLIP-SO~\cite{zhai2023siglip}                 & 384        & -              & Mamba-2.8B-Slimpj\cite{gu2023mamba}             & 2.8B           & VSS-L2\cite{qiao2024vlmamba}                                            \\
MiniCPM-V 2.0\cite{minicpm-v}              & SigLIP~\cite{zhai2023siglip}                 & -        & 0.4B              &  MiniCPM\cite{hu2024minicpm}             & 2.4B           & Perceiver Resampler~\cite{alayrac2022flamingo}                                            \\ 
DeepSeek-VL~\cite{lu2024deepseekvl}              & SigLIP-L~\cite{zhai2023siglip}                 & 384        & 0.4B              &  DeepSeek-LLM\cite{deepseek-llm}             &  1.3B           & MLP                                            \\ 
KarmaVLM\cite{karmavlm}              & SigLIP-SO~\cite{zhai2023siglip}                 & 384        & 0.4B              &  Qwen1.5\cite{bai2023qwen}             &  0.5B           & -                                           \\
moondream2\cite{moondream}              & SigLIP\cite{zhai2023siglip}                 & -        & -              &  Phi-1.5\cite{li2023phi}             &  1.3B           & -                                           \\ 
Bunny-v1.1-4B\cite{he2024bunny}              & SigLIP\cite{zhai2023siglip}                 & 1152$^\dagger$        & -              &  Phi-3-Mini-4K\cite{abdin2024phi3}             &   3.8B           & -                                           \\ \bottomrule
\end{tabular}%
}
\caption{The summary of 17 mainstream efficient MMLMs. $^*$ indicates activated parameters.$^\dagger$:High resolution support is achieved with S$^\text{2}$-Wrapper\cite{shi2024S2}.}
\label{summarytable}
\end{table}

\subsection{Vision Encoder}
\label{subsec:ve}
Taking the input image $X_v$ as input, the vision encoder compresses the original image into more compact patch features $Z_v$, as represented by the following formula:
\begin{equation}
Z_v = g(X_v).    
\end{equation} 
In line with mainstream MLLM practices, efficient MLLMs select pre-trained models that are semantically aligned with the text, represented by CLIP~\cite{radford2021clip}. This approach facilitates better alignment between the feature spaces of visual and text inputs. Since the vision encoder constitutes a relatively minor portion of the MLLM parameters, the advantages of lightweight optimization are less pronounced compared to the language model. Therefore, efficient MLLMs generally continue to employ visual encoders that are widely used in large-scale MLLMs, as detailed in Table \ref{summarytable}.

\begin{figure}[!t]
\centering
\includegraphics[width=0.9\linewidth]{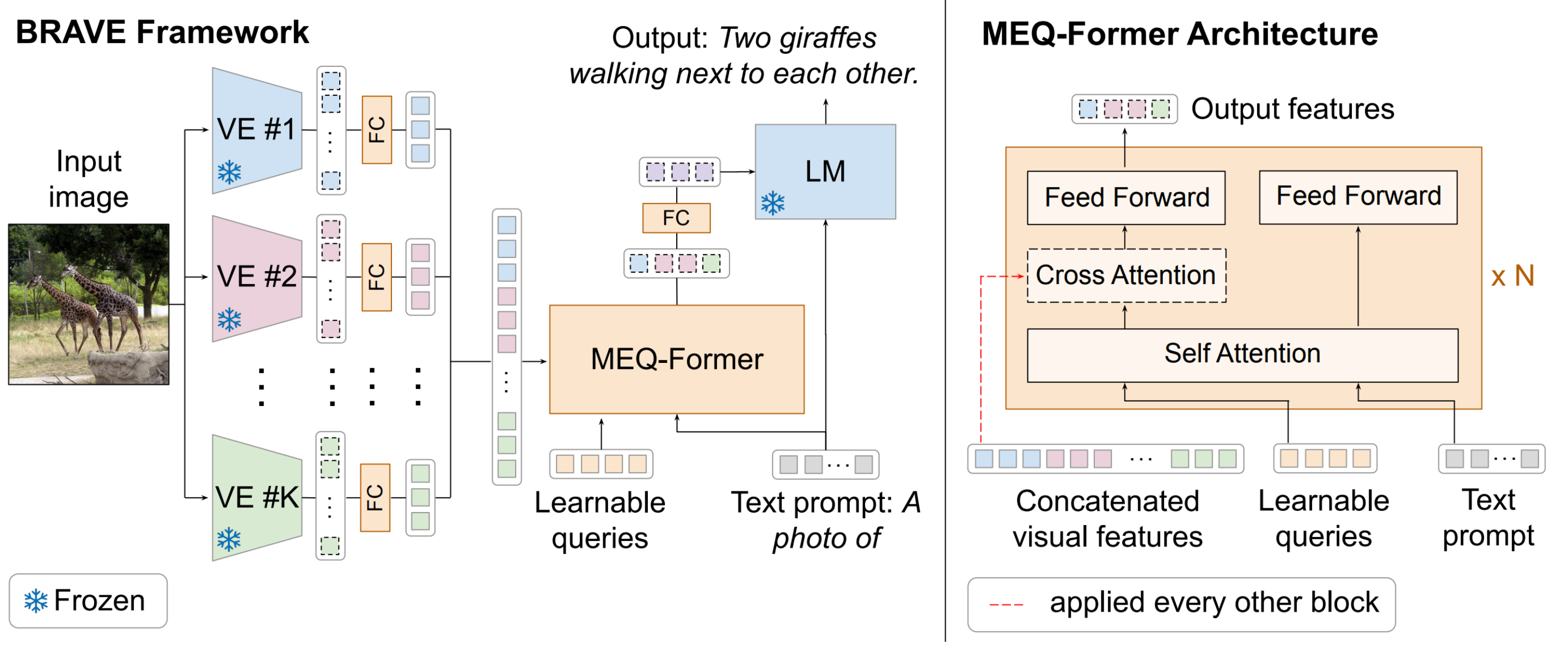}
\caption{\textbf{BRAVE}~\cite{kar2024brave} concatenates features from K different Vision Encoders in a sequence-wise manner. These concatenated features are then reduced by the MEQ-Former.}
\label{fig_brave}
\end{figure}

\paragraph{Multiple Vision Encoders}
BRAVE\cite{kar2024brave} in Figure.~\ref{fig_brave} performs an extensive ablation of various vision encoders with distinct inductive biases for tackling MLMM tasks. The results indicate that there isn't a single-encoder setup that consistently excels across different tasks, and encoders with diverse biases can yield surprisingly similar results. Presumably, incorporating multiple vision encoders contributes to capturing a wide range of visual representations, thereby enhancing the model's comprehension of visual data. Cobra\cite{zhao2024cobra} integrates DINOv2\cite{oquab2023dinov2} and SigLIP\cite{zhai2023siglip} as its vision backbone, with the rationale that merging the low-level spatial features from DINOv2 and the semantic attributes offered by SigLIP will enhance performance on subsequent tasks. SPHINX-X\cite{gao2024sphinx} employs two vision encoders – DINOv2 and CLIP-ConvNeXt. Given that these models have been pre-trained via distinct learning methodologies (self-supervised versus weakly supervised) and network architectures (ViT versus CNN), they are naturally capable of offering the most complementary and sophisticated visual knowledge.

 \paragraph{Lightweight Vision Encoder} 
Vision Transformer architectures in real-world applications pose challenges due to hardware and environmental limitations, including processing power and computational capabilities. ViTamin~\cite{chen2024vitamin} represents a lightweight vision model, specifically tailored for vision and language models. It commences with a convolutional stem, succeeded by Mobile Convolution Blocks in the first and second stages, and Transformer Blocks in the third stage. Remarkably, ViTamin-XL, with a modest count of 436M parameters, attains an 82.9\% ImageNet zero-shot accuracy. This outperforms the 82.0\% accuracy achieved by EVA-E~\cite{fang2023eva}, which operates with a parameter count ten times larger, at 4.4B. Simply replacing LLaVA's image encoder with ViTamin-L can establish new standards in various MLLM performance metrics.

\subsection{Vision-Language Projector}
\label{subsec:vlp}
The task of the vision-language projector is to map the visual patch embeddings $Z_v$ into the text feature space:
\begin{equation}
    H_v = P(Z_v),
\end{equation}
where $H_v$ denotes the projected visual embeddings. The aligned visual features are used as prompts and inputted into the language model along with the text embeddings. Vision-language projector avoids the high cost of training an end-to-end multimodal model from scratch and effectively leverages the capabilities of pre-trained language and vision models. 

\paragraph{MLP-based}As outlined in \cite{liu2023llava, liu2023llava1.5}, the vision-language projector is typically realized using a straightforward, learnable Linear Projector or a Multi-Layer Perceptron (MLP), i.e., several linear projectors interleaved with non-linear activation functions, as illustrated in Table.\ref{summarytable}. 

\paragraph{Attention-based}BLIP2~\cite{li2023blip2} introduces Q-Former, a lightweight transformer, which employs a set of learnable query vectors to extract visual features from a frozen vision model. Perceiver Resampler, proposed by Flamingo\cite{alayrac2022flamingo}, contemplates the use of learnable latent queries as Q in cross-attention, while image features are unfolded and concatenated with Q to serve as K and V in cross-attention. By this means, the transformer output at the corresponding positions of the learnable latent queries is taken as the aggregated representation of visual features, thereby standardizing variable-length video frame features into fixed-size features. MEQ-Former in BRAVE~\cite{kar2024brave} designs a multi-encoder querying transformer to amalgamate features from multiple frozen vision encoders into a versatile representation that can be directly inputted into a frozen language model.

\paragraph{CNN-based}MobileVLMv2\cite{chu2024mobilevlmv2} proposes LDPv2, a new projector consisting of three parts: feature transformation, token reduction, and positional information enhancement. By using point-wise convolution layers, average pooling, and a PEG module with a skip connection, LDPv2 achieves better efficiency, a 99.8\% reduction in parameters, and slightly faster processing compared to the original LDP\cite{chu2023mobilevlmv1}.
\paragraph{Mamba-based}VL-Mamba\cite{qiao2024vlmamba} implements the 2D vision selective scanning(VSS) technique within its vision-language projector, facilitating the amalgamation of diverse learning methodologies. The VSS module primarily resolves the distinct processing approaches between one-dimensional sequential processing and two-dimensional non-causal visual information.

\paragraph{Hybrid Structure}Honeybee~\cite{cha2023honeybee} put forward two visual projectors, namely C-Abstractor and D-Abstractor, which adhere to two primary design principles: (i) providing adaptability in terms of the number of visual tokens, and (ii) efficiently maintaining the local context. C-Abstractor, or Convolutional Abstractor, focuses on proficiently modeling the local context by employing a convolutional architecture. This structure consists of $L$ ResNet blocks, followed by adaptive average pooling and additional $L$ ResNet blocks, which facilitate the abstraction of visual features to any squared number of visual tokens. Conversely, D-Abstractor, or Deformable attention-based Abstractor utilizes deformable attention, which maintains the local context through a 2-D coordinate-based sampling process, using reference points and sampling offsets. 

\begin{figure}[!t]
\centering
\includegraphics[width=\linewidth]{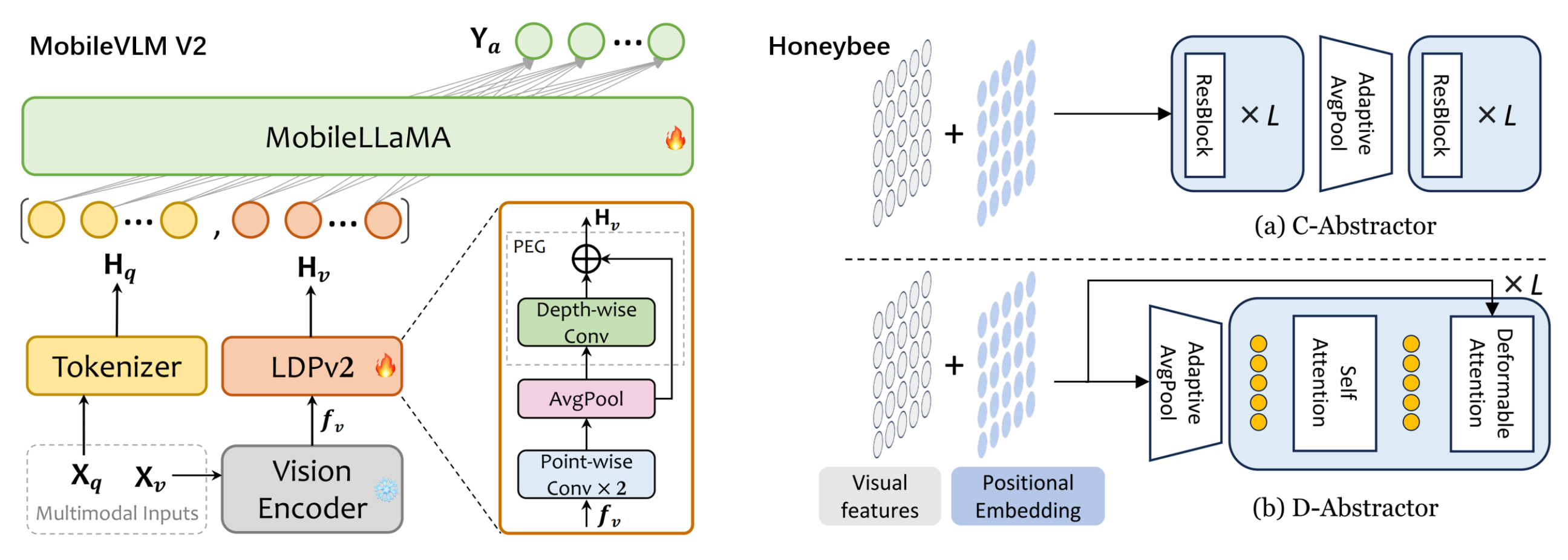}
\caption{\textbf{MobileVLM v2}~\cite{chu2024mobilevlmv2} and \textbf{Honeybee}~\cite{cha2023honeybee} efficient vision-language projector.}
\end{figure}

\subsection{Small Language Model}
\label{subsec:slm}
The pre-trained small language model(SLM) serves as the core component of MLLMs, endowing it with many outstanding capabilities, such as zero-shot generalization, instruction following, and in-context learning. The SLM accepts input sequences containing multiple modalities and outputs corresponding text sequences. A text tokenizer is typically bundled with the SLM, mapping text prompts $X_q$ to the text tokens $H_q$. The text tokens $H_q$ and the visual tokens $H_v$ are concatenated as the input of the language model, which outputs the final response sequence $Y_a$ in an autoregressive manner:
\begin{equation}
    p(Y_a|H_v,H_q)=\prod^L_{i=1}p(y_i|H_v,H_q,y_{<i}),
\end{equation}
\label{eq1}
where $L$ denotes the length of $Y_a$. As the SLM contributes the vast majority of MLLM parameters, its selection is closely related to the lightweight nature of MLLM. In comparison to conventional MLLMs with parameter sizes ranging from 7 billion to tens of billions\cite{touvron2023llama, chiang2023vicuna}, efficient MLLMs typically employ language models with less than 3 billion parameters, such as phi2-2.7B\cite{javaheripi2023phi} by Microsoft and Gemma-2B\cite{gemmateam2024gemma} by Google. Phi-2 trained on special data recipes can match the performance of models 25 times larger trained on regular data. Phi-3-mini~\cite{abdin2024phi3} can be easily deployed locally on a modern phone and achieves a quality that seems on-par with models such as Mixtral 8x7B~\cite{jiang2024moe} and GPT-3.5. In addition to utilizing pre-trained models, MobileVLM\cite{chu2023mobilevlmv1} downscales LLaMA\cite{touvron2023llama} and trains from scratch using open-source datasets. The specific model scaling is illustrated in the Table.\ref{summarytable} and Table.\ref{benchmarktable}.

\subsection{Vision Token Compression}
\label{subsec:vtc}
Initial research has underscored the potential of MLLMs across various tasks, including visual question answering and image captioning. However, MLLMs face considerable challenges in tasks necessitating intricate recognition, including crowd counting and OCR of small characters. A direct approach to address these challenges involves increasing the image resolution, practically, the number of visual tokens. This strategy, nonetheless, imposes a substantial computational burden on MLLMs, primarily due to the quadratic scaling of computational costs with the number of input tokens in the Transformer architecture.
Motivated by this challenge, vision token compression, aimed to reduce the prohibitive computation budget caused by numerous tokens, has become an essential aspect of efficient MLLMs. We will explore this topic through several key techniques, including multi-view input, token processing, multi-scale information fusion, vision expert agents and video-specific methods.

\paragraph{Multi-view Input} Directly employing high-resolution vision encoders for fine-grained perception is prohibitively costly and does not align with practical usage requirements. Therefore, to utilize low-resolution vision encoders while enabling MLLM to perceive detailed information, a common approach is to input multi-view HR images, \textit{i.e.}, a global view: low-resolution images obtained through resizing, and a local view: image patches derived from splitting. For example, LLaVA-UHD~\cite{xu2024llavauhd} proposes an image modularization strategy that divides native-resolution images into smaller variable-sized slices for efficient and extensible encoding. Inaddition, InternLM-XComposer2-4KHD~\cite{dong2024internlm} introduces a strategy that dynamically adjusts resolution with an automatic layout arrangement, which not only maintains the original aspect ratios of images but also adaptively alters patch layouts and counts, thereby enhancing the efficiency of image information extraction. By implementing an adaptive input strategy for images of varying resolutions, a balance between perceptual capability and efficiency can be achieved.

\paragraph{Token Processing} Techniques designed to process lengthy visual token squence are critical in efficient MLLMs as they address the dual challenges of preserving fine-grained details and reducing computational complexity. LLaVA-UHD~\cite{xu2024llavauhd} presents a novel approach to manage the computational burden associated with high-resolution images. It puts forward two key components: (1) a compression module that further condenses image tokens from visual encoders, significantly reducing the computational load, and (2) a spatial schema to organize slice tokens for LLMs. Notably, LLaVA-UHD demonstrates its efficiency by supporting 6 times larger resolution images using only 94\% of the inference computation compared to previous models. Furthermore, the model can be efficiently trained in academic settings, completing the process within 23 hours on 8 A100 GPUs. LLaVA-PruMerge\cite{shang2024llavaprumerge} and MADTP~\cite{cao2024madtp} propose an adaptive visual token reduction approach that significantly decreases the number of visual tokens while preserving comparable model performance. TinyChart~\cite{zhang2024tinychart} and TextHawk~\cite{yu2024texthawk} focus on document-oriented tasks, with the former adopting the Vision Token Merging module and the latter introducing the ReSampling and ReArrangement module. These modules can enhance fine-grained visual perception and information compression capabilities. 
\begin{figure}[!t]
\centering
\includegraphics[width=0.8\linewidth]{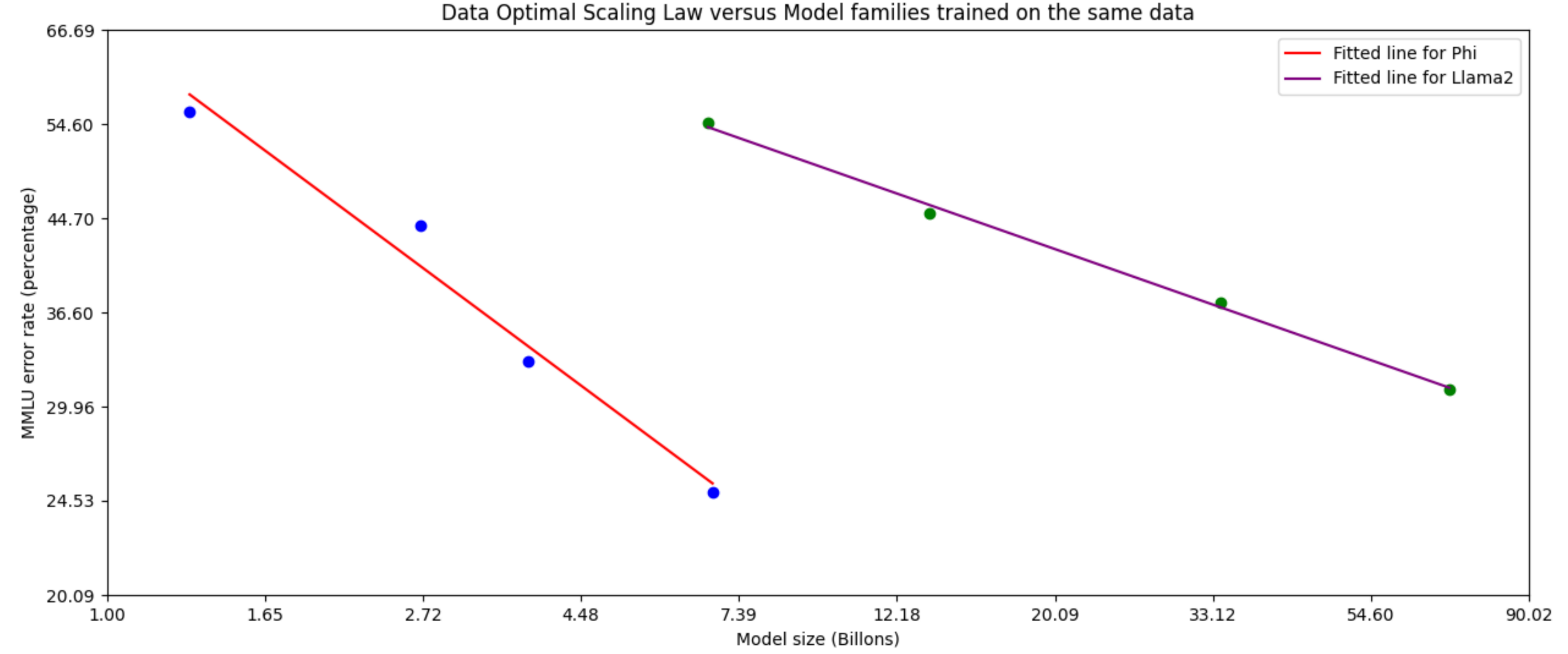}
\caption{Comparision of \textbf{Phi}~\cite{abdin2024phi3} (from left to right: phi-1.5, phi-2, phi-3-mini, phi-3-small) versus \textbf{Llama-2}~\cite{touvron2023llama2} family of models(7B, 13B, 34B, 70B) that were trained on the same fixed data.}
\end{figure}
\paragraph{Multi-Scale Information Fusion} Utilizing multi-scale image information is indeed crucial for visual feature extraction. This approach allows the model to capture both the fine-grained details present in smaller scales and the broader context available in larger scales. Mini-Gemini~\cite{li2024mini-gemini} comprises twin encoders, one for high-resolution images and the other for low-resolution visual embedding. It proposes Patch Info Mining, which uses low-resolution visual embeddings as queries to retrieve relevant visual cues from high-resolution candidates through cross-attention. Scaling on Scales ($S^2$)~\cite{shi2024S2} demonstrated that a multi-scale smaller model has comparable learning capacity to a larger model, and pre-training smaller models with $S^2$ can match or even exceed the advantage of larger models on MLLM benchmarks while being more compute-efficient. After splitting the large image into small sub-images, $S^2$-wrapper processes individual sub-images instead of using window attention, which allows using a pre-trained model that does not support window attention and avoids training additional parameters from scratch.It then interpolates the large feature map into the regular size, making sure the number of visual tokens stays acceptable.

\paragraph{Vision Expert Agents} Most MLLMs, due to their non-lossless image tokenization, struggle to fully capture the intricate details of text and objects. Leveraging vision expert agents is a solution to the problem of a single vision encoder's limited generalization ability on detail-abundant content. P2G~\cite{chen2024p2g} employs expert agents for real-time grounding, enabling efficient and purposeful reasoning through multimodal prompting. This innovative framework facilitates plug-and-play grounding of reasoning in high-resolution scenarios that are rich in natural visuals and text. It achieves this by leveraging agents to enhance both textual and visual grounding and perception, like OCR Agent (text) or Grounding Agent (image).
MoVA\cite{zong2024mova} addresses the issue of diminished generalization ability of an individual vision encoder across various contents by introducing an expert routing strategy. This approach enables the flexible and effective utilization of representations from multiple task-specific vision experts, thereby enhancing the generalization capabilities.

\begin{figure}[!t]
\centering
\includegraphics[width=0.9\linewidth]{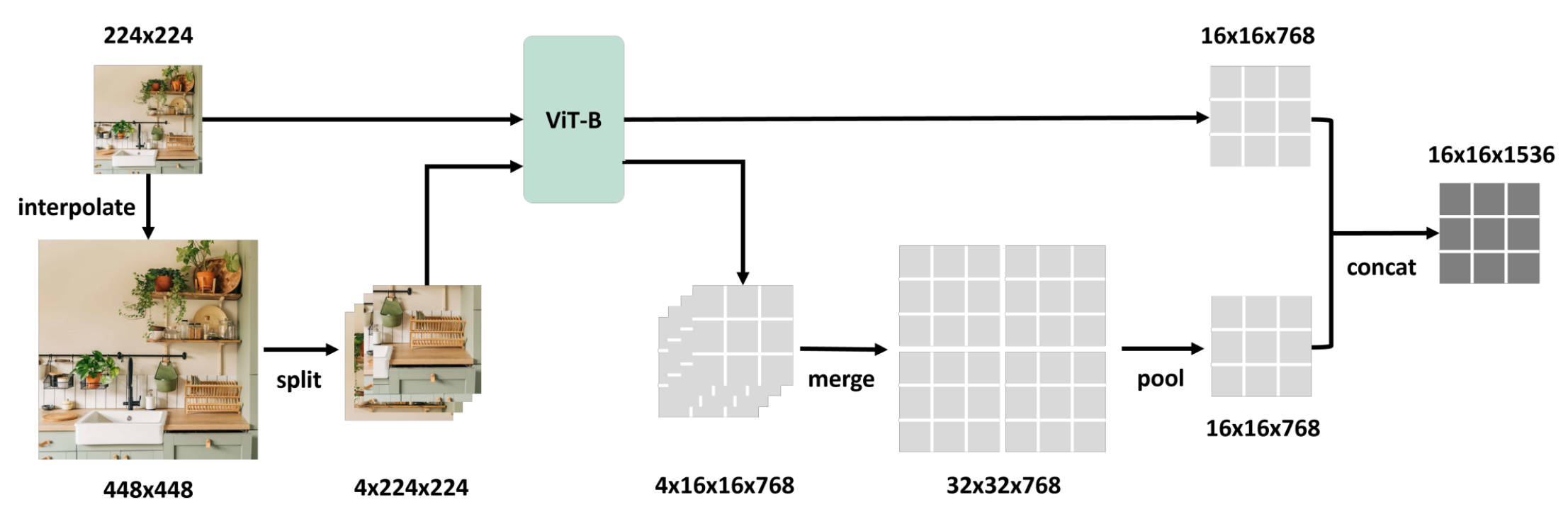}
\caption{\textbf{S$^\text{2}$-Wrapper}~\cite{shi2024S2} is a simple mechanism that extends any pre-trained vision model to multiple image scales in a parameter-free manner.}
\end{figure}

\paragraph{Video-Specific Methods} Video understanding also requires processing a large number of frames, which can pose a significant computational challenge within the context window of LLMs. Elysium~\cite{wang2024elysium} provides a trade-off between performance and visual token consumption, where T-Selector is introduced as a visual token compression network to enable LLMs to distinguish individual frames while reducing visual token use. VideoLLaVA~\cite{lin2023videollava}, building upon LanguageBind~\cite{zhu2023languagebind}, unifies visual representation into the language feature space to advance foundational LLMs towards a unified language-vision LLM without incurring a large computational burden.

\subsection{Efficient Structures}
\label{subsec:es}
Efficient structures primarily explore three directions: Mixture-of-Experts, Mamba and Inference Acceleration. 
\paragraph{Mixture of Experts}MoE enhances model capacity by modulating the total count of model parameters while maintaining the activated parameters unchanged, hence, not significantly compromising the inference speed. MoE-LLaVA\cite{lin2024moe-llava} presents an MoE-based sparse MLLM framework that effectively increases the number of parameters without compromising computational efficiency. Furthermore, it introduces MoE-Tuning, a three-stage training strategy designed to adapt MoE~\cite{jiang2024moe} to MLLMs and prevent model degradation caused by sparsity. 
MM1\cite{mckinzie2024mm1} designs two variants of MoE models. The first is a 3B-MoE model that employs 64 experts and substitutes a dense layer with a sparse one every two layers. The second is a 7B-MoE model that utilizes 32 experts and substitutes a dense layer with a sparse one every four layers.
\paragraph{Mamba}Cobra~\cite{zhao2024cobra} incorporates the efficient Mamba~\cite{gu2023mamba} language model into the vision modality and explores different modal fusion schemes to develop an effective multi-modal Mamba. Experiments show that it not only achieves competitive performance with state-of-the-art efficient methods but also boasts faster speeds due to its linear sequential modeling.It also excels in overcoming visual illusions and spatial relationship judgments in closed-set challenging prediction benchmarks and achieves performance comparable to LLaVA while using only 43\% of the parameters. VL-Mamba\cite{qiao2024vlmamba} substitutes the Transformer-based backbone language model with the pre-trained Mamba language model. It explores how to effectively implement the 2D vision selective scan mechanism for multimodal learning and the combinations of different vision encoders and pre-trained Mamba language model variants.

\paragraph{Inference Acceleration}
SPD\cite{gagrani2024speculative} proposes the speculative decoding with a language-only model to improve inference efficiency. By employing a language-only model as a draft model for speculative decoding, the need for image tokens and their associated processing components is bypassed. FastV~\cite{chen2024fastv} finds that most image tokens receive inefficient attention after the second decoder layer and achieve computation reduction by eliminating redundant visual tokens during the inference stage without sacrificing performance. VTW~\cite{lin2024vtw} asserts that visual tokens are not essential in the deeper layers of MLLM. It strategically removes all of them at a specific layer, allowing only text tokens to participate in the subsequent layers. This approach by VTW can reduce computational overhead by more than 40\% across a variety of multimodal tasks, without compromising performance.


\begin{figure}[!t]
\centering
\includegraphics[width=\linewidth]{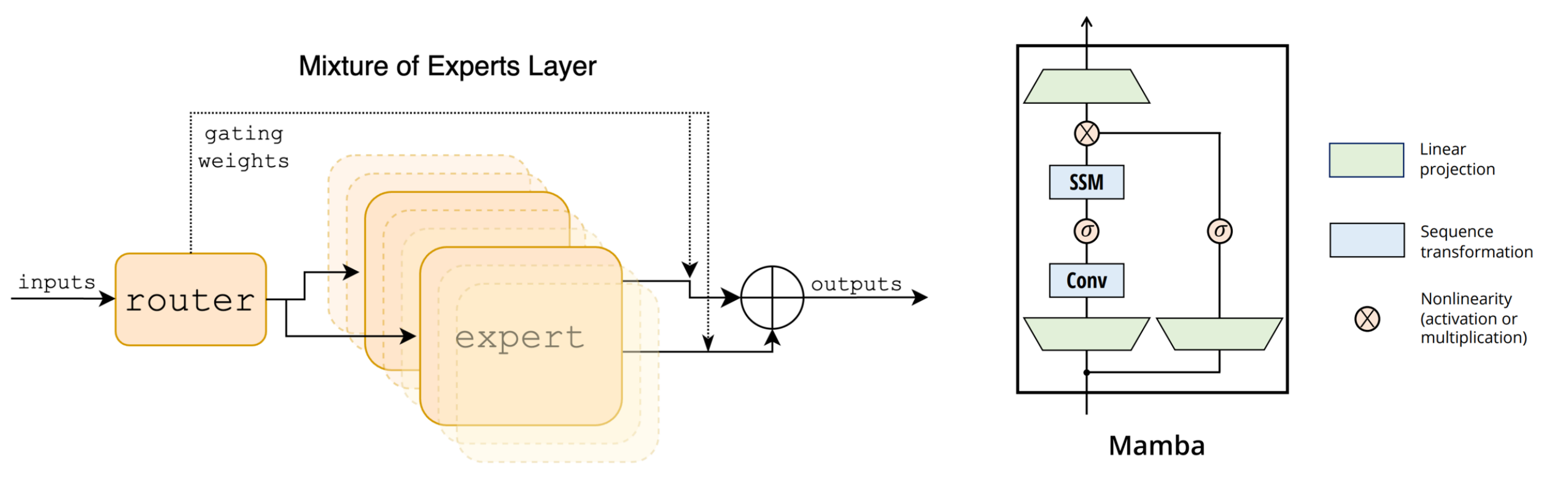}
\caption{Structure of \textbf{MOE}~\cite{jiang2024moe}(left) and \textbf{Mamba}~\cite{gu2023mamba}(right).}
\end{figure}

\section{Efficient Vision}
Vision Transformer (ViT)~\cite{dosovitskiy2020vit} architectures have gained significant popularity and are widely used in computer vision applications. However, as ViT models have grown in size, the number of trainable parameters and operations has also increased, impacting their deployment and performance. Additionally, the computational and memory cost of self-attention grows quadratically with image resolution. Referring to the paper~\cite{Papa_2024vit_survey}, this survey aims to explore the most efficient vision encoding methodologies that may be used for efficient MLLMs. 
\tikzstyle{my-box}=[
    rectangle,
    draw=hidden-draw,
    rounded corners,
    text opacity=1,
    minimum height=1.5em,
    minimum width=5em,
    inner sep=2pt,
    align=center,
    fill opacity=.5,
    line width=0.8pt,
]
\tikzstyle{leaf}=[my-box, minimum height=1.5em,
    fill=hidden-pink!80, text=black, align=left,font=\normalsize,
    inner xsep=2pt,
    inner ysep=4pt,
    line width=0.8pt,
]

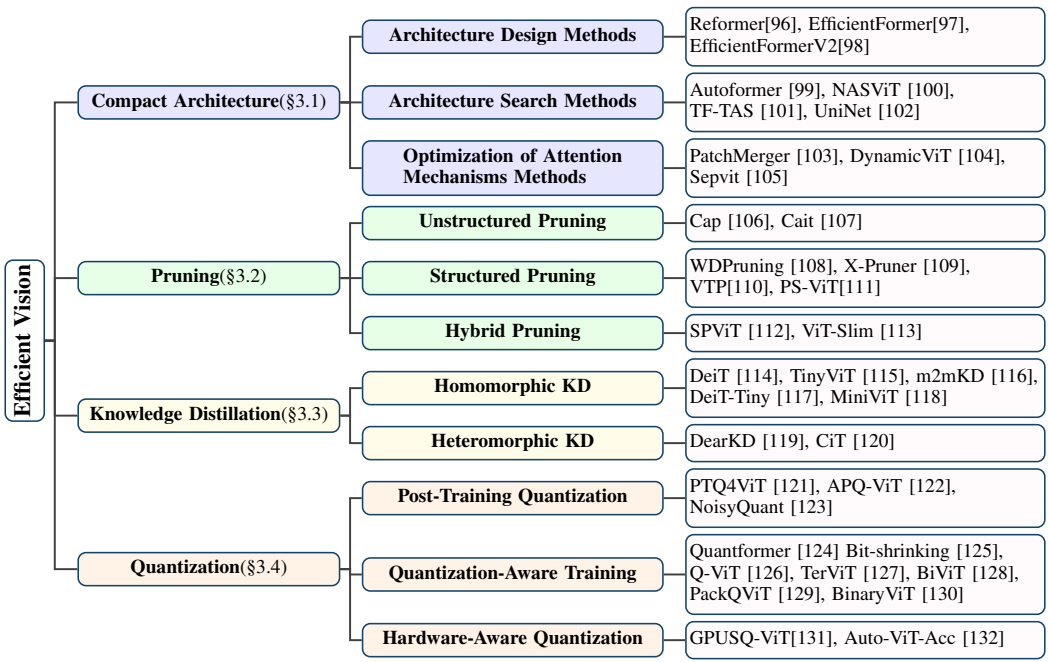
\begin{figure*}[ht]
    \centering
    \resizebox{\textwidth}{!}{
        \begin{forest}
            forked edges,
            for tree={
                grow=east,
                reversed=true,
                anchor=base west,
                parent anchor=east,
                child anchor=west,
                base=center,
                font=\large,
                rectangle,
                draw=hidden-draw,
                rounded corners,
                align=left,
                text centered,
                minimum width=4em,
                edge+={darkgray, line width=1pt},
                s sep=3pt,
                inner xsep=2pt,
                inner ysep=3pt,
                line width=0.8pt,
                ver/.style={rotate=90, child anchor=north, parent anchor=south, anchor=center},
            },
            where level=1{text width=13em,font=\normalsize,}{},
            where level=2{text width=15em,font=\normalsize,}{},
            where level=3{text width=16em,font=\normalsize,}{},
            where level=4{text width=18em,font=\normalsize,}{},
            where level=5{text width=18em,font=\normalsize,}{},
            [
                \textbf{Efficient Vision}, ver
                [
                        \textbf{Compact Architecture}(\S\ref{subsec:ca}), fill=blue!10    
                        [
                        \textbf{Architecture Design Methods}, fill=blue!10
                        [
                             Reformer\cite{kitaev2020reformer}{,}
                             EfficientFormer\cite{li2022efficientformer}{,}
                             \\ EfficientFormerV2\cite{li2023rethinking}, leaf, text width=18em
                        ]
                        ]
                        [
                            \textbf{Architecture Search Methods}, fill=blue!10
                            [
                              Autoformer~\cite{chen2021autoformer}{,}
                              NASViT~\cite{gong2022nasvit}{,}
                              \\TF-TAS~\cite{zhou2022training}{,}
                              UniNet~\cite{liu2022uninet}, leaf, text width=18em
                            ]
                        ]
                        [
                            \textbf{Optimization of Attention}
                            \\\textbf{Mechanisms Methods}, fill=blue!10
                            [
                                PatchMerger~\cite{renggli2022learning}{,}
                                DynamicViT~\cite{rao2021dynamicvit}{,}
                                \\Sepvit~\cite{li2022sepvit}, leaf, text width=18em
                            ]
                        ] 
                    ]                  
                [
                        \textbf{Pruning}(\S\ref{subsec:pru}), fill=green!10
                        [
                            \textbf{Unstructured Pruning}, fill=green!10
                            [
                            Cap~\cite{kuznedelev2024cap}{,}
                            Cait~\cite{wang2023cait}, leaf, text width=18em
                            ]
                        ]
                        [
                            \textbf{Structured Pruning}, fill=green!10
                              [
                             WDPruning~\cite{yu2022width}{,}
                             X-Pruner~\cite{yu2023x}{,}
                              \\VTP\cite{zhu2021vision}{,}
                              PS-ViT\cite{tang2022patch}, leaf, text width=18em
                              ]
                        ]
                        [
                            \textbf{Hybrid Pruning}, fill=green!10
                              [
                             SPViT~\cite{kong2022spvit}{,}
                             ViT-Slim~\cite{chavan2022vision}, leaf, text width=18em
                              ]
                        ]
                    ]
                     [
                        \textbf{Knowledge Distillation}(\S\ref{subsec:kd}), fill=yellow!10
                        [
                            \textbf{Homomorphic KD}, fill=yellow!10
                            [
                            DeiT~\cite{touvron2021training}{,}
                            TinyViT~\cite{wu2022tinyvit}{,}
                            m2mKD~\cite{lo2024m2mkd}{,}
                            \\DeiT-Tiny~\cite{hao2022learning}{,}
                             MiniViT~\cite{zhang2022minivit}, leaf, text width=18em
                            ]
                        ]
                        [
                            \textbf{Heteromorphic KD}, fill=yellow!10
                            [
                            DearKD~\cite{chen2022dearkd}{,}
                            CiT~\cite{ren2022co}, leaf, text width=18em
                            ]
                        ]
                    ]
                [
                    \textbf{Quantization}(\S\ref{subsec:quant}), fill=orange!10
                    [
                        \textbf{Post-Training Quantization},  fill=orange!10
                        [
                        PTQ4ViT~\cite{yuan2022ptq4vit}{,}
                        APQ-ViT~\cite{ding2022towards}{,}
                        \\NoisyQuant~\cite{liu2023noisyquant}, leaf, text width=18em
                        ]
                    ]
                    [
                        \textbf{Quantization-Aware Training},  fill=orange!10
                         [
                        Quantformer~\cite{wang2022quantformer}
                        Bit-shrinking~\cite{lin2023bit}{,}
                        \\Q-ViT~\cite{li2022q}{,}
                         TerViT~\cite{xu2022tervit}{,}
                         BiViT~\cite{he2023bivit}{,}
                         \\PackQViT~\cite{dong2024packqvit}{,}
                         BinaryViT~\cite{le2023binaryvit}, leaf, text width=18em
                        ]
                    ]
                    [
                        \textbf{Hardware-Aware Quantization},  fill=orange!10
                         [
                        GPUSQ-ViT\cite{yu2023boost}{,}
                        Auto-ViT-Acc~\cite{li2022auto}, leaf, text width=18em
                        ]
                    ]
                ]
            ]
        \end{forest}
        }
    \vspace{-0mm}
    \caption{Organization of efficient vision advancements.}
    \label{fig:efficient Vision}
    \vspace{0mm}
\end{figure*}

\subsection{Compact Architecture} 
\label{subsec:ca}
Compact Architecture refers to the design of lightweight and efficient models while maintaining high performance in downstream tasks. It encompasses various strategies and methodologies to reduce model size, computational complexity, and memory footprint without compromising performance. These strategies can be broadly categorized into three categories, 1) Architecture Design Methods, 2) Architecture Search Methods, and 3) Optimization of Attention Mechanisms Methods.

\textbf{Architecture Design Methods} involve creating new architectures~\cite{wang2023crossformer++} or adjusting existing ones~\cite{hassani2021escaping} to achieve compactness without sacrificing performance. For example, Reformer~\cite{kitaev2020reformer} introduced locality-sensitive hashing in attention mechanisms to reduce complexity, while also employing reversible residual layers to store activations more efficiently. Furthermore, EfficientFormer~\cite{li2022efficientformer} analyzed ViT-based model architectures and operators, introducing a dimension-consistent pure transformer paradigm and employing latency-driven slimming to produce optimized models. Additionally, EfficientFormerV2~\cite{li2023rethinking} proposed a supernet with low latency and high parameter efficiency. 

\paragraph{Architecture Search Methods} involve employing neural architecture search algorithms~\cite{chavan2022vision} to explore and discover compact architectures tailored to specific tasks or constraints. For instance, Autoformer~\cite{chen2021autoformer} intertwined weights within layers, enabling thorough training of thousands of subnets. NASViT~\cite{gong2022nasvit} introduced a gradient projection algorithm, switchable layer scaling, and streamlined data augmentation, enhancing convergence and performance. Additionally, TF-TAS~\cite{zhou2022training} investigated training-free architecture search methods, proposing an efficient scheme. UniNet~\cite{liu2022uninet} introduced context-aware down-sampling modules improving information accommodation by transformer and MLP operators.

\paragraph{Optimization of Attention Mechanisms Methods} focus on reducing computational complexity by introducing adaptive attention, learning sparse attention patterns, and dynamically adjusting attention mechanisms. Fayyaz \textit{et al.}~\cite{fayyaz2022adaptive} implemented adaptive attention by scoring and adaptively sampling significant tokens. PatchMerger~\cite{renggli2022learning} extracted global information among regional tokens and exchanged local self-attention with information among regional tokens via self-attention. DynamicViT~\cite{rao2021dynamicvit} proposed an attention masking strategy to differentiably prune tokens by blocking interactions with other tokens. Additionally, Sepvit~\cite{li2022sepvit} conducted local-global information interaction within and across windows using depthwise separable self-attention. These methods collectively optimize attention mechanisms, enhancing computational efficiency and performance.

\subsection{Pruning}
\label{subsec:pru}
Pruning involves removing less essential weights from vision transformer models, typically categorized as unstructured pruning, structured pruning, and hybrid pruning techniques. 

\paragraph{Unstructured Pruning} focuses on eliminating individual weights without considering their structural arrangement within the model. Rao \textit{et al}.~\cite{rao2021dynamicvit} introduced a dynamic token sparsification framework for progressive and adaptive pruning of redundant tokens based on input, integrating a lightweight prediction module to estimate token importance scores and employing an attention masking strategy to differentiate token interactions and optimize the prediction module in an end-to-end fashion. Cap~\cite{kuznedelev2024cap} proposed a novel theoretically-grounded pruner capable of accurately and efficiently handling intricate weight correlations during pruning, alongside an effective fine-tuning procedure for post-compression recovery. Cait~\cite{wang2023cait} introduced asymmetric token merging to integrate neighboring tokens efficiently while preserving the spatial structure, paired with consistent dynamic channel pruning for uniform pruning of unimportant channels in Vision Transformers, enhancing model compression.

\paragraph{Structured Pruning} aims to remove structural components, such as attention heads or layers based on predefined criteria. For example, WDPruning~\cite{yu2022width} employed a binary mask to discern insignificant parameters based on their magnitudes. Additionally, Yu \textit{et al}.~\cite{yu2022unified} presented a unified framework integrating pruning to generate compact transformers. X-Pruner~\cite{yu2023x} utilizes an end-to-end learned explainability-aware mask to measure each unit's contribution to predicting target classes and adaptively searches layer-wise thresholds to preserve the most informative unit while determining the pruning rate. Additionally, VTP \cite{zhu2021vision} reduces embedding dimensions through the integration of control coefficients, concurrently removing neurons with negligible coefficients. Tang \textit{et al.} \cite{tang2022patch} eliminate redundant patches by first identifying effective patches in the last layer and then leveraging them to guide the selection process of previous layers, where patches with minimal impact on the final output feature are subsequently discarded.

\paragraph{Hybrid Pruning}, such as~\cite{chen2021chasing}, investigates both unstructured and structured sparsity, introducing a first-order importance approximation approach for attention head removal. SPViT~\cite{kong2022spvit} develops a dynamic attention-based multi-head token selector for adaptive instance-wise token selection, alongside a soft pruning technique consolidating less informative tokens into package tokens rather than discarding them. ViT-Slim~\cite{chavan2022vision} utilizes a learnable and unified sparsity constraint with pre-defined factors to represent global importance within the continuous search space across various dimensions.

\begin{figure}[!t]
\centering
\includegraphics[width=\linewidth]{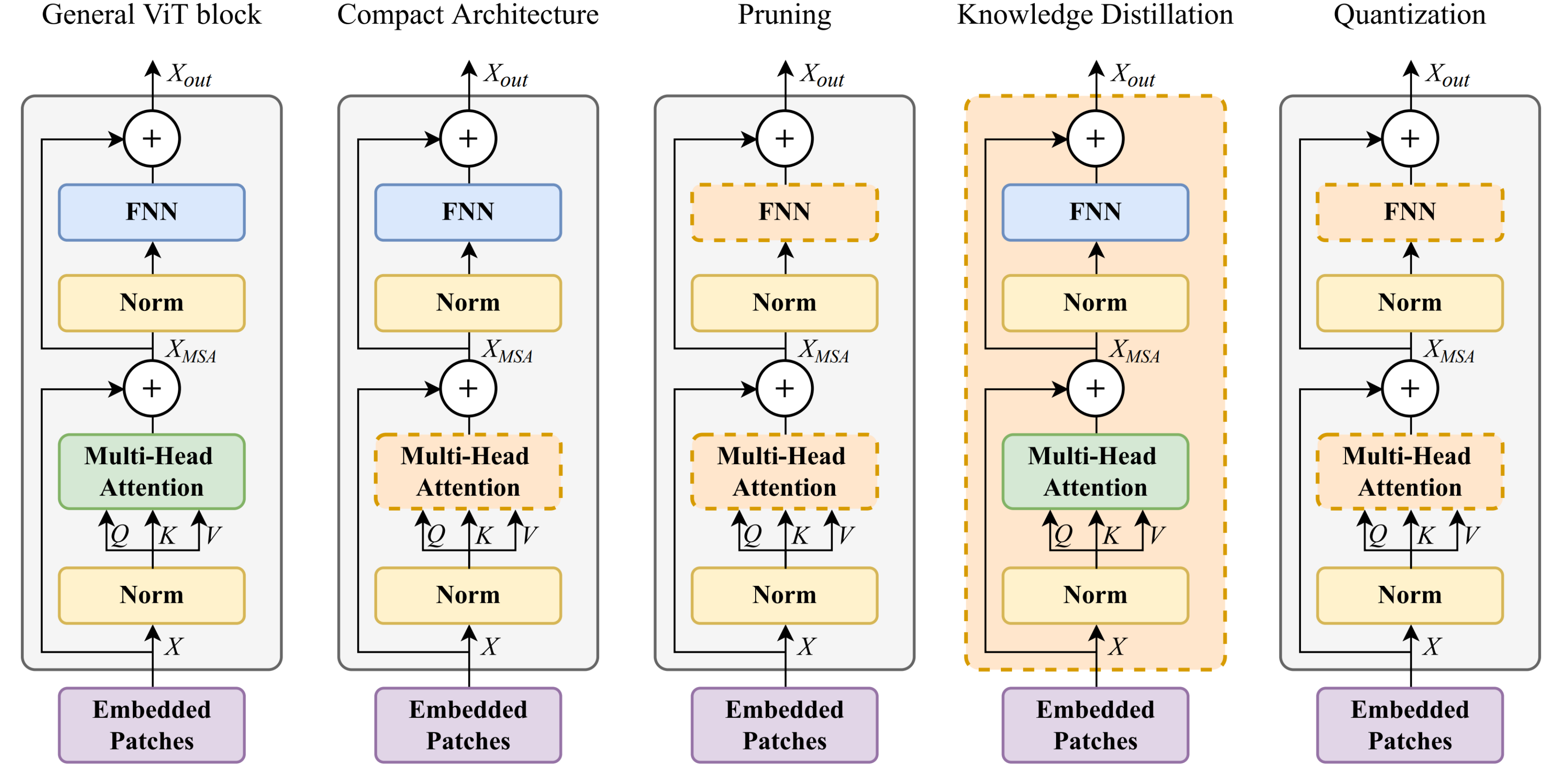}
\caption{Efficient vision transformer techniques in ~\cite{papa2024survey}. The dashed orange block highlights the component on which each optimization technique mainly focuses.}
\end{figure}

\subsection{Knowledge Distillation}
\label{subsec:kd}
Knowledge distillation is a technique in which a smaller model learns from a larger, more complex model to replicate its performance, enabling efficient deployment while maintaining predictive accuracy~\cite{hinton2015distilling}. Knowledge distillation (KD) techniques for Vision Transformers (ViTs) can be categorized into two main types: 1) homomorphic KDs and 2) heteromorphic KDs. 

\paragraph{Homomorphic KDs} can further classified into logit-level~\cite{touvron2021training,wu2022tinyvit}, patch-level~\cite{hao2022learning}, module-level~\cite{lo2024m2mkd}, and feature-level KDs~\cite{zhang2022minivit}. For logit-level methods, in DeiT~\cite{touvron2021training}, a distillation token is incorporated into the self-attention module to emulate the class label inferred by the teacher model, facilitating interaction between the student attention and layers, thus enabling the learning of hard labels during back-propagation. TinyViT~\cite{wu2022tinyvit} applies distillation during pretraining, where logits from large teacher models are pre-stored in the hardware, enabling memory and computational efficiency when transferring knowledge to scaled-down student transformers. Patch-level techniques like DeiT-Tiny~\cite{hao2022learning} train a small student model to match a pre-trained teacher model on patch-level structures, then optimize with a decomposed manifold matching loss for reduced computational costs. Module-level methods involve segregating teacher modules from a pre-trained unified model, and student modules from a modular model. In m2mKD~\cite{lo2024m2mkd}, these modules are combined with a shared meta-model, allowing the student module to emulate the behavior of the teacher module. 
Feature-level KD methods, as demonstrated by MiniViT~\cite{zhang2022minivit}, combine the weights of consecutive transformer blocks. This entails sharing weights across layers while introducing transformations to enhance diversity. Additionally, weight distillation over self-attention is utilized to transfer knowledge from large-scale ViT models to compact models with multiplexed weights.

\paragraph{Heteromorphic KDs} involves transferring knowledge between models with differing architectures. For example, DearKD~\cite{chen2022dearkd} proposes a novel two-stage framework, DearKD, departing from traditional methods for ViT architectures. In the first stage, they use a vanilla KD strategy to transfer CNN features to the ViT student model, representing a heteromorphic transfer. In the subsequent phase, if real samples are limited, they introduce a boundary-preserving intra-divergence loss to enhance the process. Similarly, CiT~\cite{ren2022co} proposes a heteromorphic KD strategy, where knowledge is transferred from diverse models, such as a CNN and an involution neural network, resulting in improved performance for the ViT student model.

\subsection{Quantization}
\label{subsec:quant}
ViT quantization is the process of reducing the precision of numerical representations in ViT models, typically transitioning from floating-point to fixed-point arithmetic~\cite{du2024model}. This reduction in precision aims to decrease memory usage, computational complexity, and energy consumption while preserving model accuracy to an acceptable level. Current research can be mainly categorized into post-training quantization, quantization-aware training, and hardware-aware quantization.

\paragraph{Post-Training Quantization (PTQ)} compresses trained ViT models by converting their parameters from high-precision floating-point numbers to lower-precision fixed-point numbers, such as 8-bit integers. For example, Liu \textit{et al.}~\cite{liu2021post}  introduced a ranking loss method to identify optimal low-bit quantization intervals for weights and inputs, ensuring the functionality of the attention mechanism. They also conducted an analysis to understand the relationship between quantization loss in different layers and feature diversity, exploring a mixed-precision quantization approach leveraging the nuclear norm of each attention map and output feature. Additionally, PTQ4ViT~\cite{yuan2022ptq4vit} introduced the twin uniform quantization method to minimize quantization error on activation values following softmax and GELU functions, incorporating a Hessian-guided metric to enhance calibration accuracy. APQ-ViT~\cite{ding2022towards} proposed a unified Bottom-elimination Blockwise Calibration scheme to optimize the calibration metric, prioritizing crucial quantization errors and designing a Matthew-effect Preserving Quantization for Softmax to maintain the power-law character and attention mechanism functionality. NoisyQuant~\cite{liu2023noisyquant} proposes to add a fixed Uniform noisy bias to quantized values, the quantization error is significantly reduced under certain conditions. This technique successfully modifies heavy-tailed activation distributions to fit a given quantizer.

\paragraph{Quantization-Aware Training (QAT) } integrates quantization into the training cycle. This integration is particularly advantageous when scaling down to ultra-low bit precision, such as 4 bits or lower, where PTQ struggles with significant performance loss. For example, Quantformer~\cite{wang2022quantformer} leverages entropy information to maintain consistency in self-attention ranks and introduces a differentiable search mechanism to optimally group patch feature dimensions, reducing rounding and clipping inaccuracies. Q-ViT~\cite{li2022q} incorporates a distillation token and Information Rectification Module (IRM) to counteract altered distributions in quantized attention modules. TerViT~\cite{xu2022tervit} and Bit-shrinking~\cite{lin2023bit} progressively reduce model bit-width while regulating sharpness to maintain accuracy throughout quantization. PackQViT~\cite{dong2024packqvit} mitigates outlier effects during quantization. BiViT~\cite{he2023bivit} introduces Softmax-aware Binarization to adjust the binarization process, minimizing errors in binarizing softmax attention values. Xiao \textit{et al.}~\cite{xiao2023binaryvit} integrated a gradient regularization scheme to curb weight oscillation during binarization training and introduced an activation shift module to reduce information distortion in activations. Additionally, BinaryViT~\cite{le2023binaryvit} integrates essential architectural elements from CNNs into a pure ViT framework, enhancing its capabilities.

\paragraph{Hardware-Aware Quantization} optimizes the quantization process of neural network models for specific hardware platforms (\textit{e.g.}, GPUs~\cite{yu2023boost}, FPGA~\cite{li2022auto}). It adjusts precision levels and quantization strategies to maximize performance and energy efficiency during inference. For example, Yu \textit{et al.}~\cite{yu2023boost} propose a compression scheme utilizing GPU-friendly 2:4 fine-grained structured sparsity and quantization. They prune a dense model into a sparse one using 2:4 structured pruning, leveraging GPU acceleration. Then, they quantize the sparse model into fixed-point representation through sparse-distillation-aware quantization-aware training, exploiting GPU speedup. Throughout the process, they employ mixed-strategy knowledge distillation, enabling support for supervised and unsupervised learning styles. Auto-ViT-Acc~\cite{li2022auto} proposed a framework designed for quantizing ViT architectures to run inference on FPGA-powered devices. They apply the quantization function from a prior study specifically to the FNN module within the attention block, aiming to optimize FPGA resource utilization and accelerate inference.

\section{Efficient LLMs}
\tikzstyle{my-box}=[
    rectangle,
    draw=hidden-draw,
    rounded corners,
    text opacity=1,
    minimum height=1.5em,
    minimum width=5em,
    inner sep=2pt,
    align=center,
    fill opacity=.5,
    line width=0.8pt,
]
\tikzstyle{leaf}=[my-box, minimum height=1.5em,
    fill=hidden-pink!80, text=black, align=left,font=\normalsize,
    inner xsep=2pt,
    inner ysep=4pt,
    line width=0.8pt,
]

\begin{figure*}[ht]
    \centering
    \resizebox{\textwidth}{!}{
        \begin{forest}
            forked edges,
            for tree={
                grow=east,
                reversed=true,
                anchor=base west,
                parent anchor=east,
                child anchor=west,
                base=center,
                font=\large,
                rectangle,
                draw=hidden-draw,
                rounded corners,
                align=left,
                text centered,
                minimum width=4em,
                edge+={darkgray, line width=1pt},
                s sep=3pt,
                inner xsep=2pt,
                inner ysep=3pt,
                line width=0.8pt,
                ver/.style={rotate=90, child anchor=north, parent anchor=south, anchor=center},
            },
            where level=1{text width=10em,font=\normalsize,}{},
            where level=2{text width=12em,font=\normalsize,}{},
            where level=3{text width=16em,font=\normalsize,}{},
            where level=4{text width=18em,font=\normalsize,}{},
            where level=5{text width=18em,font=\normalsize,}{},
            [
                \textbf{Efficient LLM}, ver
                [
                        \textbf{Attention}(\S\ref{subsec:att}), fill=blue!10    
                        [
                        \textbf{Sharing-based Attention}, fill=blue!10
                            [
                            GQA~\citep{ainslie2023gqa}{,}
                            Multi-Query Attention\cite{shazeer2019fast}, leaf, text width=18em
                            ]
                        ]
                        [
                            \textbf{Feature Information}
                            \\ \textbf{Reduction}, fill=blue!10
                            [
                                Funnel-Transformer~\citep{dai2020funnel}{,}
                                \\Set Transformer\cite{lee2019settransformer}, leaf, text width=18em
                            ]
                        ]
                        [
                            \textbf{Approximate Attention}, fill=blue!10
                            [
                               Linformer~\cite{wang2020linformer}{,}
                               Performer\cite{choromanski2020masked}, leaf, text width=18em
                            ]
                        ]
                    ]                  
                [
                        \textbf{Framework}(\S\ref{subsec:fra}), fill=green!10
                        [
                            \textbf{Mixture of Experts}, fill=green!10
                            [
                            GShard\cite{lepikhin2020gshard}{,}
                            Switch Transformer \cite{fedus2022switch}, leaf, text width=18em
                            ]
                        ]
                        [
                            \textbf{Transformer-Alternative}
                            \\ \textbf{Architecture}, fill=green!10
                              [
                              RWKV\cite{peng2023rwkv}{,}
                              S4\cite{gu2021efficiently}{,}
                               \\DSS\cite{gupta2022diagonal}{,}
                               Mamba\cite{gu2023mamba}, leaf, text width=18em
                              ]
                        ]
                    ]
                     [
                        \textbf{Fine-tuning}(\S\ref{subsec:ft}), fill=yellow!10
                        [
                            \textbf{Parameter-Efficient} 
                            \\ \textbf{Fine-Tuning}, fill=yellow!10
                            [
                             LLM-Adapters \cite{hu2023llm}{,}
                            (IA)$^\text{3}$~\cite{liu2020tfew}{,}
                            \\LoRA-FA~\cite{zhang2023lora}{,}
                            DyLoRa~\cite{valipour2022dylora}, leaf, text width=18em
                            ]
                        ]
                        [
                            \textbf{Full-Parameter fine-tuning}, fill=yellow!10
                            [
                            LOMO~\cite{lv2023full}{,}
                            MeZO\cite{malladi2023mezo}, leaf, text width=18em
                            ]
                        ]
                    ]
            ]
        \end{forest}
        }
    \vspace{-0mm}
    \caption{Organization of efficient large language models advancements.}
    \label{fig:efficient LLMs}
    \vspace{0mm}
\end{figure*}
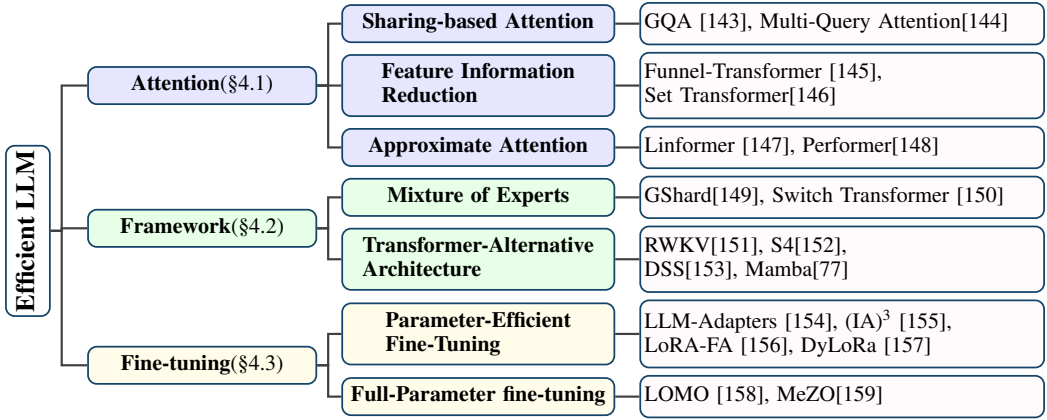

Occupying a significant majority of the parameter volume in MLLMs, LLM serves as a crucial entry point for enhancing the efficiency of MLLMs. In this section, similar to the survey paper~\cite{wan2024efficientllm_survey}, we provide a brief overview of the research progress in efficient LLMs, offering inspiration for the development of Efficient MLLMs.

\subsection{Attention}
\label{subsec:att}
In the standard self-attention mechanism, the time complexity is $O(n^2)$, where $n$ is the sequence length. This quadratic complexity arises due to the pairwise interactions between all input tokens, which can lead to scalability issues, especially when dealing with long sequences in LLMs. To tackle this, researchers have developed techniques to expedite attention mechanisms and reduce time complexity, such as sharing-based attention, feature information reduction, kernelization or low-rank, fixed and learnable pattern strategies, and hardware-assisted attention.

\paragraph{Sharing-based Attention} Sharing-based Attention aims to expedite attention computation during inference by by sharing computation resources across multiple Key-Value heads. For example, Llama-2~\cite{touvron2023llama2} incorporates a technique called grouped-query attention (GQA)~\cite{ainslie2023gqa} to optimize memory bandwidth during the autoregressive decoding. GQA is a Sharing-based Attention technique that seeks to achieve a balance between performance and efficiency, positioned between multi-head attention and multi-query attention~\cite{shazeer2019fast} mechanisms. In multi-head attention, each head utilizes a distinct set of linear transformation parameters for queries, keys, and values. Conversely, multi-query attention shares a single set of key-value heads across all queries. GQA partitions all query heads into several groups, with each group's query heads sharing a common key-value head, thereby establishing a rigorous equilibrium between effectiveness and computational cost.
\begin{figure}[!t]
\centering
\includegraphics[width=0.8\linewidth]{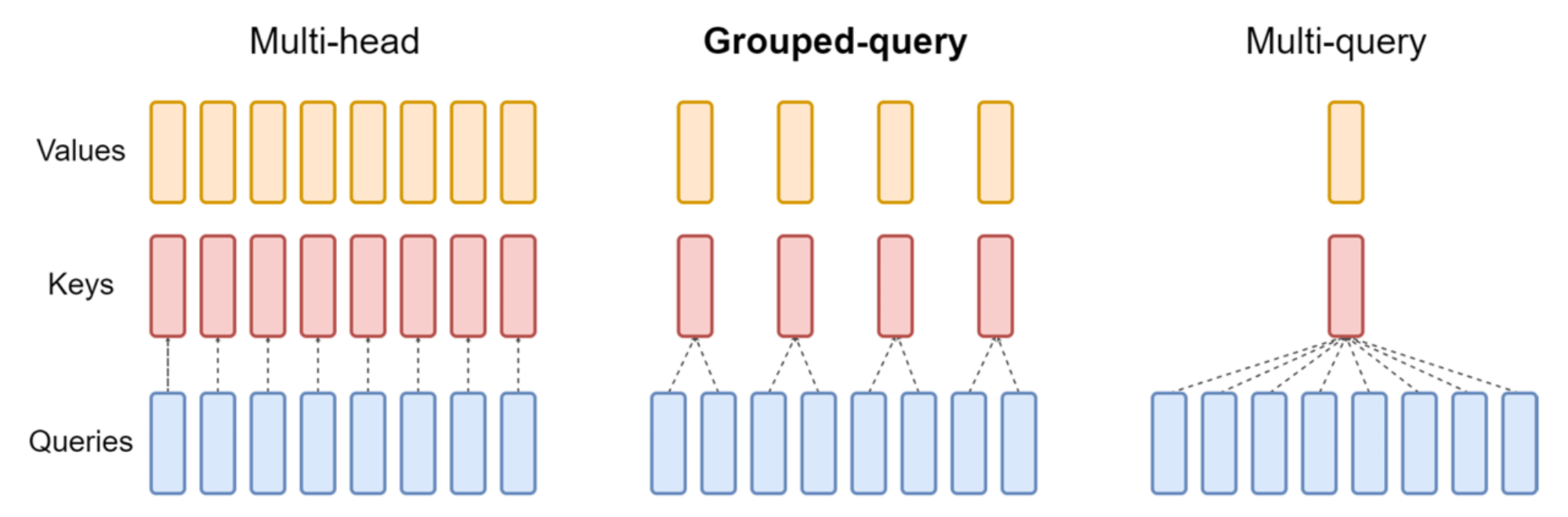}
\caption{In GQA\cite{hudson2019gqa}, a single set of key and value heads is allocated for each group of query heads, providing a balance between multi-head and multi-query attention mechanisms. }
\end{figure}

\paragraph{Feature Information Reduction}Feature Information Reduction, as evidenced by models
 such as Funnel-Transformer\cite{dai2020funnel} and Set Transformer\cite{lee2019settransformer}, addresses the crucial need for computational efficiency in attention mechanisms, specifically by reducing the dimensionality or quantity of input features while preserving the essential information embedded within the data. A key motivation behind this strategy stems from the potential redundancy in maintaining full-length hidden representations across all layers in Transformer models. Funnel-Transformer~\cite{dai2020funnel} tackles this issue by progressively reducing the sequence size of hidden representations in self-attention models, such as sequence length. This reduction not only decreases computational complexity and memory usage but also frees up resources that can be allocated toward building deeper or wider models.

\paragraph{Approximate Attention}Approximate Attention facilitates models to efficiently focus on task-relevant information when processing long texts. Two pivotal concepts within Approximate Attention are Kernelization and Low-Rank. Kernelization, e.g.~\cite{choromanski2020masked}, involves transforming a problem into a kernel-based framework with the goal of converting the original issue into a more manageable one in a higher-dimensional space. Kernelization is primarily employed to map text sequences into a high-dimensional space, where task-related information can be more readily captured. In this new space, each word in the text sequence is represented as a high-dimensional vector, and the distances between these vectors serve to measure their similarities. Low-Rank~\cite{wang2020linformer} aims to decompose a high-dimensional matrix into the product of two lower-dimensional matrices. Consequently, by calculating the inverses of these two lower-dimensional matrices, an approximate inverse of the attention matrix can be obtained, thereby significantly reducing computational complexity.

\subsection{Framework}
\label{subsec:fra}
\paragraph{Mixture of Experts}
The core idea behind MoE~\cite{jiang2024moe} is to decompose a large-scale model into several smaller models, each of which focuses on learning a specific part of the input data. During the training process, each expert is assigned a weight that determines its importance within the overall model. During the inference phase, given an input, all experts are ranked, and the most relevant ones are selected for computation. This approach considerably reduces the amount of computation, as only a subset of experts is involved in the calculation.By distributing computational tasks among different experts, MoE achieves more efficient utilization of computational resources during both training and inference phases. In MoE, each expert has its own set of parameters; however, these parameters are shared during the training process. This parameter-sharing strategy reduces the overall number of parameters in the model, consequently lowering storage and computational costs. GShard~\cite{lepikhin2020gshard} is a module composed of a set of lightweight annotation APIs and XLA compiler extensions, which offers an elegant way to express various parallel computation patterns while making minimal changes to existing model code. It enables us to scale multi-lingual neural machine translation Transformer models with sparse gated mixtures of experts to over 600 billion parameters using automatic sharding. Switch Transformer~\cite{fedus2022switch} replaces the feedforward network (FFN) layer in the standard Transformer with a MoE routing layer, where each expert operates independently on the tokens in the sequence. Its training speed is four times faster than Google's previously developed largest model, T5-XXL, under the same computational resources. The proposed training techniques have eliminated instability during the training process, demonstrating that large sparse models can also be trained in a low-precision format, such as bfloat16.

\paragraph{Transformer-Alternative Structures}

\begin{figure}[!t]
\centering
\includegraphics[width=0.9\linewidth]{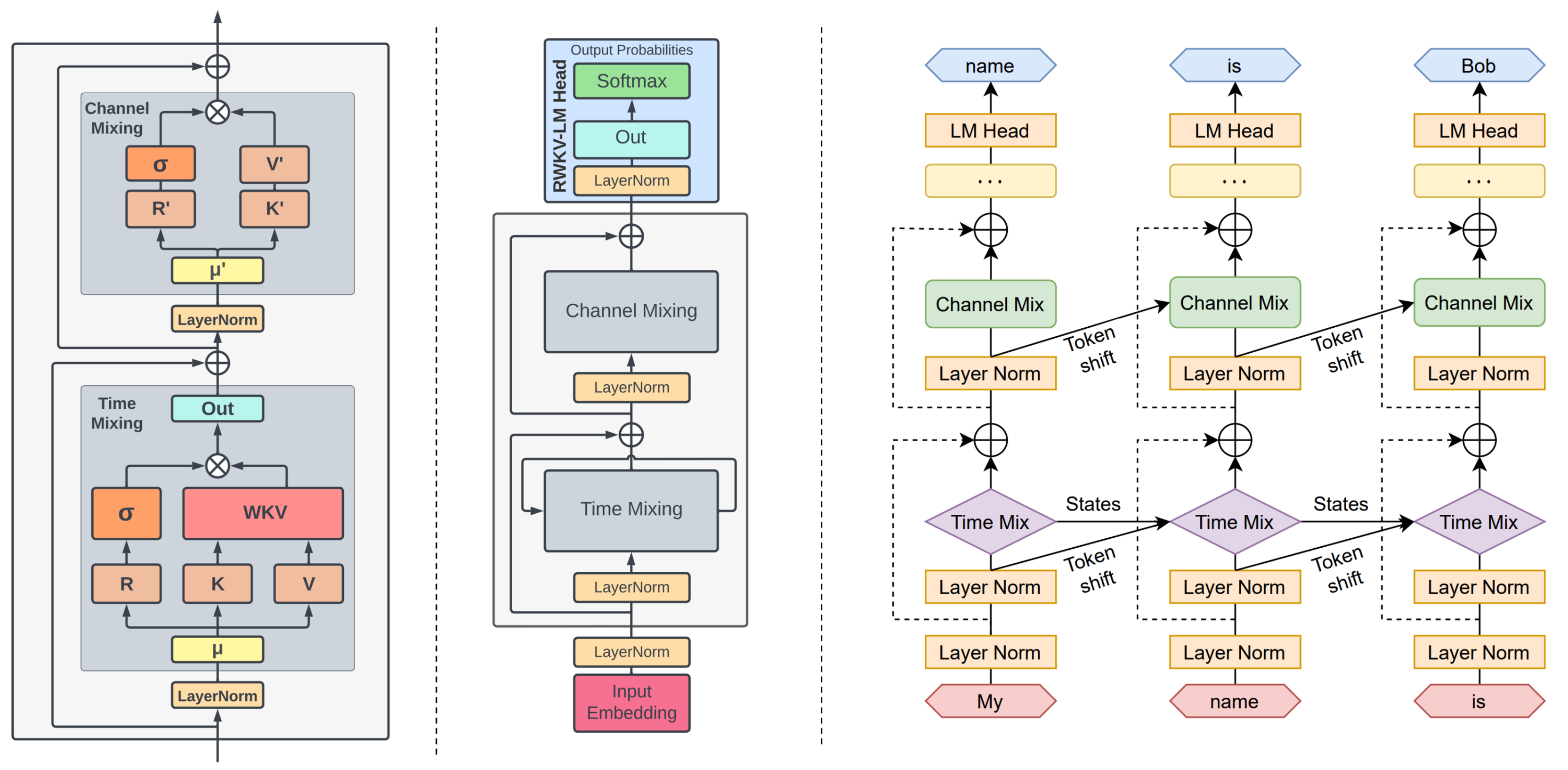}
\caption{The elements(left) block(middle) and architecture(right) in RWKV~\cite{peng2023rwkv}.}
\end{figure}

Although the Transformer is the dominant architecture in current large-scale language models, models like RWKV~\cite{peng2023rwkv} and Mamba~\cite{gu2023mamba} have emerged as popular solutions for achieving heightened efficiency and processing lengthy texts. These innovative models have demonstrated attributes similar to transformers, including the ability to handle long-range dependencies and parallel processing.RWKV model leverages a linear attention mechanism, enabling us to formulate the model as either a Transformer or a Recurrent Neural Network (RNN). This approach parallelizes computations during training and maintains constant computational and memory complexity during inference.

State Space Models (SSMs)~\cite{gu2021efficiently} can be formulated as a type of RNN for efficient autoregressive inference and have emerged as a promising alternative to attention mechanisms, offering near-linear computational complexity compared to the quadratic complexity of attention. SSMs are formulated as x'(t) = Ax(t) + Bu(t), y(t) = Cx(t) + Du(t), mapping a single-dimension input signal u(t) to an N-dimension latent state x(t) before projecting it to a single-dimension output signal y(t), with A, B, C, and D being parameters learned by gradient descent~\cite{gu2021efficiently}.
Several techniques have been proposed to enhance SSMs, such as the Structured State Space sequence model (S4)~\cite{gu2021efficiently}, which refines SSMs by conditioning matrix A with a low-rank correction, and the Diagonal State Space (DSS) model~\cite{gupta2022diagonal}, which proposes fully diagonal parameterization of state spaces for greater efficiency. H3 stacks two SSMs to interact with their output and input projection, bridging the gap between SSMs and attention while adapting to modern hardware.
Mamba~\cite{gu2023mamba}, a selective state space model, has been introduced as a strong competitor to the Transformer architecture in large language models. Mamba incorporates a selection mechanism to eliminate irrelevant data and develops a hardware-aware parallel algorithm for recurrent operation. This results in competitive performance compared to LLMs of the same capacity, with faster inference speeds that scale linearly with time and constant memory usage.
In conclusion, State Space Models offer significant potential as an alternative to attention mechanisms by providing near-linear computational complexity and effectively capturing long-range dependencies. With continuous advancements and refinements, SSMs are poised to become an influential approach in the field of deep learning and sequence processing.

\subsection{Fine-Tuning}
\label{subsec:ft}
Fine-tuning, as the primary stage for adapting LLMs to downstream tasks and training MLLLMs to follow visual instructions, plays a crucial role in enhancing the efficiency of LLMs.

\paragraph{Parameter-Efficient Fine-Tuning}
Parameter-Efficient Fine-Tuning (PEFT) is an approach that aims to achieve high performance with fewer parameters in Large Language Models (LLMs). Techniques such as adapter-based tuning and low-rank adaptation provide effective solutions to mitigate the computational and memory challenges associated with fine-tuning LLMs while maintaining their expressiveness and generalization capabilities.
Adapter-based tuning introduces lightweight adapter modules into the pre-trained model's architecture. These adapter modules, typically composed of feed-forward neural networks with a small number of parameters, are inserted between the layers of the original model. During fine-tuning, only the adapter parameters are updated, while the pre-trained model's parameters remain fixed. This method significantly reduces the number of trainable parameters, leading to faster training and inference times without compromising the model's performance. LLM-Adapters~\cite{hu2023llm} presents a framework for integrating various adapters into large language models, enabling parameter-efficient fine-tuning for diverse tasks. This framework encompasses state-of-the-art openly accessible large language models and a wide range of widely-used adapters.(IA)$^\text{3}$~\cite{liu2020tfew} introduces a novel Parameter-Efficient Fine-Tuning method, Infused Adapters by Inhibiting and Amplifying Inner Activations, which learns vectors to weight model parameters through multiplication with activations, enabling robust few-shot performance and task mixing within batches during inference without manual model structure adjustments.
Low-rank adaptation~\cite{hu2022lora} employs matrix factorization techniques to reduce the number of parameters in the model. By decomposing the original weight matrices into lower-rank matrices, low-rank adaptation captures the most significant components of the model's representations while discarding less important information. This results in a more compact model with a reduced number of parameters, which can be fine-tuned more efficiently.In LoRA-FA~\cite{zhang2023lora}, a variant of LoRA, the first low-rank matrix is frozen after initialization and used as a random projection, while the other is trained. This leads to a reduction in the number of parameters by half, while maintaining a performance comparable to the conventional LoRA technique.DyLoRa~\cite{valipour2022dylora} introduces a dynamic low-rank adaptation technique that enables the training of LoRA blocks for a range of ranks instead of a single rank, which is achieved by sorting the representations learned by the adapter modules during training across different ranks.

\paragraph{Full-Parameter fine-tuning}Full-parameter fine-tuning is an approach in which all the parameters of a pre-trained model are updated during the fine-tuning process. This method aims to achieve optimal performance on a specific downstream task by leveraging the entire capacity of the pre-trained model. While full-parameter fine-tuning often leads to state-of-the-art results and improved task-specific performance, it comes with higher resource requirements in terms of computational power and memory consumption. In an effort to lessen the burden associated with training, numerous studies have concentrated on enhancing memory efficiency during full-parameter fine-tuning. This strategic approach has effectively diminished the obstacles that once hindered progress in this field of research. LOMO~\cite{lv2023full} introduces a Low-Memory Optimization technique derived from Stochastic Gradient Descent (SGD) to reduce memory consumption. Typically, the ADAM optimizer is employed; however, the optimizer states in this approach occupy a significant amount of memory. By utilizing the modified SGD-based LOMO, memory usage can be reduced. While SGD itself faces three challenges, these issues tend to resolve themselves during model fine-tuning. The specific modification involves updating the parameters within the gradient computation rather than after an entire layer. MeZO\cite{malladi2023mezo} proposes an optimizer that computes gradients using merely two forward passes, enabling the fine-tuning of LLMs with a memory footprint equivalent to that of inference. With a GPU memory requirement of 55GB, it allows for the comprehensive fine-tuning of a 30B parameter model.

\section{Training}
\label{sec:train}
\begin{figure}[!t]
\centering
\includegraphics[width=\linewidth]{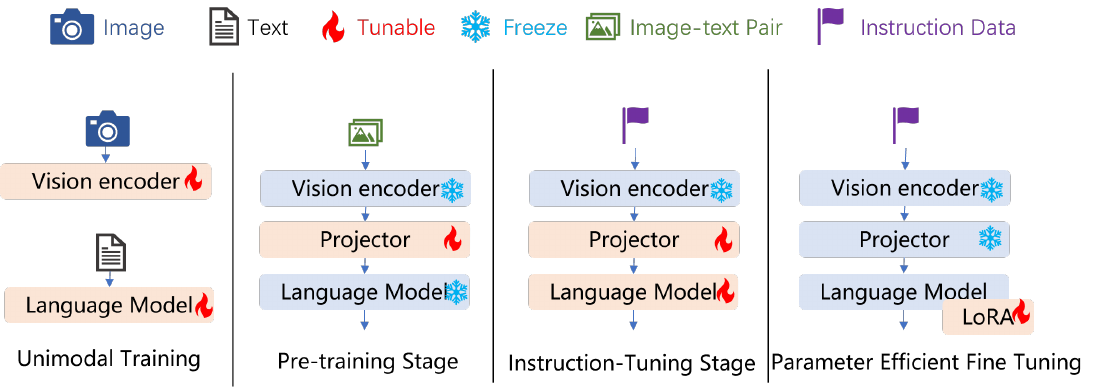}
\caption{Training stages of efficient MLLMs.}
\label{fig_train}
\end{figure}

The training process of efficient MLLMs is a critical aspect that determines their performance on downstream tasks and their ability to handle diverse modalities. In this section, we provide an overview of various training methodologies, including pre-training, instruction-tuning, diverse training steps, and parameter-efficient transfer learning strategies. These approaches aim to optimize the alignment between different modalities, fine-tune the models on specific tasks, and minimize the computational and parameter costs associated with the transfer learning process. 
Figure.\ref{fig_train} presents a schematic representation of the different training stages involved in the development of efficient MLLMs. In the following subsections, we delve deeper into each of these aspects and discuss their significance in the context of efficient MLLMs.

\subsection{Pre-Training}
\label{subsec:ptt}
In the pre-training stage, the primary focus is on aligning different modalities in the embedding space, enabling the language model to accept inputs from various modalities. This phase of training mainly involves large-scale text-paired data, predominantly in the form of image-caption pairs. An image-caption pair $(X, Y)$ is typically expanded into a single-turn conversation $(X_{instruct}, X_a)$, where $X_{instruct}$ contains an image $X_v$ and a randomly sampled question $X_q$ from a set of instructions asking the assistant to briefly describe the image, and $X_a$ is the original image description. Given such a conversation, the model is trained to autoregressively predict the image description. Consequently, we can compute the probability of predicting $X_a$ conditioned by $X_v$ and optimize it using a standard cross-entropy loss function:
\begin{equation}
    \mathop{\max}_{\theta} \sum_{i=1}^L \log p_\theta(x_i|X_v,X_{instruct},X_{a,<i}),
\end{equation}
where $L$ is the length of $X_a$ and $\theta$ denotes the trainable parameters. In order to better align different modalities of knowledge and avoid catastrophic forgetting during the pre-training stage, $\theta$ typically includes only a learnable modality interface, \textit{i.e.}, a vision-language projector. 

\paragraph{Which part to unfreeze?}Considering that only training the connector may not well align the vision and text information when using SLMs, TinyLlava\cite{zhou2024tinyllava} also opt to partially freeze pre-trained modules (i.e. vision encoder and SLM) to activate more parameters for learning alignment. VILA\cite{lin2023vila} reveals that updating the base LLM throughout the pre-training stage is essential to inheriting some of the appealing LLM properties like in-context learning. ShareGPT4V\cite{chen2023sharegpt4v} found that unfreezing more parameters, particularly in the latter half of the vision encoder's layers, proves beneficial when learning larger and more diverse datasets, showing the choice of training recipe is closely related to the quality of the data.

\paragraph{Multi-stage pre-training}To maximize compute efficiency, Idefics2~\cite{laurençon2024idefics2} decomposes the pre-training in two stages. In the first stage, it limits the max image resolution to 384 pixels and use a large global batch size. In the second stage, PDF documents are introduced to increase image resolution to a maximum of 980 pixels for the text to be legible.

\subsection{Instruction-Tuning}
\label{subsec:itt}
Instruction-tuning (IT) is a crucial aspect of efficient MLLMs, which aims to fine-tune the models on specific tasks by leveraging task-specific instructions. This approach is built upon the concept that MLLMs can understand and follow instructions provided in natural language, thereby enhancing their performance on the target task. The benefits of IT in efficient MLLMs are manifold. Firstly, it enables the model to adapt to a wide range of tasks with minimal changes to its architecture or training data. This makes it a flexible and efficient approach for fine-tuning on diverse tasks. Secondly, IT allows for better generalization, as the model learns to follow instructions and apply its knowledge to new and unseen tasks. 

The IT stage is typically conducted within the paradigm of Supervised Fine-Tuning (SFT). SFT datasets are often derived from a portion of the pre-training data, which is transformed into an instruction-based format, presented in the form of single-turn or multi-turn dialogue structures. Given an image $X_v$ and its caption, a conversation data $(X_q^1,X_a^1,\ldots,X_q^T,X_a^T)$ can be generated, where T is the total number of turns. 
Typically, we can organize the data into a sequence of instructions and responses following \cite{liu2023llava}, where the instruction $X_{instruct}^t$ at the $t$-th turn as:
\begin{equation}
X_{instruct}^t=\left\{
\begin{array}{lr}
\text{Randomly choose }[X_q^1,X_v]\ or\ [X_v,X_q^1],            & {\text{the first turn } t = 1}\\
X_q^t.           & {\text{the remaining turns } t > 1}
\end{array} \right. 
\end{equation}
With this multimodal instruction-following sequence, IT can be performed by using the same auto-regressive training objective as that of the pre-training stage. A prevalent strategy involves maintaining the visual encoder weights in a fixed state while continuing to update the pre-trained weights of both the projector and the SLM during the IT process. 

\paragraph{Efficient IT}Current IT solutions are prohibitively expensive, requiring optimization of a large number of parameters and additional large-scale training. LaVIN~\cite{luo2024lavin} introduces an innovative and cost-effective solution for efficient instruction tuning of MLLMs. The Mixture-of-Modality Adaptation (MMA) in LaVIN uses lightweight modules to bridge the gap between LLMs and VL tasks. This also facilitates the joint optimization of vision and language models. The actual cost of implementing LaVIN is remarkably low, for instance, it only requires 1.4 training hours with 3.8M trainable parameters. HyperLLaVA~\cite{zhang2024hyperllava} studies the under-explored dynamic tuning strategy for MLLMs and leverages the visual and language-guided dynamic tuning for the projector and LLM in two-stage training.

\begin{figure}[!t]
\centering
\includegraphics[width=0.95\linewidth]{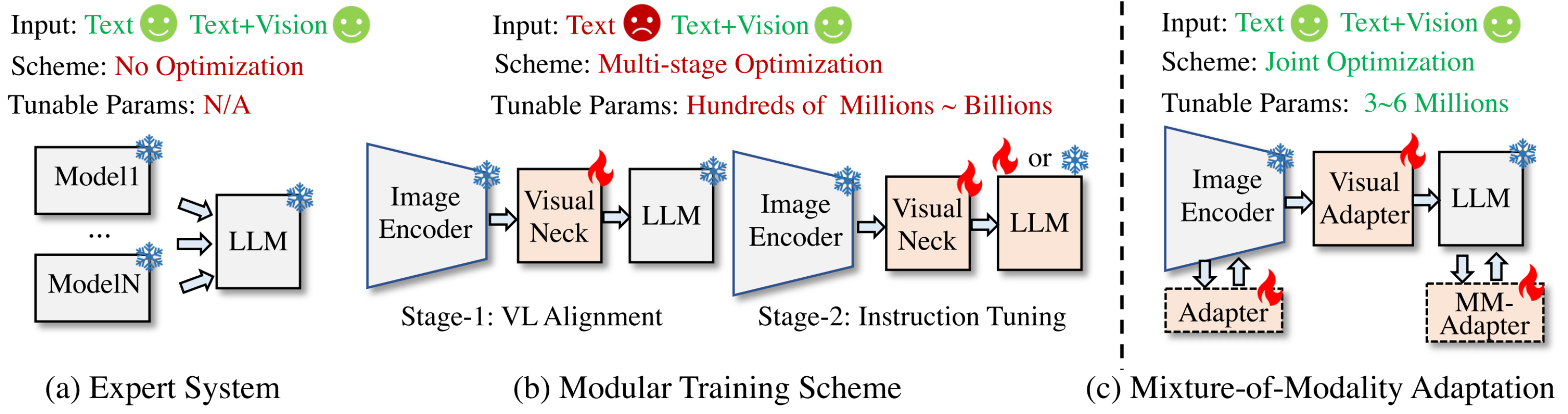}
\caption{Comparison of different multimodal adaptation schemes for LLMs in \textbf{LaVIN}~\cite{luo2024lavin}.}
\end{figure}

\subsection{Diverse Training Steps}
\label{subsec:div}
The traditional two-stage strategy, which demands the manual assignment of various adjustable parameters and dataset combinations to different training stages, can be a laborious task. To mitigate this, SPHINX-X\cite{gao2024sphinx} devises a single-stage, all-encompassing training pipeline that impartially treats all gathered datasets and consistently converts them into multi-modal, multi-turn dialogue formats. Throughout this unified training phase, all parameters except vision ecnoders within SPHINX-X are activated. Cobra\cite{zhao2024cobra} also argues that the initial phase of pre-alignment may not be requisite, with the model remaining underfitted even post-finetuning. Consequently, it discards the pre-alignment stage, opting instead to directly finetune the entire SLM backbone along with the projector.
TinyGPT-V\cite{yuan2023tinygpt-v} training process consists of four stages: an initial pre-training stage for vision-language understanding, a second stage for refining image modality processing, a third stage for human-like learning through fine-tuning, and a fourth stage for multi-task learning to enhance its conversational abilities as a chatbot.

\subsection{Parameter Efficient Transfer Learning}
\label{subsec:pe}
Several studies adopt Parameter-Efficient Fine-Tuning (PEFT) techniques for transfer learning, like LoRA~\cite{hu2022lora}, to safeguard against the loss of pre-trained knowledge. Efficient Attention Skipping (EAS) module\cite{wu2024EAS} proposes a novel parameter and computation-efficient tuning method for MLLMs to retain the high performance and reduce both parameter and computation expenditures on downstream tasks. MemVP~\cite{jie2024MemVP} argues that this transfer learning paradigm still exhibits inefficiency since it significantly increases the input length of the language models. Visual prompts in MemVP are concatenated with the weights of Feed Forward Networks for visual knowledge injection to reduce the training time and inference latency of the finetuned MLLMs and surpass the performance of previous PEFT methods.

\section{Data and Benchmarks}
\label{datasets}
In this section, we provide an overview of the data and benchmarks used for training and evaluating efficient MLLMs. We discuss the significance of pre-training data, instruction-tuning data, and the benchmarks employed to assess the performance of these models. The discussion highlights the importance of diverse and high-quality datasets in achieving robust and accurate MLLMs, as well as the various strategies employed to generate and refine these datasets. Furthermore, we present a comprehensive comparison of MLLM performance across established benchmarks, emphasizing the need for a thorough evaluation to ensure the effectiveness of these models in real-world applications.

\subsection{Pre-Training Data}
\label{subsec:pt}
Pre-training data primarily serve two critical objectives: (1) promoting the integration of various modalities and (2) conveying comprehensive knowledge. Large-scale image-caption pair datasets naturally fulfill these requirements. Firstly, they predominantly originate from the internet, providing an extensive data volume with a broad knowledge coverage. Secondly, the direct alignment between the two modalities is beneficial for training modality projectors. However, captions in such corpora are often brief and contain noise, which can be refined and filtered using automated methods, such as employing the CLIP [13] model to eliminate image-text pairs with low similarity scores.
A summary of frequently used pre-training datasets can be found in Figure\ref{pt_6}.

\begin{table}[ht]
\centering
\resizebox{\textwidth}{!}{%
\begin{tabular}{lcccccc}
\toprule
\textbf{Dataset Name}   & \textbf{X Modality} & \textbf{\#.X} & \textbf{\#.T} & \textbf{\#.X-T}    & \textbf{Representative Publications}                                                                              \\ \midrule
CC3M~\cite{sharma2018conceptual}                    & Image               & 3.3M          & 3.3M          & 3.3M               & TinyGPT-V\cite{yuan2023tinygpt-v},MM1\cite{mckinzie2024mm1}                                                                                     \\ \arrayrulecolor{gray!50} \midrule
CC12M~\cite{changpinyo2021conceptual}                   & Image               & 12.4M         & 12.4M         & 12.4M              & MM1\cite{mckinzie2024mm1}                                                                                               \\ \midrule 
SBU~\cite{ordonez2011im2text}                     & Image               & 1M            & 1M            & 1M                 & TinyGPT-V\cite{yuan2023tinygpt-v}                                                                                         \\ \midrule
LAION-5B~\cite{schuhmann2022laion}                & Image               & 5.9B          & 5.9B          & 5.9B               & TinyGPT-V\cite{yuan2023tinygpt-v}                                                                                         \\\midrule
LAION-COCO\cite{schuhmann2021laion}              & Image               & 600M          & 600M          & 600M               & Vary-toy~\cite{wei2024vary-toy}                                                                                          \\\midrule
COYO \cite{kakaobrain2022coyo-700m}                  & Image               & 747M          & 747M          & 747M               & MM1\cite{mckinzie2024mm1}                                                                                               \\\midrule
COCO Caption\cite{chen2015microsoft}            & Image               & 164K          & 1M            & 1M                 & Vary-toy~\cite{wei2024vary-toy}                                                                                          \\\midrule
CC595k \cite{liu2023llava}                 & Image               & 595K          & 595K          & 595K               & \begin{tabular}[c]{@{}c@{}}MobileVLM~\cite{chu2023mobilevlmv1},LLaVA-Phi~\cite{zhu2024llava-phi},\\ LLaVA-Gemma~\cite{hinck2024llava-gemma},Mini-Gemini~\cite{li2024mini-gemini}\end{tabular}            \\\midrule
RefCOCO\cite{kazemzadeh2014referitgame}                & Image               & 20K           & 142K          & 142K               & Vary-toy~\cite{wei2024vary-toy}                                                                                          \\\midrule
DocVQA\cite{mathew2021docvqa}                  & Image               & 12K           & 50K           & 50K                & Vary-toy~\cite{wei2024vary-toy}                                                                                          \\\midrule
LLava-1.5-PT\cite{liu2023llava1.5}            & Image               & 558K          & 558K          & 558K               & \begin{tabular}[c]{@{}c@{}}Imp-v1~\cite{imp2024},MoE-LLaVA~\cite{lin2024moe-llava},\\ Vary-toy~\cite{wei2024vary-toy},Mipha~\cite{zhu2024mipha},\\ VL-Mamba~\cite{qiao2024vlmamba},Tiny-LLaVA~\cite{zhou2024tinyllava}\end{tabular} \\\midrule
ShareGPT4V-PT \cite{chen2023sharegpt4v}          & Image               & 1246K         & 1246K         & 1246K              & Tiny-LLava~\cite{zhou2024tinyllava},MobileVLM V2~\cite{chu2024mobilevlmv2}                                                                           \\\midrule
ShareGPT4   \cite{chen2023sharegpt4v}        & Image               & 100K          & 100K          & 100K               & ALLaVA~\cite{chen2024allava}                                                                                            \\\midrule
Bunny-pretrain-LAION-2M\cite{he2024bunny} & Image               & 2M            & 2M            & 2M                 & Bunny~\cite{he2024bunny}                                                                                             \\\midrule
ALLaVA-Caption-4V \cite{chen2024allava}      & Image               & 715K          & 715K          & 715K               & Mini-Gemini~\cite{li2024mini-gemini}, ALLaVA\cite{chen2024allava}                                                                               \\\midrule
MMC4 (Interleaved) \cite{zhu2024multimodal}     & Image               & 571M          & 43B           & 101.2M (Instances) & DeepSeek-VL~\cite{lu2024deepseekvl}                                                                                       \\\midrule \arrayrulecolor{black}
Obelics (Interleaved)\cite{laurenccon2024obelics}   & Image               & 353M          & 115M          & 141M (Instances)   & MM1\cite{mckinzie2024mm1}                                                \\ \bottomrule                                                  
\end{tabular}%
}
\caption{The statistics for common MLLM PT datasets.\#.X represents the quantity of X, \#.T represents the quantity of Text, and \#.X-T represents the quantity of X-Text pairs, where X can be Image, Video, or Audio.
}
\label{pt_6}
\end{table}

A growing number of studies have investigated the production of high-quality fine-grained pre-trained data by leveraging powerful MLLMs like GPT-4V. These datasets typically offer more detailed and accurate image descriptions compared to their coarse-grained counterparts, enabling a closer alignment of image and text modalities. However, this method often requires the use of commercial MLLMs, leading to increased costs and a smaller data volume. ShareGPT4V\cite{chen2023sharegpt4v} addresses this issue by first training a captioner on 100K GPT-4V-generated data and then expanding the dataset to 1.2M using the pretrained captioner. Moreover,VILA's\cite{lin2023vila} findings indicate that incorporating interleaved pre-training data proves advantageous, while solely relying on image-text pairs is suboptimal in achieving the desired outcomes.
\subsection{Instruction-Tuning Data}
\label{subsec:it}
Instruction tuning(IT) is a crucial step in refining efficient MLLMs' capacity to accurately interpret user instructions and effectively carry out the desired tasks. This procedure bears a strong connection to the concept of multi-task prompting.

 A summary of frequently used pre-training datasets can be found in Table.\ref{it_6}. High-quality IT data can be derived from task-specific datasets. For instance, consider a sample from VQA datasets where the input includes an image and a natural language question, and the output is the text-based answer to the question based on the image. This could easily form the multimodal input and response of the instruction sample. The instructions, or task descriptions, can be obtained either through manual creation or semi-automatic generation with the help of GPT. In addition to utilizing publicly available task-specific datasets, SPHINX-X\cite{gao2024sphinx} assembles a dataset focused on OCR from a wide range of PDF data sourced from the internet. Specifically, it begins by gathering a large-scale PDF dataset from the web. It then obtains the rendering results of each page in the PDF file, while simultaneously saving all text annotations along with their respective bounding boxes. Ultimately, these elements are converted into a unified question-answering format.

While multi-task datasets provide an abundant source of data, they may not always be suitable for complex real-world situations, such as engaging in multi-turn conversations. To address this challenge, some research has explored the use of self-instruction by leveraging LLMs to generate text-based or multimodal instruction-following data from a limited number of hand-annotated samples. 
SPHINX-X\cite{gao2024sphinx} assembles a rich multi-domain dataset with fine-grained correspondence between images and texts.It gathers images from diverse sources and then employs annotations to apply various markers onto the original images. 
By prompting GPT-4V with these marked images and tailored domain-specific guidelines,
the system generates captions that offer an image overview, regional details and object relationships insight. During the training process, SPHINX-X utilizes the unaltered images rather than the marked ones.
ALLaVA\cite{chen2024allava} propose to distill a caption and a QA pair for an image within a single session. Specifically, it prompts GPT-4V with an image, and ask it to first generate a fine-grained caption then a VQA pair. 

Additionally, excluding multimodal instructional data, conversations solely based on language between users and assistants can significantly contribute to enhancing a model's conversational expertise and responsiveness to directives.For example, VILA's\cite{lin2023vila} research demonstrates that integrating text-only instructional data with image-text data during the fine-tuning process not only mitigates the decline in performance for text-only tasks but also enhances the accuracy of MLLM-related tasks.

\begin{table}[ht]
\centering
\resizebox{\textwidth}{!}{%
\begin{tabular}{lccccccc}
\toprule
\textbf{Dataset Name} & \textbf{Type} & \textbf{I→O} & \textbf{Source}                           & \textbf{Method} & \textbf{\#.Instance} & \textbf{Representative Publications}                                                                             \\ \midrule
LLaVA’s IT~\cite{liu2023llava}            & SFT           & I+T→T        & MS-COCO\cite{lin2014microsoft}                                   & Auto.           & 150K                 & \makecell[c]{MobileVLM~\cite{chu2023mobilevlmv1}, LLaVA-Phi~\cite{zhu2024llava-phi},\\ Mini-Gemini~\cite{li2024mini-gemini}, Vary-toy~\cite{wei2024vary-toy}, \\TinyGPT-V~\cite{yuan2023tinygpt-v}, Imp-v1~\cite{imp2024}, \\ALLaVA~\cite{chen2024allava},SPHINX-X~\cite{gao2024sphinx},\\LLaVA-Gemma~\cite{hinck2024llava-gemma},MM1~\cite{mckinzie2024mm1}} \\ \arrayrulecolor{gray!50} \midrule
ShareGPT4V’s IT\cite{chen2023sharegpt4v}       & SFT           & I+T→T        & \makecell[c]{LCS, COCO~\cite{lin2014microsoft}, SAM~\cite{kirillov2023segment}, \\TextCaps\cite{sidorov2020textcaps}, WikiArt~\cite{saleh2015large}}         & Auto.+Manu.     & –                    & \makecell[c]{Tiny-LLaVA~\cite{zhou2024tinyllava}, Mini-Gemini~\cite{li2024mini-gemini}, \\ MM1~\cite{mckinzie2024mm1}, DeepSeek-VL~\cite{lu2024deepseekvl},\\SPHINX-X~\cite{gao2024sphinx}}                                               \\ \midrule
LLaVA-1.5's IT~\cite{liu2023llava1.5}        & SFT           & I+T→T        & \makecell[c]{LLaVA\cite{liu2023llava},Visual Genome\cite{krishna2017visual},\\ VQAv2\cite{goyal2017vqav2}, ShareGPT\cite{ShareGPT2023}, \\A-OKVQA\cite{schwenk2022okvqa},TextCaps\cite{sidorov2020textcaps}, \\GQA\cite{hudson2019gqa}, OKVQA\cite{marino2019ok},\\OCRVQA, RefCOCO\cite{kazemzadeh2014referitgame, mao2016generation}}   & Auto.+Manu.     & 665K                 & \makecell[c]{Tiny-LLava~\cite{zhou2024tinyllava}, VL-Mamba~\cite{qiao2024vlmamba}, \\Cobra~\cite{zhao2024cobra}, LLaVA-Gemma~\cite{hinck2024llava-gemma}, \\Mipha~\cite{zhu2024mipha}, MoE-LLaVA~\cite{lin2024moe-llava}}                                        \\ \midrule
LRV-Instruct~\cite{liu2023mitigating}          & SFT           & I+T→T        & Visual Genome\cite{krishna2017visual}                             & Auto.           & 300K                 & MoE-LLaVA~\cite{lin2024moe-llava}, Cobra\cite{zhao2024cobra}                                                                                  \\ \midrule
LVIS-INSTRUCT-4V\cite{wang2023see}      & SFT           & I+T→T        & LVIS~\cite{gupta2019lvis}       & Auto.           & 220K                 & \makecell[c]{MoE-LLaVA~\cite{lin2024moe-llava}, SPHINX-X~\cite{gao2024sphinx},\\DeepSeek-VL~\cite{lu2024deepseekvl}, Cobra\cite{zhao2024cobra}}                                                           \\ \midrule
LAION GPT4V\cite{laion2023gpt4v}           & SFT           & I+T→T        & LAION~\cite{schuhmann2021laion}                                     & Auto.           & 12.4k                & \makecell[c]{Mini-Gemini~\cite{li2024mini-gemini}, SPHINX-X~\cite{gao2024sphinx},\\ DeepSeek-VL~\cite{lu2024deepseekvl}}                                                                 \\ \midrule
MiniGPT-4’s IT \cite{zhu2023minigpt4}       & SFT           & I+T→T        & CC3M~\cite{sharma2018conceptual}, CC12M~\cite{changpinyo2021conceptual}                               & Auto.           & 5K                   & TinyGPT-V~\cite{yuan2023tinygpt-v},                                                                                         \\ \midrule
SVIT  \cite{zhao2023svit}                & SFT           & I+T→T        & \makecell[c]{MS-COCO\cite{lin2014microsoft}, \\Visual Genome\cite{krishna2017visual}}                   & Auto.           & 3.2M                 & MoE-LLaVA~\cite{lin2024moe-llava}                                                                                         \\ \midrule
Bunny-695K~\cite{he2024bunny}            & SFT           & I+T→T        & \makecell[c]{SVIT-mix-665K\cite{zhao2023svit}, \\WizardLM-evol-instruct-70K\cite{xu2023wizardlm}} & Auto.     & 695K                 & Bunny~\cite{he2024bunny}                                                                                             \\ \midrule \arrayrulecolor{black}
GQA\cite{hudson2019gqa}                   & SFT             & I+T→T            & Visual Genome\cite{krishna2017visual}                                         & Auto.               &  22M                    & \makecell[c]{MM1~\cite{mckinzie2024mm1}, SPHINX-X~\cite{gao2024sphinx}, \\LLaVA-Gemma~\cite{hinck2024llava-gemma}}                                                        \\ \bottomrule                   
\end{tabular}%
}
\caption{The statistics for common MLLM IT datasets. I→O:Input to Output Modalities,T:Text.}
\label{it_6}
\end{table}

\subsection{Benchmarks}
\label{subsec:bench}
\begin{table}[ht]
\centering
\setlength{\tabcolsep}{2pt}
\resizebox{\textwidth}{!}{%
\begin{tabular}{lc|cccccccccccccc}
\toprule
\textbf{Model}                                     & \textbf{LLM Backbone}       & \textbf{VQA$^\text{v2}$} & \textbf{GQA}  & \textbf{SQA$^\text{I}$} & \textbf{VQA$^\text{T}$} & \textbf{VizWiz} & \textbf{MMMU}      & \textbf{MathV} & \textbf{MME$^\text{P}$}   & \textbf{MME$^\text{C}$}   & \textbf{MMB}  & \textbf{SEED}      & \textbf{POPE} & \textbf{LLAVA$^\text{W}$} & \textbf{MM-Vet} \\ \midrule
Flamingo~\cite{alayrac2022flamingo}                & Chinchilla-7B      & -     & -    & -    & -    & 28.8   & -         & -     & -      & -      & -    & -         & -    & -      & -      \\
BLIP-2~\cite{li2023blip2}                  & Flan-T5XXL(13B)    & 65.0  & 44.7 & 61.0 & 42.5 & 19.6   & -         & -     & 1293.8 & 290.0  & -    & -/46.4    & 85.3 & 38.1   & 22.4   \\
LLaVA~\cite{liu2023llava}                 & Vicuna-13B         & -     & 41.3 & -    & 38.9 & -      & -         & -     & -      & -      & -    & -         & -    & -      & -      \\
MiniGPT-4~\cite{zhu2023minigpt4}          & Vicuna-13B         & -     & 30.8 & -    & 19.4 & -      & -         & -     & -      & -      & -    & -         & -    & -      & -      \\
InstructBLIP~\cite{dai2024instructblip}   & Vicuna-13B         & -     & 49.5 & 63.1 & 50.7 & 33.4   & -         & -     & 1212.8 & 291.8  & -    & -         & 78.9 & 58.2   & 25.6   \\
Qwen-VL-Chat~\cite{bai2023qwen-vl}        & Qwen-7B            & 78.2$^*$  & 57.5$^*$ & 68.2 & 61.5 & 38.9   & 35.9/32.9 & -     & 1487.5 & 360.7  & 60.6 & -/58.2    & -    & -      & -      \\
LLaVA-1.5~\cite{liu2023llava1.5}          & Vicuna-1.5-13B     & 80.0$^*$  & 63.3$^*$ & \color{blue}\textbf{71.6} & 61.3 & 53.6   & -         & -     & 1531.3 & 295.4  & 67.7 & -/68.2    & 85.9 & \color{blue}\textbf{70.7}   & 35.4   \\
MiniGPT-v2-Chat~\cite{chen2023minigpt-v2} & LLaMA-2-Chat-7B    & -     & 58.8 & -    & 52.3 & 42.4   & -         & -     & -      & -      & -    & -         & -    & -      & -      \\
InternVL-Chat~\cite{chen2023internvl}                             & Vicuna-13B         & 81.2$^*$  & \textcolor{blue}{\textbf{66.6}}$^*$ & -    & 61.5 & \color{blue}58.5   & -         & -     & \color{blue}\textbf{1586.4} & -      & -    & -         & \color{blue}\textbf{87.6} & -      & -      \\
Emu2-Chat~\cite{sun2023generative}                                 & LLaMA-33B          & \textcolor{blue}{\textbf{84.9}}$^*$  & 65.1$^*$ & -    & 66.6$^*$    & 54.9   & -/34.1      & -     & -      & -      & -    & 62.8      & -    & -      & 48.5   \\
Gemini Pro~\cite{team2023gemini}                                & -                  & 71.2  & -    & -    & 74.6 & -      & 47.9/–    & 45.2  & -      & 436.79 & 73.6 & –/70.7    & -    & -      & 64.3   \\
Gemini Ultra~\cite{team2023gemini}                              & -                  & 77.8  & -    & -    & \color{blue}\textbf{82.3} & -      & \color{blue}\textbf{59.4/–}    & \color{blue}\textbf{53.0}    & -      & -      & -    & -         & -    & -      & -      \\
GPT4V~\cite{achiam2023gpt}                                     & -                  & 77.2  & -    & -    & 78.0 & -      & 56.8/55.7 & 49.9  & -      & \color{blue}\textbf{517.14} & \color{blue}\textbf{75.8} & \color{blue}\textbf{67.3/69.1} & -    & -      & \color{blue}\textbf{67.6}  \\ \midrule
MobileVLM~\cite{chu2023mobilevlmv1}       & MobileLLaMA (2.7B) & -     & 59.0$^*$ & 61.0 & 47.5 & -      & -         & -     & 1288.9 & -      & 59.6 & -         & 84.9 & -      & -      \\
LLaVA-Phi~\cite{zhu2024llava-phi}         & Phi-2 (2.7B)       & 71.4$^*$  & -    & 68.4 & 48.6 & 35.9   & -         & -     & 1335.1 & -      & 59.8 & -         & 85.0 & -      & 28.9   \\
Imp-v1~\cite{imp2024}                     & Phi-2 (2.7B)       & 79.5  & -    & 70.0 & 59.4 & -      & -         & -     & 1434.0 & -      & 66.5 & -         & \color{red}\textbf{88.0} & -      & 33.1   \\
TinyLLaVA~\cite{zhou2024tinyllava}        & Phi-2 (2.7B)       & 79.9$^*$  & 62.0$^*$ & 69.1 & 59.1 & -      & -         & -     & 1464.9 & -      & 66.9 & -         & 86.4 & 75.8   & 32.0   \\
Bunny~\cite{he2024bunny}                  & Phi-2 (2.7B)       & 79.8  & 62.5 & 70.9 & -    & -      & 38.2/33.0 & -     & 1488.8 & 289.3  & 68.6 & 62.5/-    & 86.8 & -      & -      \\
Gemini   Nano-2~\cite{team2023gemini}     & -                  & 67.5  & -    & -    & 65.9 & -      & 32.6/-    & 30.6  & -      & -      & -    & -         & -    & -      & -      \\
MobileVLM-v2~\cite{chu2024mobilevlmv2} & MobileLLaMA(2.7B)  & -     & 61.1 & 70.0 & 57.5 & -      & -         & -     & 1440.5 & -      & -    & -         & 84.7 & -      & -      \\
MoE-LLaVA~\cite{lin2024moe-llava}    & Phi-2 (2.7B)       & 79.9$^*$  & 62.6$^*$ & 70.3 & 57.0 & 43.7   & -         & -     & 1431.3 & -      & 68.0 & -         & 85.7 & -      & 35.9   \\
Cobra~\cite{zhao2024cobra}                & Mamba-2.8B  & 75.9  & 58.5 & -    & 46.0 & \color{red}\textbf{52.0}   & -         & -     & -      & -      & -    & -         & \color{red}\textbf{88.0} & -      & -      \\
Mini-Gemini~\cite{li2024mini-gemini}      & Gemma-2B           & -     & -    & -    & 56.2 & -      & 31.7/29.1 & 29.4  & 1341.0 & 312.0  & 59.8 & -         & -    & -      & 31.1   \\
Vary-toy~\cite{wei2024vary-toy}           & Qwen-1.8B          & -     & -    & -    & -    & -      & -         & -     & -      & -      & -    & -         & -    & -      & 29.0   \\
TinyGPT-V~\cite{yuan2023tinygpt-v}        & Phi-2 (2.7B)       & -     & 33.6 & -    & -    & 24.8   & -         & -     & -      & -      & -    & -         & -    & -      & -      \\
SPHINX-Tiny~\cite{gao2024sphinx}          & TinyLlama-1.1B     & -     & -    & -    & 57.8 & -      & -         & 26.4  & 1261.2 & 242.1  & 56.6 & 17.1/-    & 82.2 & 52.3   & 23.8   \\
ALLaVA-Longer~\cite{chen2024allava}       & Phi-2 (2.7B)                     & -        & 50.0 & -     & 50.3  & -      & 33.2/-    & -     & \multicolumn{2}{c}{1564.6$^\dagger$} & 64.6 & -         & -    & 71.7    & 35.5   \\
MM1-3B~\cite{mckinzie2024mm1}    & MM1-3B                  & \color{red}\textbf{82.5}  & -    & 76.1 & \color{red}\textbf{72.9} & -      & 38.6/35.7 & 32.6  & 1469.4 & 303.1  & 70.8 & 63.9/69.4 & 87.6 & \color{red}\textbf{76.8}   & \color{red}\textbf{42.2}   \\
LLaVA-Gemma~\cite{hinck2024llava-gemma}   & Gemma-2b-it        & 71.4  & 58.7 & -    & -    & -      & -         & -     & 1133.0 & 307.0  & -    & -         & 85.3 & -      & 19.1   \\
Mipha-3B~\cite{zhu2024mipha}              & Phi-2 (2.7B)       & 81.3$^*$  & \textcolor{red}{\textbf{63.9}}$^*$ & 70.9 & 56.6 & 47.5   & -         & -     & 1488.9 & 295.0  & 69.7 & -         & 86.7 & -      & 32.1   \\
VL-Mamba~\cite{qiao2024vlmamba}           & Mamba-2.8B     & 76.6  & 56.2 & 65.4 & 48.9 & -      & -         & -     & 1369.6 & -      & 57.0 & -         & 84.4 & -      & 32.6   \\ 
MiniCPM-V 2.0\cite{minicpm-v}                          & MiniCPM-2.4B       & -        & -    & -     & 74.1  & -      & 38.2/-    & \color{red}\textbf{38.7}  & \multicolumn{2}{c}{1808.6$^\dagger$} & 69.6 & -         & -    & -       & -   \\ 
DeepSeek-VL~\cite{lu2024deepseekvl}           & DeepSeek-LLM-1B     & -  & - & - & - & -      & 32.2/-         & 31.1     & - & -      & 64.6 & -/66.7         & 87.6 & -      & 36.8   \\ 
KarmaVLM\cite{karmavlm}           & Qwen1.5-0.5B     & -  & - & 53.86 & 45.25 & -      & -         & -     & - & -      & 55.8 & -        & - & 47.5      & -   \\ 
moondream2\cite{moondream}           & Phi-1.5(1.3B)     & 77.7  & 61.7 & - & 49.7 & -      & -         & -     & - & -      & - & -        & - & -      & -   \\ 
Bunny-v1.1-4B\cite{he2024bunny}           & Phi-3-Mini-4K     & 81.7  & 63.4 & \color{red}\textbf{76.3} & - & -      & \color{red}\textbf{40.2/38.8}         & -     & \color{red}\textbf{1503.9} & \color{red}\textbf{362.9}      & \color{red}\textbf{74.1} & \color{red}\textbf{64.6/71.7}        & 87.0 & -      & -   \\ \bottomrule
\end{tabular}%
}
\caption{Comparison of mainstream MLLMs and efficient MLLMs on 14 VL benchmarks. VQA$^\text{v2}$~\cite{goyal2017vqav2}; 
VQA$^\text{T}$: TextVQA~\cite{singh2019vqat}; GQA~\cite{hudson2019gqa}; SQA$^\text{I}$: ScienceQA-IMG~\cite{lu2022scienceQA}; VizWiz~\cite{gurari2018vizwiz}; MMMU~\cite{yue2023mmmu}; MathV: MathVista~\cite{lu2023mathvista}; MME$^\text{P/C}$: the Perception/Cognition split of MME~\cite{fu2024mme}; MMB: MMBench~\cite{liu2023mmbench}; SEED: SEED-Bench~\cite{li2023seedbench}; POPE~\cite{li2023pope}; LLaVA$^\text{W}$: LLaVA-Bench (In-the-Wild)~\cite{liu2023llava}; MM-Vet~\cite{yu2023mmvet}. The two numbers reported in MMMU denote the performance on the val and test split, respectively. The two numbers reported in SEED denote the performance on the whole SEED-Bench and the image part, respectively. $^\dagger$ denotes the combined points of two splits. $^*$ indicates that training images of the datasets are observed during training.The \textcolor{red}{\textbf{red}} denotes the highest result of efficient MLLMs, and the \textcolor{blue}{\textbf{blue}} denotes that of large-scale MLLMs.}
\label{benchmarktable}
\end{table}

With the aim of delivering an all-encompassing performance evaluation, we have assembled a table that demonstrates the effectiveness of 22 MLLMs across 14 well-established VL benchmarks, as depicted in Table.\ref{benchmarktable}. Additionally, for further reference, we have incorporated a comparison of results from 13 prominent and larger MLLMs.

\section{Applications}
\label{sec:app}
From the preceding analysis, it's clear that many efficient MLLM approaches evaluate their performances across a range of scenarios, like VQA, visual grounding, image segmentation, \textit{etc}. However, it's also crucial to explore these efficient architectures in well-established tasks to achieve their ultimate performance. Therefore, we have chosen to introduce several downstream tasks, such as medical analysis, document understanding, and video comprehension.

\subsection{Biomedical Analysis}
\label{subsec:bio}
Due to the high cost of annotating biomedical data, foundation models
are poised to become a new paradigm in biomedicine, achieving state-of-the-art results on many applications, including medical question answering~\cite{tu2024Med-PaLM} and medical image classification~\cite{azizi2021medical_cls}. Recently, multimodal generative AI has emerged as an exciting frontier in the biomedical domain, expanding the application scope from single-modality to multi-modality, such as VQA and radiology report generation. 

The mixture of Expert Tuning has effectively enhanced the performance of general MLLMs with fewer parameters, yet its application in resource-limited medical settings has not been fully explored. MoE-TinyMed~\cite{jiang2024moetinymed} is a model tailored for medical applications that significantly lower parameter demands. 
LLaVA-Rad~\cite{chaves2024llava-rad} is a state-of-the-art tool that demonstrates rapid performance on a single V100 GPU in private settings, making it highly applicable for real-world clinical scenarios. It employs a modular approach, integrating unimodal pre-trained models and emphasizing the training of lightweight adapters. As a result, LLaVA-Rad outperforms larger models such as GPT-4V and Med-PaLM in terms of standard metrics, showcasing its superior efficiency and effectiveness.

\subsection{Document Understanding}
\label{subsec:doc}
Documents or charts serve as a crucial source of information, offering an intuitive visualization of data in various forms. They have become an indispensable part of information dissemination, business decision-making, and academic research. However, current chart understanding models still face two primary limitations: (1) The considerable number of parameters makes training and deployment challenging. For instance, ChartLlama~\cite{han2023chartllama}, a 13-billion-parameter model, is difficult to deploy on a single consumer-grade GPU. (2) These models struggle with efficiently encoding high-resolution images, as vision transformers tend to produce lengthy feature sequences.

To address the challenges of fine-grained visual perception and visual information compression for document-oriented MLLMs. TinyChart~\cite{zhang2024tinychart} outperforms several 13B MLLMs with Program-of-Thoughts (PoT) learning and Visual Token Merging strategy while excelling in faster inference speed at the same time. TextHawk~\cite{yu2024texthawk} explores efficient fine-grained perception by designing four dedicated components to address challenges posed by document-oriented tasks. HRVDA~\cite{liu2024hrvda} and Monkey~\cite{li2024monkey} are also large multimodal models designed to address the challenges posed by high-resolution requirements in visual document understanding tasks.

\subsection{Video Comprehension}
\label{subsec:video}
Videos provide an impressively accurate representation of how humans continuously perceive the visual world. Intelligent video understanding is vital for a variety of real-world applications, including video category classification, video captioning, and video-text retrieval. Several works like videoChat~\cite{li2023videochat} and
Video-LLaMA~\cite{zhang2023videollama} are LLM-based large multimodal models for end-to-end chat-centric video comprehension. However, these methods can only take in a limited number of frames for short video understanding.

To address the computational challenges associated with processing long videos due to the excessive number of visual tokens, several approaches have been developed. mPLUG-video~\cite{xu2023mplug2} is designed for video understanding tasks and begins with a TimeSformer-based video encoder to extract features from sparsely sampled video frames effectively, followed by a visual abstractor module to reduce sequence length. Video-LLaVA~\cite{lin2023videollava} excels in various video understanding tasks by unifying visual representations of images and videos into a single language feature space before projection. This approach enables effective learning of multi-modal interactions with LanguageBind~\cite{zhu2023languagebind}. LLaMA-VID~\cite{li2023llamavid} addresses this issue by representing each frame with two distinct tokens, namely context token and content token. The context token encodes the overall image context based on user input, whereas the content token encapsulates visual cues in each frame. This dual-token strategy significantly reduces the overload of long videos while preserving critical information. Instead of trying to process more frames simultaneously like most existing work, MA-LMM~\cite{he2024malmm} proposes to process videos in an online manner and store past video information in a memory bank to reference historical video content for long-term analysis without exceeding LLMs’ context length constraints or GPU memory limits.

\section{Discussion and Conclusion}

\subsection{Limitations and Future work}
The development of efficient MLLMs is still in its nascent stage, and there is ample room for improvement. We summarize the current state of affairs as follows: 
\begin{itemize}
    \item At present, efficient MLLMs face challenges in processing extended-context multimodal information, and they are typically limited to accepting single images. This constrains the advancement of more sophisticated models capable of handling an increased number of multimodal tokens. Such models would be beneficial for applications like comprehending lengthy videos and analyzing extensive documents that incorporate a mix of images and text, creating more versatile and powerful systems.
    \item The predominant efficient MLLMs mainly support dual input modalities - images and texts, and a singular output modality - text. However, the tangible world encompasses a more extensive array of modalities. By expanding the scope of efficient MLLMs to accommodate a richer diversity of input modalities, and augmenting their generative capacities, we can significantly bolster their multifunctionality and widen their applicability.
    \item There are two principal pathways to fortify efficient MLLM models. Firstly, the incorporation of a more varied set of lightweight LLMs can render the design of MLLMs more adaptable, facilitating their customization to cater to a broad spectrum of requirements. Secondly, leveraging high-quality instruction tuning datasets can empower efficient MLLMs to better comprehend and implement a vast array of instructions, thereby amplifying their zero-shot learning capabilities.
    \item The development of embodied agents capable of deployment on edge devices represents a crucial application prospect for efficient MLLMs. An agent possessing specialized knowledge and the capability to interact with the real world has far-reaching implications, potentially revolutionizing fields such as robotics, automation, and artificial intelligence.
\end{itemize}

\subsection{Conclusion}

In this study, we take a deep dive into the realm of efficient MLLM literature, providing an all-encompassing view of its central themes, including foundational theories and their extensions. Our goal is to identify and highlight areas that require further research and suggest potential avenues for future studies. We aim to provide a comprehensive perspective on the current state of efficient MLLM, with the hope of inspiring additional research.
Given the dynamic nature of this field, it's possible that some recent developments may not be fully covered. To counter this, we've set up a dedicated website that uses crowdsourcing to keep up with the latest advancements. This platform is intended to serve as a continually updated source of information, promoting ongoing growth in the field.
Due to space constraints, we can't cover all technical details in depth but have provided brief overviews of the key contributions in the field. In the future, we plan to continuously update and enhance the information on our website, adding new insights as they come to light.

\newpage
{
    \small
    \bibliographystyle{unsrt}
    \bibliography{neurips_2023}

@String(ICLR = {Int. Conf. Learn. Represent.})

@String(AAAI = {AAAI})

@String(ICLR  = {ICLR})

@article{chu2023mobilevlmv1,
  title={Mobilevlm: A fast, reproducible and strong vision language assistant for mobile devices},
  author={Chu, Xiangxiang and Qiao, Limeng and Lin, Xinyang and Xu, Shuang and Yang, Yang and Hu, Yiming and Wei, Fei and Zhang, Xinyu and Zhang, Bo and Wei, Xiaolin and others},
  journal={arXiv preprint arXiv:2312.16886},
  year={2023}
}

@article{chu2024mobilevlmv2,
  title={MobileVLM V2: Faster and Stronger Baseline for Vision Language Model},
  author={Chu, Xiangxiang and Qiao, Limeng and Zhang, Xinyu and Xu, Shuang and Wei, Fei and Yang, Yang and Sun, Xiaofei and Hu, Yiming and Lin, Xinyang and Zhang, Bo and others},
  journal={arXiv preprint arXiv:2402.03766},
  year={2024}
}

@article{wei2024vary-toy,
  title={Small Language Model Meets with Reinforced Vision Vocabulary},
  author={Wei, Haoran and Kong, Lingyu and Chen, Jinyue and Zhao, Liang and Ge, Zheng and Yu, En and Sun, Jianjian and Han, Chunrui and Zhang, Xiangyu},
  journal={arXiv preprint arXiv:2401.12503},
  year={2024}
}

@article{he2024bunny,
  title={Efficient Multimodal Learning from Data-centric Perspective},
  author={He, Muyang and Liu, Yexin and Wu, Boya and Yuan, Jianhao and Wang, Yueze and Huang, Tiejun and Zhao, Bo},
  journal={arXiv preprint arXiv:2402.11530},
  year={2024}
}

@article{yuan2023tinygpt-v,
  title={Tinygpt-v: Efficient multimodal large language model via small backbones},
  author={Yuan, Zhengqing and Li, Zhaoxu and Sun, Lichao},
  journal={arXiv preprint arXiv:2312.16862},
  year={2023}
}

@article{hinck2024llava-gemma,
  title={LLaVA-Gemma: Accelerating Multimodal Foundation Models with a Compact Language Model},
  author={Hinck, Musashi and Olson, Matthew L and Cobbley, David and Tseng, Shao-Yen and Lal, Vasudev},
  journal={arXiv preprint arXiv:2404.01331},
  year={2024}
}

@article{zhu2024llava-phi,
  title={LLaVA-phi: Efficient Multi-Modal Assistant with Small Language Model},
  author={Zhu, Yichen and Zhu, Minjie and Liu, Ning and Ou, Zhicai and Mou, Xiaofeng and Tang, Jian},
  journal={arXiv preprint arXiv:2401.02330},
  year={2024}
}

@article{lin2024moe-llava,
  title={Moe-llava: Mixture of experts for large vision-language models},
  author={Lin, Bin and Tang, Zhenyu and Ye, Yang and Cui, Jiaxi and Zhu, Bin and Jin, Peng and Zhang, Junwu and Ning, Munan and Yuan, Li},
  journal={arXiv preprint arXiv:2401.15947},
  year={2024}
}

@article{zhao2024cobra,
  title={Cobra: Extending Mamba to Multi-Modal Large Language Model for Efficient Inference},
  author={Zhao, Han and Zhang, Min and Zhao, Wei and Ding, Pengxiang and Huang, Siteng and Wang, Donglin},
  journal={arXiv preprint arXiv:2403.14520},
  year={2024}
}

@article{li2024mini-gemini,
  title={Mini-Gemini: Mining the Potential of Multi-modality Vision Language Models},
  author={Li, Yanwei and Zhang, Yuechen and Wang, Chengyao and Zhong, Zhisheng and Chen, Yixin and Chu, Ruihang and Liu, Shaoteng and Jia, Jiaya},
  journal={arXiv preprint arXiv:2403.18814},
  year={2024}
}

@article{chen2024allava,
  title={ALLaVA: Harnessing GPT4V-synthesized Data for A Lite Vision-Language Model},
  author={Chen, Guiming Hardy and Chen, Shunian and Zhang, Ruifei and Chen, Junying and Wu, Xiangbo and Zhang, Zhiyi and Chen, Zhihong and Li, Jianquan and Wan, Xiang and Wang, Benyou},
  journal={arXiv preprint arXiv:2402.11684},
  year={2024}
}

@article{zhou2024tinyllava,
  title={TinyLLaVA: A Framework of Small-scale Large Multimodal Models},
  author={Zhou, Baichuan and Hu, Ying and Weng, Xi and Jia, Junlong and Luo, Jie and Liu, Xien and Wu, Ji and Huang, Lei},
  journal={arXiv preprint arXiv:2402.14289},
  year={2024}
}

@article{gao2024sphinx,
  title={SPHINX-X: Scaling Data and Parameters for a Family of Multi-modal Large Language Models},
  author={Gao, Peng and Zhang, Renrui and Liu, Chris and Qiu, Longtian and Huang, Siyuan and Lin, Weifeng and Zhao, Shitian and Geng, Shijie and Lin, Ziyi and Jin, Peng and others},
  journal={arXiv preprint arXiv:2402.05935},
  year={2024}
}

@article{team2023gemini,
  title={Gemini: a family of highly capable multimodal models},
  author={Team, Gemini and Anil, Rohan and Borgeaud, Sebastian and Wu, Yonghui and Alayrac, Jean-Baptiste and Yu, Jiahui and Soricut, Radu and Schalkwyk, Johan and Dai, Andrew M and Hauth, Anja and others},
  journal={arXiv preprint arXiv:2312.11805},
  year={2023}
}

@misc{imp2024,
  author = {Shao, Zhenwei and Ouyang, Xuecheng and Gai, Zhenbiao and Yu, Zhou and Yu, Jun},
  title = {Imp: An emprical study of multimodal small language models},
  year = {2024},
  url = {https://huggingface.co/MILVLG/imp-v1-3b}
}

@article{zhu2024mipha,
  title={A Comprehensive Overhaul of Multimodal Assistant with Small Language Models},
  author={Zhu, Minjie and Zhu, Yichen and Liu, Xin and Liu, Ning and Xu, Zhiyuan and Shen, Chaomin and Peng, Yaxin and Ou, Zhicai and Feng, Feifei and Tang, Jian},
  journal={arXiv preprint arXiv:2403.06199},
  year={2024}
}

@article{mckinzie2024mm1,
  title={Mm1: Methods, analysis \& insights from multimodal llm pre-training},
  author={McKinzie, Brandon and Gan, Zhe and Fauconnier, Jean-Philippe and Dodge, Sam and Zhang, Bowen and Dufter, Philipp and Shah, Dhruti and Du, Xianzhi and Peng, Futang and Weers, Floris and others},
  journal={arXiv preprint arXiv:2403.09611},
  year={2024}
}

@misc{shang2024llavaprumerge,
      title={LLaVA-PruMerge: Adaptive Token Reduction for Efficient Large Multimodal Models}, 
      author={Yuzhang Shang and Mu Cai and Bingxin Xu and Yong Jae Lee and Yan Yan},
      year={2024},
      eprint={2403.15388},
      archivePrefix={arXiv},
      primaryClass={cs.CV}
}

@misc{gagrani2024speculative,
      title={On Speculative Decoding for Multimodal Large Language Models}, 
      author={Mukul Gagrani and Raghavv Goel and Wonseok Jeon and Junyoung Park and Mingu Lee and Christopher Lott},
      year={2024},
      eprint={2404.08856},
      archivePrefix={arXiv},
      primaryClass={cs.CL}
}

@article{yu2024texthawk,
  title={TextHawk: Exploring Efficient Fine-Grained Perception of Multimodal Large Language Models},
  author={Yu, Ya-Qi and Liao, Minghui and Wu, Jihao and Liao, Yongxin and Zheng, Xiaoyu and Zeng, Wei},
  journal={arXiv preprint arXiv:2404.09204},
  year={2024}
}

@article{liu2024hrvda,
  title={HRVDA: High-Resolution Visual Document Assistant},
  author={Liu, Chaohu and Yin, Kun and Cao, Haoyu and Jiang, Xinghua and Li, Xin and Liu, Yinsong and Jiang, Deqiang and Sun, Xing and Xu, Linli},
  journal={arXiv preprint arXiv:2404.06918},
  year={2024}
}

@article{chen2024p2g,
  title={Plug-and-Play Grounding of Reasoning in Multimodal Large Language Models},
  author={Chen, Jiaxing and Liu, Yuxuan and Li, Dehu and An, Xiang and Feng, Ziyong and Zhao, Yongle and Xie, Yin},
  journal={arXiv preprint arXiv:2403.19322},
  year={2024}
}

@misc{wang2024elysium,
      title={Elysium: Exploring Object-level Perception in Videos via MLLM}, 
      author={Han Wang and Yanjie Wang and Yongjie Ye and Yuxiang Nie and Can Huang},
      year={2024},
      eprint={2403.16558},
      archivePrefix={arXiv},
      primaryClass={cs.CV}
}

@article{wu2024EAS,
  title={Not All Attention is Needed: Parameter and Computation Efficient Transfer Learning for Multi-modal Large Language Models},
  author={Wu, Qiong and Ye, Weihao and Zhou, Yiyi and Sun, Xiaoshuai and Ji, Rongrong},
  journal={arXiv preprint arXiv:2403.15226},
  year={2024}
}

@article{qiao2024vlmamba,
  title={VL-Mamba: Exploring State Space Models for Multimodal Learning},
  author={Qiao, Yanyuan and Yu, Zheng and Guo, Longteng and Chen, Sihan and Zhao, Zijia and Sun, Mingzhen and Wu, Qi and Liu, Jing},
  journal={arXiv preprint arXiv:2403.13600},
  year={2024}
}

@misc{zhang2024hyperllava,
      title={HyperLLaVA: Dynamic Visual and Language Expert Tuning for Multimodal Large Language Models}, 
      author={Wenqiao Zhang and Tianwei Lin and Jiang Liu and Fangxun Shu and Haoyuan Li and Lei Zhang and He Wanggui and Hao Zhou and Zheqi Lv and Hao Jiang and Juncheng Li and Siliang Tang and Yueting Zhuang},
      year={2024},
      eprint={2403.13447},
      archivePrefix={arXiv},
      primaryClass={cs.AI}
}

@article{shi2024S2,
  title={When Do We Not Need Larger Vision Models?},
  author={Shi, Baifeng and Wu, Ziyang and Mao, Maolin and Wang, Xin and Darrell, Trevor},
  journal={arXiv preprint arXiv:2403.13043},
  year={2024}
}

@misc{lu2024deepseekvl,
      title={DeepSeek-VL: Towards Real-World Vision-Language Understanding},
      author={Haoyu Lu and Wen Liu and Bo Zhang and Bingxuan Wang and Kai Dong and Bo Liu and Jingxiang Sun and Tongzheng Ren and Zhuoshu Li and Hao Yang and Yaofeng Sun and Chengqi Deng and Hanwei Xu and Zhenda Xie and Chong Ruan},
      year={2024},
      eprint={2403.05525},
      archivePrefix={arXiv},
      primaryClass={cs.AI}
}

@misc{chen2024vitamin,
      title={ViTamin: Designing Scalable Vision Models in the Vision-Language Era}, 
      author={Jieneng Chen and Qihang Yu and Xiaohui Shen and Alan Yuille and Liang-Chieh Chen},
      year={2024},
      eprint={2404.02132},
      archivePrefix={arXiv},
      primaryClass={cs.CV}
}

@article{tu2024Med-PaLM,
  title={Towards generalist biomedical ai},
  author={Tu, Tao and Azizi, Shekoofeh and Driess, Danny and Schaekermann, Mike and Amin, Mohamed and Chang, Pi-Chuan and Carroll, Andrew and Lau, Charles and Tanno, Ryutaro and Ktena, Ira and others},
  journal={NEJM AI},
  volume={1},
  number={3},
  pages={AIoa2300138},
  year={2024},
  publisher={Massachusetts Medical Society}
}

@inproceedings{azizi2021medical_cls,
  title={Big self-supervised models advance medical image classification},
  author={Azizi, Shekoofeh and Mustafa, Basil and Ryan, Fiona and Beaver, Zachary and Freyberg, Jan and Deaton, Jonathan and Loh, Aaron and Karthikesalingam, Alan and Kornblith, Simon and Chen, Ting and others},
  booktitle={Proceedings of the IEEE/CVF international conference on computer vision},
  pages={3478--3488},
  year={2021}
}

@misc{jiang2024moetinymed,
      title={MoE-TinyMed: Mixture of Experts for Tiny Medical Large Vision-Language Models}, 
      author={Songtao Jiang and Tuo Zheng and Yan Zhang and Yeying Jin and Zuozhu Liu},
      year={2024},
      eprint={2404.10237},
      archivePrefix={arXiv},
      primaryClass={cs.CV}
}

@misc{chaves2024llava-rad,
      title={Training Small Multimodal Models to Bridge Biomedical Competency Gap: A Case Study in Radiology Imaging}, 
      author={Juan Manuel Zambrano Chaves and Shih-Cheng Huang and Yanbo Xu and Hanwen Xu and Naoto Usuyama and Sheng Zhang and Fei Wang and Yujia Xie and Mahmoud Khademi and Ziyi Yang and Hany Awadalla and Julia Gong and Houdong Hu and Jianwei Yang and Chunyuan Li and Jianfeng Gao and Yu Gu and Cliff Wong and Mu Wei and Tristan Naumann and Muhao Chen and Matthew P. Lungren and Serena Yeung-Levy and Curtis P. Langlotz and Sheng Wang and Hoifung Poon},
      year={2024},
      eprint={2403.08002},
      archivePrefix={arXiv},
      primaryClass={cs.CL}
}

@misc{han2023chartllama,
      title={ChartLlama: A Multimodal LLM for Chart Understanding and Generation}, 
      author={Yucheng Han and Chi Zhang and Xin Chen and Xu Yang and Zhibin Wang and Gang Yu and Bin Fu and Hanwang Zhang},
      year={2023},
      eprint={2311.16483},
      archivePrefix={arXiv},
      primaryClass={cs.CV}
}

@misc{li2024monkey,
      title={Monkey: Image Resolution and Text Label Are Important Things for Large Multi-modal Models}, 
      author={Zhang Li and Biao Yang and Qiang Liu and Zhiyin Ma and Shuo Zhang and Jingxu Yang and Yabo Sun and Yuliang Liu and Xiang Bai},
      year={2024},
      eprint={2311.06607},
      archivePrefix={arXiv},
      primaryClass={cs.CV}
}

@article{li2023videochat,
  title={Videochat: Chat-centric video understanding},
  author={Li, KunChang and He, Yinan and Wang, Yi and Li, Yizhuo and Wang, Wenhai and Luo, Ping and Wang, Yali and Wang, Limin and Qiao, Yu},
  journal={arXiv preprint arXiv:2305.06355},
  year={2023}
}

@article{zhang2023videollama,
  title={Video-llama: An instruction-tuned audio-visual language model for video understanding},
  author={Zhang, Hang and Li, Xin and Bing, Lidong},
  journal={arXiv preprint arXiv:2306.02858},
  year={2023}
}

@misc{he2024malmm,
      title={MA-LMM: Memory-Augmented Large Multimodal Model for Long-Term Video Understanding}, 
      author={Bo He and Hengduo Li and Young Kyun Jang and Menglin Jia and Xuefei Cao and Ashish Shah and Abhinav Shrivastava and Ser-Nam Lim},
      year={2024},
      eprint={2404.05726},
      archivePrefix={arXiv},
      primaryClass={cs.CV}
}

@article{lin2023videollava,
  title={Video-LLaVA: Learning United Visual Representation by Alignment Before Projection},
  author={Lin, Bin and Zhu, Bin and Ye, Yang and Ning, Munan and Jin, Peng and Yuan, Li},
  journal={arXiv preprint arXiv:2311.10122},
  year={2023}
}

@article{zhu2023languagebind,
  title={LanguageBind: Extending Video-Language Pretraining to N-modality by Language-based Semantic Alignment},
  author={Zhu, Bin and Lin, Bin and Ning, Munan and Yan, Yang and Cui, Jiaxi and Wang, HongFa and Pang, Yatian and Jiang, Wenhao and Zhang, Junwu and Li, Zongwei and others},
  journal={arXiv preprint arXiv:2310.01852},
  year={2023}
}

@misc{xu2023mplug2,
      title={mPLUG-2: A Modularized Multi-modal Foundation Model Across Text, Image and Video}, 
      author={Haiyang Xu and Qinghao Ye and Ming Yan and Yaya Shi and Jiabo Ye and Yuanhong Xu and Chenliang Li and Bin Bi and Qi Qian and Wei Wang and Guohai Xu and Ji Zhang and Songfang Huang and Fei Huang and Jingren Zhou},
      year={2023},
      eprint={2302.00402},
      archivePrefix={arXiv},
      primaryClass={cs.CV}
}

@misc{xu2024llavauhd,
      title={LLaVA-UHD: an LMM Perceiving Any Aspect Ratio and High-Resolution Images}, 
      author={Ruyi Xu and Yuan Yao and Zonghao Guo and Junbo Cui and Zanlin Ni and Chunjiang Ge and Tat-Seng Chua and Zhiyuan Liu and Maosong Sun and Gao Huang},
      year={2024},
      eprint={2403.11703},
      archivePrefix={arXiv},
      primaryClass={cs.CV}
}

@misc{li2023llamavid,
      title={LLaMA-VID: An Image is Worth 2 Tokens in Large Language Models}, 
      author={Yanwei Li and Chengyao Wang and Jiaya Jia},
      year={2023},
      eprint={2311.17043},
      archivePrefix={arXiv},
      primaryClass={cs.CV}
}

@misc{lin2024vtw,
      title={Boosting Multimodal Large Language Models with Visual Tokens Withdrawal for Rapid Inference}, 
      author={Zhihang Lin and Mingbao Lin and Luxi Lin and Rongrong Ji},
      year={2024},
      eprint={2405.05803},
      archivePrefix={arXiv},
      primaryClass={cs.CV}
}

@misc{chen2024fastv,
      title={An Image is Worth 1/2 Tokens After Layer 2: Plug-and-Play Inference Acceleration for Large Vision-Language Models}, 
      author={Liang Chen and Haozhe Zhao and Tianyu Liu and Shuai Bai and Junyang Lin and Chang Zhou and Baobao Chang},
      year={2024},
      eprint={2403.06764},
      archivePrefix={arXiv},
      primaryClass={cs.CV}
}

@misc{cao2024madtp,
      title={MADTP: Multimodal Alignment-Guided Dynamic Token Pruning for Accelerating Vision-Language Transformer}, 
      author={Jianjian Cao and Peng Ye and Shengze Li and Chong Yu and Yansong Tang and Jiwen Lu and Tao Chen},
      year={2024},
      eprint={2403.02991},
      archivePrefix={arXiv},
      primaryClass={cs.CV}
}

@article{luo2024lavin,
  title={Cheap and quick: Efficient vision-language instruction tuning for large language models},
  author={Luo, Gen and Zhou, Yiyi and Ren, Tianhe and Chen, Shengxin and Sun, Xiaoshuai and Ji, Rongrong},
  journal={Advances in Neural Information Processing Systems},
  volume={36},
  year={2024}
}

@misc{laurençon2024idefics2,
      title={What matters when building vision-language models?}, 
      author={Hugo Laurençon and Léo Tronchon and Matthieu Cord and Victor Sanh},
      year={2024},
      eprint={2405.02246},
      archivePrefix={arXiv},
      primaryClass={cs.CV}
}

@misc{jie2024MemVP,
      title={Memory-Space Visual Prompting for Efficient Vision-Language Fine-Tuning}, 
      author={Shibo Jie and Yehui Tang and Ning Ding and Zhi-Hong Deng and Kai Han and Yunhe Wang},
      year={2024},
      eprint={2405.05615},
      archivePrefix={arXiv},
      primaryClass={cs.CV}
}

@misc{zong2024mova,
      title={MoVA: Adapting Mixture of Vision Experts to Multimodal Context}, 
      author={Zhuofan Zong and Bingqi Ma and Dazhong Shen and Guanglu Song and Hao Shao and Dongzhi Jiang and Hongsheng Li and Yu Liu},
      year={2024},
      eprint={2404.13046},
      archivePrefix={arXiv},
      primaryClass={cs.CV}
}

@article{dong2024internlm,
  title={InternLM-XComposer2-4KHD: A Pioneering Large Vision-Language Model Handling Resolutions from 336 Pixels to 4K HD},
  author={Dong, Xiaoyi and Zhang, Pan and Zang, Yuhang and Cao, Yuhang and Wang, Bin and Ouyang, Linke and Zhang, Songyang and Duan, Haodong and Zhang, Wenwei and Li, Yining and others},
  journal={arXiv preprint arXiv:2404.06512},
  year={2024}
}

@article{chen2023sharegpt4v,
  title={Sharegpt4v: Improving large multi-modal models with better captions},
  author={Chen, Lin and Li, Jisong and Dong, Xiaoyi and Zhang, Pan and He, Conghui and Wang, Jiaqi and Zhao, Feng and Lin, Dahua},
  journal={arXiv preprint arXiv:2311.12793},
  year={2023}
}

@inproceedings{liu2023llava,
author      = {Liu, Haotian and Li, Chunyuan and Wu, Qingyang and Lee, Yong Jae},
title       = {Visual Instruction Tuning},
booktitle   = {NeurIPS},
year        = {2023}
}

@article{alayrac2022flamingo,
  title={Flamingo: a visual language model for few-shot learning},
  author={Alayrac, Jean-Baptiste and Donahue, Jeff and Luc, Pauline and Miech, Antoine and Barr, Iain and Hasson, Yana and Lenc, Karel and Mensch, Arthur and Millican, Katherine and Reynolds, Malcolm and others},
  journal={Advances in neural information processing systems},
  volume={35},
  pages={23716--23736},
  year={2022}
}

@article{dai2024instructblip,
  title={Instructblip: Towards general-purpose vision-language models with instruction tuning},
  author={Dai, Wenliang and Li, Junnan and Li, Dongxu and Tiong, Anthony Meng Huat and Zhao, Junqi and Wang, Weisheng and Li, Boyang and Fung, Pascale N and Hoi, Steven},
  journal={Advances in Neural Information Processing Systems},
  volume={36},
  year={2024}
}

@inproceedings{li2023blip2,
  title={Blip-2: Bootstrapping language-image pre-training with frozen image encoders and large language models},
  author={Li, Junnan and Li, Dongxu and Savarese, Silvio and Hoi, Steven},
  booktitle={International conference on machine learning},
  pages={19730--19742},
  year={2023},
  organization={PMLR}
}

@article{bai2023qwen-vl,
  title={Qwen-vl: A frontier large vision-language model with versatile abilities},
  author={Bai, Jinze and Bai, Shuai and Yang, Shusheng and Wang, Shijie and Tan, Sinan and Wang, Peng and Lin, Junyang and Zhou, Chang and Zhou, Jingren},
  journal={arXiv preprint arXiv:2308.12966},
  year={2023}
}

@article{liu2023llava1.5,
  title={Improved baselines with visual instruction tuning},
  author={Liu, Haotian and Li, Chunyuan and Li, Yuheng and Lee, Yong Jae},
  journal={arXiv preprint arXiv:2310.03744},
  year={2023}
}

@article{chen2023minigpt-v2,
  title={Minigpt-v2: large language model as a unified interface for vision-language multi-task learning},
  author={Chen, Jun and Zhu, Deyao and Shen, Xiaoqian and Li, Xiang and Liu, Zechun and Zhang, Pengchuan and Krishnamoorthi, Raghuraman and Chandra, Vikas and Xiong, Yunyang and Elhoseiny, Mohamed},
  journal={arXiv preprint arXiv:2310.09478},
  year={2023}
}

@article{lin2023vila,
  title={Vila: On pre-training for visual language models},
  author={Lin, Ji and Yin, Hongxu and Ping, Wei and Lu, Yao and Molchanov, Pavlo and Tao, Andrew and Mao, Huizi and Kautz, Jan and Shoeybi, Mohammad and Han, Song},
  journal={arXiv preprint arXiv:2312.07533},
  year={2023}
}

@article{zhu2023minigpt4,
  title={Minigpt-4: Enhancing vision-language understanding with advanced large language models},
  author={Zhu, Deyao and Chen, Jun and Shen, Xiaoqian and Li, Xiang and Elhoseiny, Mohamed},
  journal={arXiv preprint arXiv:2304.10592},
  year={2023}
}

@article{chen2023internvl,
  title={Internvl: Scaling up vision foundation models and aligning for generic visual-linguistic tasks},
  author={Chen, Zhe and Wu, Jiannan and Wang, Wenhai and Su, Weijie and Chen, Guo and Xing, Sen and Muyan, Zhong and Zhang, Qinglong and Zhu, Xizhou and Lu, Lewei and others},
  journal={arXiv preprint arXiv:2312.14238},
  year={2023}
}

@article{sun2023generative,
  title={Generative multimodal models are in-context learners},
  author={Sun, Quan and Cui, Yufeng and Zhang, Xiaosong and Zhang, Fan and Yu, Qiying and Luo, Zhengxiong and Wang, Yueze and Rao, Yongming and Liu, Jingjing and Huang, Tiejun and others},
  journal={arXiv preprint arXiv:2312.13286},
  year={2023}
}

@article{achiam2023gpt,
  title={Gpt-4 technical report},
  author={Achiam, Josh and Adler, Steven and Agarwal, Sandhini and Ahmad, Lama and Akkaya, Ilge and Aleman, Florencia Leoni and Almeida, Diogo and Altenschmidt, Janko and Altman, Sam and Anadkat, Shyamal and others},
  journal={arXiv preprint arXiv:2303.08774},
  year={2023}
}

@misc{ye2023mplugowl,
      title={mPLUG-Owl: Modularization Empowers Large Language Models with Multimodality}, 
      author={Qinghao Ye and Haiyang Xu and Guohai Xu and Jiabo Ye and Ming Yan and Yiyang Zhou and Junyang Wang and Anwen Hu and Pengcheng Shi and Yaya Shi and Chaoya Jiang and Chenliang Li and Yuanhong Xu and Hehong Chen and Junfeng Tian and Qi Qian and Ji Zhang and Fei Huang},
      year={2023},
      eprint={2304.14178},
      archivePrefix={arXiv},
      primaryClass={cs.CL}
}

@misc{ye2023mplugowl2,
      title={mPLUG-Owl2: Revolutionizing Multi-modal Large Language Model with Modality Collaboration}, 
      author={Qinghao Ye and Haiyang Xu and Jiabo Ye and Ming Yan and Anwen Hu and Haowei Liu and Qi Qian and Ji Zhang and Fei Huang and Jingren Zhou},
      year={2023},
      eprint={2311.04257},
      archivePrefix={arXiv},
      primaryClass={cs.CL}
}

@misc{fu2024mme,
      title={MME: A Comprehensive Evaluation Benchmark for Multimodal Large Language Models}, 
      author={Chaoyou Fu and Peixian Chen and Yunhang Shen and Yulei Qin and Mengdan Zhang and Xu Lin and Jinrui Yang and Xiawu Zheng and Ke Li and Xing Sun and Yunsheng Wu and Rongrong Ji},
      year={2024},
      eprint={2306.13394},
      archivePrefix={arXiv},
      primaryClass={cs.CV}
}

@article{liu2023mmbench,
  title={Mmbench: Is your multi-modal model an all-around player?},
  author={Liu, Yuan and Duan, Haodong and Zhang, Yuanhan and Li, Bo and Zhang, Songyang and Zhao, Wangbo and Yuan, Yike and Wang, Jiaqi and He, Conghui and Liu, Ziwei and others},
  journal={arXiv preprint arXiv:2307.06281},
  year={2023}
}

@article{li2023seedbench,
  title={Seed-bench: Benchmarking multimodal llms with generative comprehension},
  author={Li, Bohao and Wang, Rui and Wang, Guangzhi and Ge, Yuying and Ge, Yixiao and Shan, Ying},
  journal={arXiv preprint arXiv:2307.16125},
  year={2023}
}

@article{yue2023mmmu,
  title={Mmmu: A massive multi-discipline multimodal understanding and reasoning benchmark for expert agi},
  author={Yue, Xiang and Ni, Yuansheng and Zhang, Kai and Zheng, Tianyu and Liu, Ruoqi and Zhang, Ge and Stevens, Samuel and Jiang, Dongfu and Ren, Weiming and Sun, Yuxuan and others},
  journal={arXiv preprint arXiv:2311.16502},
  year={2023}
}

@article{yu2023mmvet,
  title={Mm-vet: Evaluating large multimodal models for integrated capabilities},
  author={Yu, Weihao and Yang, Zhengyuan and Li, Linjie and Wang, Jianfeng and Lin, Kevin and Liu, Zicheng and Wang, Xinchao and Wang, Lijuan},
  journal={arXiv preprint arXiv:2308.02490},
  year={2023}
}

@article{lu2023mathvista,
  title={Mathvista: Evaluating mathematical reasoning of foundation models in visual contexts},
  author={Lu, Pan and Bansal, Hritik and Xia, Tony and Liu, Jiacheng and Li, Chunyuan and Hajishirzi, Hannaneh and Cheng, Hao and Chang, Kai-Wei and Galley, Michel and Gao, Jianfeng},
  journal={arXiv preprint arXiv:2310.02255},
  year={2023}
}

@inproceedings{goyal2017vqav2,
  title={Making the v in vqa matter: Elevating the role of image understanding in visual question answering},
  author={Goyal, Yash and Khot, Tejas and Summers-Stay, Douglas and Batra, Dhruv and Parikh, Devi},
  booktitle={Proceedings of the IEEE conference on computer vision and pattern recognition},
  pages={6904--6913},
  year={2017}
}

@inproceedings{singh2019vqat,
  title={Towards vqa models that can read},
  author={Singh, Amanpreet and Natarajan, Vivek and Shah, Meet and Jiang, Yu and Chen, Xinlei and Batra, Dhruv and Parikh, Devi and Rohrbach, Marcus},
  booktitle={Proceedings of the IEEE/CVF conference on computer vision and pattern recognition},
  pages={8317--8326},
  year={2019}
}

@inproceedings{hudson2019gqa,
  title={Gqa: A new dataset for real-world visual reasoning and compositional question answering},
  author={Hudson, Drew A and Manning, Christopher D},
  booktitle={Proceedings of the IEEE/CVF conference on computer vision and pattern recognition},
  pages={6700--6709},
  year={2019}
}

@article{lu2022scienceQA,
  title={Learn to explain: Multimodal reasoning via thought chains for science question answering},
  author={Lu, Pan and Mishra, Swaroop and Xia, Tanglin and Qiu, Liang and Chang, Kai-Wei and Zhu, Song-Chun and Tafjord, Oyvind and Clark, Peter and Kalyan, Ashwin},
  journal={Advances in Neural Information Processing Systems},
  volume={35},
  pages={2507--2521},
  year={2022}
}

@article{li2023pope,
  title={Evaluating object hallucination in large vision-language models},
  author={Li, Yifan and Du, Yifan and Zhou, Kun and Wang, Jinpeng and Zhao, Wayne Xin and Wen, Ji-Rong},
  journal={arXiv preprint arXiv:2305.10355},
  year={2023}
}

@inproceedings{gurari2018vizwiz,
  title={Vizwiz grand challenge: Answering visual questions from blind people},
  author={Gurari, Danna and Li, Qing and Stangl, Abigale J and Guo, Anhong and Lin, Chi and Grauman, Kristen and Luo, Jiebo and Bigham, Jeffrey P},
  booktitle={Proceedings of the IEEE conference on computer vision and pattern recognition},
  pages={3608--3617},
  year={2018}
}

@misc{wan2024efficientllm_survey,
      title={Efficient Large Language Models: A Survey}, 
      author={Zhongwei Wan and Xin Wang and Che Liu and Samiul Alam and Yu Zheng and Jiachen Liu and Zhongnan Qu and Shen Yan and Yi Zhu and Quanlu Zhang and Mosharaf Chowdhury and Mi Zhang},
      year={2024},
      eprint={2312.03863},
      archivePrefix={arXiv},
      primaryClass={cs.CL}
}

@article{touvron2023llama,
  title={Llama: Open and efficient foundation language models},
  author={Touvron, Hugo and Lavril, Thibaut and Izacard, Gautier and Martinet, Xavier and Lachaux, Marie-Anne and Lacroix, Timoth{\'e}e and Rozi{\`e}re, Baptiste and Goyal, Naman and Hambro, Eric and Azhar, Faisal and others},
  journal={arXiv preprint arXiv:2302.13971},
  year={2023}
}

@article{chiang2023vicuna,
  title={Vicuna: An open-source chatbot impressing gpt-4 with 90\%* chatgpt quality},
  author={Chiang, Wei-Lin and Li, Zhuohan and Lin, Zi and Sheng, Ying and Wu, Zhanghao and Zhang, Hao and Zheng, Lianmin and Zhuang, Siyuan and Zhuang, Yonghao and Gonzalez, Joseph E and others},
  journal={See https://vicuna. lmsys. org (accessed 14 April 2023)},
  volume={2},
  number={3},
  pages={6},
  year={2023}
}

@misc{abdin2024phi3,
      title={Phi-3 Technical Report: A Highly Capable Language Model Locally on Your Phone}, 
      author={Marah Abdin and Sam Ade Jacobs and Ammar Ahmad Awan and Jyoti Aneja and Ahmed Awadallah and Hany Awadalla and Nguyen Bach and Amit Bahree and Arash Bakhtiari and Harkirat Behl and Alon Benhaim and Misha Bilenko and Johan Bjorck and Sébastien Bubeck and Martin Cai and Caio César Teodoro Mendes and Weizhu Chen and Vishrav Chaudhary and Parul Chopra and Allie Del Giorno and Gustavo de Rosa and Matthew Dixon and Ronen Eldan and Dan Iter and Amit Garg and Abhishek Goswami and Suriya Gunasekar and Emman Haider and Junheng Hao and Russell J. Hewett and Jamie Huynh and Mojan Javaheripi and Xin Jin and Piero Kauffmann and Nikos Karampatziakis and Dongwoo Kim and Mahoud Khademi and Lev Kurilenko and James R. Lee and Yin Tat Lee and Yuanzhi Li and Chen Liang and Weishung Liu and Eric Lin and Zeqi Lin and Piyush Madan and Arindam Mitra and Hardik Modi and Anh Nguyen and Brandon Norick and Barun Patra and Daniel Perez-Becker and Thomas Portet and Reid Pryzant and Heyang Qin and Marko Radmilac and Corby Rosset and Sambudha Roy and Olatunji Ruwase and Olli Saarikivi and Amin Saied and Adil Salim and Michael Santacroce and Shital Shah and Ning Shang and Hiteshi Sharma and Xia Song and Masahiro Tanaka and Xin Wang and Rachel Ward and Guanhua Wang and Philipp Witte and Michael Wyatt and Can Xu and Jiahang Xu and Sonali Yadav and Fan Yang and Ziyi Yang and Donghan Yu and Chengruidong Zhang and Cyril Zhang and Jianwen Zhang and Li Lyna Zhang and Yi Zhang and Yue Zhang and Yunan Zhang and Xiren Zhou},
      year={2024},
      eprint={2404.14219},
      archivePrefix={arXiv},
      primaryClass={cs.CL}
}

@misc{gemmateam2024gemma,
      title={Gemma: Open Models Based on Gemini Research and Technology}, 
      author={Gemma Team and Thomas Mesnard and Cassidy Hardin and Robert Dadashi and Surya Bhupatiraju and Shreya Pathak and Laurent Sifre and Morgane Rivière and Mihir Sanjay Kale and Juliette Love and Pouya Tafti and Léonard Hussenot and Pier Giuseppe Sessa and Aakanksha Chowdhery and Adam Roberts and Aditya Barua and Alex Botev and Alex Castro-Ros and Ambrose Slone and Amélie Héliou and Andrea Tacchetti and Anna Bulanova and Antonia Paterson and Beth Tsai and Bobak Shahriari and Charline Le Lan and Christopher A. Choquette-Choo and Clément Crepy and Daniel Cer and Daphne Ippolito and David Reid and Elena Buchatskaya and Eric Ni and Eric Noland and Geng Yan and George Tucker and George-Christian Muraru and Grigory Rozhdestvenskiy and Henryk Michalewski and Ian Tenney and Ivan Grishchenko and Jacob Austin and James Keeling and Jane Labanowski and Jean-Baptiste Lespiau and Jeff Stanway and Jenny Brennan and Jeremy Chen and Johan Ferret and Justin Chiu and Justin Mao-Jones and Katherine Lee and Kathy Yu and Katie Millican and Lars Lowe Sjoesund and Lisa Lee and Lucas Dixon and Machel Reid and Maciej Mikuła and Mateo Wirth and Michael Sharman and Nikolai Chinaev and Nithum Thain and Olivier Bachem and Oscar Chang and Oscar Wahltinez and Paige Bailey and Paul Michel and Petko Yotov and Rahma Chaabouni and Ramona Comanescu and Reena Jana and Rohan Anil and Ross McIlroy and Ruibo Liu and Ryan Mullins and Samuel L Smith and Sebastian Borgeaud and Sertan Girgin and Sholto Douglas and Shree Pandya and Siamak Shakeri and Soham De and Ted Klimenko and Tom Hennigan and Vlad Feinberg and Wojciech Stokowiec and Yu-hui Chen and Zafarali Ahmed and Zhitao Gong and Tris Warkentin and Ludovic Peran and Minh Giang and Clément Farabet and Oriol Vinyals and Jeff Dean and Koray Kavukcuoglu and Demis Hassabis and Zoubin Ghahramani and Douglas Eck and Joelle Barral and Fernando Pereira and Eli Collins and Armand Joulin and Noah Fiedel and Evan Senter and Alek Andreev and Kathleen Kenealy},
      year={2024},
      eprint={2403.08295},
      archivePrefix={arXiv},
      primaryClass={cs.CL}
}

@misc{li2023phi,
      title={Textbooks Are All You Need II: phi-1.5 technical report}, 
      author={Yuanzhi Li and Sébastien Bubeck and Ronen Eldan and Allie Del Giorno and Suriya Gunasekar and Yin Tat Lee},
      year={2023},
      eprint={2309.05463},
      archivePrefix={arXiv},
      primaryClass={cs.CL}
}

@misc{zhang2024tinyllama,
      title={TinyLlama: An Open-Source Small Language Model}, 
      author={Peiyuan Zhang and Guangtao Zeng and Tianduo Wang and Wei Lu},
      year={2024},
      eprint={2401.02385},
      archivePrefix={arXiv},
      primaryClass={cs.CL}
}

@misc{bai2023qwen,
      title={Qwen Technical Report}, 
      author={Jinze Bai and Shuai Bai and Yunfei Chu and Zeyu Cui and Kai Dang and Xiaodong Deng and Yang Fan and Wenbin Ge and Yu Han and Fei Huang and Binyuan Hui and Luo Ji and Mei Li and Junyang Lin and Runji Lin and Dayiheng Liu and Gao Liu and Chengqiang Lu and Keming Lu and Jianxin Ma and Rui Men and Xingzhang Ren and Xuancheng Ren and Chuanqi Tan and Sinan Tan and Jianhong Tu and Peng Wang and Shijie Wang and Wei Wang and Shengguang Wu and Benfeng Xu and Jin Xu and An Yang and Hao Yang and Jian Yang and Shusheng Yang and Yang Yao and Bowen Yu and Hongyi Yuan and Zheng Yuan and Jianwei Zhang and Xingxuan Zhang and Yichang Zhang and Zhenru Zhang and Chang Zhou and Jingren Zhou and Xiaohuan Zhou and Tianhang Zhu},
      year={2023},
      eprint={2309.16609},
      archivePrefix={arXiv},
      primaryClass={cs.CL}
}

@misc{touvron2023llama2,
      title={Llama 2: Open Foundation and Fine-Tuned Chat Models}, 
      author={Hugo Touvron and Louis Martin and Kevin Stone and Peter Albert and Amjad Almahairi and Yasmine Babaei and Nikolay Bashlykov and Soumya Batra and Prajjwal Bhargava and Shruti Bhosale and Dan Bikel and Lukas Blecher and Cristian Canton Ferrer and Moya Chen and Guillem Cucurull and David Esiobu and Jude Fernandes and Jeremy Fu and Wenyin Fu and Brian Fuller and Cynthia Gao and Vedanuj Goswami and Naman Goyal and Anthony Hartshorn and Saghar Hosseini and Rui Hou and Hakan Inan and Marcin Kardas and Viktor Kerkez and Madian Khabsa and Isabel Kloumann and Artem Korenev and Punit Singh Koura and Marie-Anne Lachaux and Thibaut Lavril and Jenya Lee and Diana Liskovich and Yinghai Lu and Yuning Mao and Xavier Martinet and Todor Mihaylov and Pushkar Mishra and Igor Molybog and Yixin Nie and Andrew Poulton and Jeremy Reizenstein and Rashi Rungta and Kalyan Saladi and Alan Schelten and Ruan Silva and Eric Michael Smith and Ranjan Subramanian and Xiaoqing Ellen Tan and Binh Tang and Ross Taylor and Adina Williams and Jian Xiang Kuan and Puxin Xu and Zheng Yan and Iliyan Zarov and Yuchen Zhang and Angela Fan and Melanie Kambadur and Sharan Narang and Aurelien Rodriguez and Robert Stojnic and Sergey Edunov and Thomas Scialom},
      year={2023},
      eprint={2307.09288},
      archivePrefix={arXiv},
      primaryClass={cs.CL}
}

@misc{jiang2024moe,
      title={Mixtral of Experts}, 
      author={Albert Q. Jiang and Alexandre Sablayrolles and Antoine Roux and Arthur Mensch and Blanche Savary and Chris Bamford and Devendra Singh Chaplot and Diego de las Casas and Emma Bou Hanna and Florian Bressand and Gianna Lengyel and Guillaume Bour and Guillaume Lample and Lélio Renard Lavaud and Lucile Saulnier and Marie-Anne Lachaux and Pierre Stock and Sandeep Subramanian and Sophia Yang and Szymon Antoniak and Teven Le Scao and Théophile Gervet and Thibaut Lavril and Thomas Wang and Timothée Lacroix and William El Sayed},
      year={2024},
      eprint={2401.04088},
      archivePrefix={arXiv},
      primaryClass={cs.LG}
}

@article{Papa_2024vit_survey,
   title={A Survey on Efficient Vision Transformers: Algorithms, Techniques, and Performance Benchmarking},
   ISSN={1939-3539},
   url={http://dx.doi.org/10.1109/TPAMI.2024.3392941},
   DOI={10.1109/tpami.2024.3392941},
   journal={IEEE Transactions on Pattern Analysis and Machine Intelligence},
   publisher={Institute of Electrical and Electronics Engineers (IEEE)},
   author={Papa, Lorenzo and Russo, Paolo and Amerini, Irene and Zhou, Luping},
   year={2024},
   pages={1–20} }

@article{rao2021dynamicvit,
  title={Dynamicvit: Efficient vision transformers with dynamic token sparsification},
  author={Rao, Yongming and Zhao, Wenliang and Liu, Benlin and Lu, Jiwen and Zhou, Jie and Hsieh, Cho-Jui},
  journal={Advances in neural information processing systems},
  volume={34},
  pages={13937--13949},
  year={2021}
}

@article{dosovitskiy2020vit,
  title={An Image is Worth 16x16 Words: Transformers for Image Recognition at Scale},
  author={Dosovitskiy, Alexey and Beyer, Lucas and Kolesnikov, Alexander and Weissenborn, Dirk and Zhai, Xiaohua and Unterthiner, Thomas and  Dehghani, Mostafa and Minderer, Matthias and Heigold, Georg and Gelly, Sylvain and Uszkoreit, Jakob and Houlsby, Neil},
  journal={ICLR},
  year={2021}
}

@inproceedings{zhai2023siglip,
  title={Sigmoid loss for language image pre-training},
  author={Zhai, Xiaohua and Mustafa, Basil and Kolesnikov, Alexander and Beyer, Lucas},
  booktitle={Proceedings of the IEEE/CVF International Conference on Computer Vision},
  pages={11975--11986},
  year={2023}
}

@inproceedings{radford2021clip,
  title={Learning transferable visual models from natural language supervision},
  author={Radford, Alec and Kim, Jong Wook and Hallacy, Chris and Ramesh, Aditya and Goh, Gabriel and Agarwal, Sandhini and Sastry, Girish and Askell, Amanda and Mishkin, Pamela and Clark, Jack and others},
  booktitle={International conference on machine learning},
  pages={8748--8763},
  year={2021},
  organization={PMLR}
}

@article{oquab2023dinov2,
  title={Dinov2: Learning robust visual features without supervision},
  author={Oquab, Maxime and Darcet, Timoth{\'e}e and Moutakanni, Th{\'e}o and Vo, Huy and Szafraniec, Marc and Khalidov, Vasil and Fernandez, Pierre and Haziza, Daniel and Massa, Francisco and El-Nouby, Alaaeldin and others},
  journal={arXiv preprint arXiv:2304.07193},
  year={2023}
}

@inproceedings{fang2023eva,
  title={Eva: Exploring the limits of masked visual representation learning at scale},
  author={Fang, Yuxin and Wang, Wen and Xie, Binhui and Sun, Quan and Wu, Ledell and Wang, Xinggang and Huang, Tiejun and Wang, Xinlong and Cao, Yue},
  booktitle={Proceedings of the IEEE/CVF Conference on Computer Vision and Pattern Recognition},
  pages={19358--19369},
  year={2023}
}

@misc{ainslie2023gqa,
      title={GQA: Training Generalized Multi-Query Transformer Models from Multi-Head Checkpoints}, 
      author={Joshua Ainslie and James Lee-Thorp and Michiel de Jong and Yury Zemlyanskiy and Federico Lebrón and Sumit Sanghai},
      year={2023},
      eprint={2305.13245},
      archivePrefix={arXiv},
      primaryClass={cs.CL}
}

@article{shazeer2019fast,
  title={Fast transformer decoding: One write-head is all you need},
  author={Shazeer, Noam},
  journal={arXiv preprint arXiv:1911.02150},
  year={2019}
}

@article{dai2020funnel,
  title={Funnel-transformer: Filtering out sequential redundancy for efficient language processing},
  author={Dai, Zihang and Lai, Guokun and Yang, Yiming and Le, Quoc},
  journal={Advances in neural information processing systems},
  volume={33},
  pages={4271--4282},
  year={2020}
}

@misc{wang2020linformer,
    title={Linformer: Self-Attention with Linear Complexity},
    author={Sinong Wang and Belinda Z. Li and Madian Khabsa and Han Fang and Hao Ma},
    year={2020},
    eprint={2006.04768},
    archivePrefix={arXiv},
    primaryClass={cs.LG}
}

@article{choromanski2020masked,
  title={Masked language modeling for proteins via linearly scalable long-context transformers},
  author={Choromanski, Krzysztof and Likhosherstov, Valerii and Dohan, David and Song, Xingyou and Gane, Andreea and Sarlos, Tamas and Hawkins, Peter and Davis, Jared and Belanger, David and Colwell, Lucy and others},
  journal={arXiv preprint arXiv:2006.03555},
  year={2020}
}

@article{lepikhin2020gshard,
  title={Gshard: Scaling giant models with conditional computation and automatic sharding},
  author={Lepikhin, Dmitry and Lee, HyoukJoong and Xu, Yuanzhong and Chen, Dehao and Firat, Orhan and Huang, Yanping and Krikun, Maxim and Shazeer, Noam and Chen, Zhifeng},
  journal={arXiv preprint arXiv:2006.16668},
  year={2020}
}

@article{fedus2022switch,
  title={Switch transformers: Scaling to trillion parameter models with simple and efficient sparsity},
  author={Fedus, William and Zoph, Barret and Shazeer, Noam},
  journal={Journal of Machine Learning Research},
  volume={23},
  number={120},
  pages={1--39},
  year={2022}
}

@article{gu2021efficiently,
  title={Efficiently modeling long sequences with structured state spaces},
  author={Gu, Albert and Goel, Karan and R{\'e}, Christopher},
  journal={arXiv preprint arXiv:2111.00396},
  year={2021}
}

@article{gupta2022diagonal,
  title={Diagonal state spaces are as effective as structured state spaces},
  author={Gupta, Ankit and Gu, Albert and Berant, Jonathan},
  journal={Advances in Neural Information Processing Systems},
  volume={35},
  pages={22982--22994},
  year={2022}
}

@article{peng2023rwkv,
  title={Rwkv: Reinventing rnns for the transformer era},
  author={Peng, Bo and Alcaide, Eric and Anthony, Quentin and Albalak, Alon and Arcadinho, Samuel and Cao, Huanqi and Cheng, Xin and Chung, Michael and Grella, Matteo and GV, Kranthi Kiran and others},
  journal={arXiv preprint arXiv:2305.13048},
  year={2023}
}

@article{gu2023mamba,
  title={Mamba: Linear-time sequence modeling with selective state spaces},
  author={Gu, Albert and Dao, Tri},
  journal={arXiv preprint arXiv:2312.00752},
  year={2023}
}

@article{hu2023llm,
  title={LLM-Adapters: An Adapter Family for Parameter-Efficient Fine-Tuning of Large Language Models},
  author={Hu, Zhiqiang and Lan, Yihuai and Wang, Lei and Xu, Wanyu and Lim, Ee-Peng and Lee, Roy Ka-Wei and Bing, Lidong and Poria, Soujanya},
  journal={arXiv preprint arXiv:2304.01933},
  year={2023}
}

@article{liu2020tfew,
  title={Few-Shot Parameter-Efficient Fine-Tuning is Better and Cheaper than In-Context Learning},
  author={Liu, Haokun and Tam, Derek and Muqeeth, Mohammed and Mohta, Jay and Huang, Tenghao and Bansal, Mohit and Raffel, Colin},
  journal={arXiv preprint arXiv:2205.05638},
  year={2022}
}

@inproceedings{
hu2022lora,
title={Lo{RA}: Low-Rank Adaptation of Large Language Models},
author={Edward J Hu and Yelong Shen and Phillip Wallis and Zeyuan Allen-Zhu and Yuanzhi Li and Shean Wang and Lu Wang and Weizhu Chen},
booktitle={International Conference on Learning Representations},
year={2022},
url={https://openreview.net/forum?id=nZeVKeeFYf9}
}

@article{zhang2023lora,
  title={Lora-fa: Memory-efficient low-rank adaptation for large language models fine-tuning},
  author={Zhang, Longteng and Zhang, Lin and Shi, Shaohuai and Chu, Xiaowen and Li, Bo},
  journal={arXiv preprint arXiv:2308.03303},
  year={2023}
}

@article{valipour2022dylora,
  title={Dylora: Parameter efficient tuning of pre-trained models using dynamic search-free low-rank adaptation},
  author={Valipour, Mojtaba and Rezagholizadeh, Mehdi and Kobyzev, Ivan and Ghodsi, Ali},
  journal={arXiv preprint arXiv:2210.07558},
  year={2022}
}

@article{lv2023full,
  title={Full Parameter Fine-tuning for Large Language Models with Limited Resources},
  author={Lv, Kai and Yang, Yuqing and Liu, Tengxiao and Gao, Qinghui and Guo, Qipeng and Qiu, Xipeng},
  journal={arXiv preprint arXiv:2306.09782},
  year={2023}
}

@article{internlmxcomposer2_4khd,
      title={InternLM-XComposer2-4KHD: A Pioneering Large Vision-Language Model Handling Resolutions from 336 Pixels to 4K HD},
      author={Xiaoyi Dong and Pan Zhang and Yuhang Zang and Yuhang Cao and Bin Wang and Linke Ouyang and Songyang Zhang and Haodong Duan and Wenwei Zhang and Yining Li and Hang Yan and Yang Gao and Zhe Chen and Xinyue Zhang and Wei Li and Jingwen Li and Wenhai Wang and Kai Chen and Conghui He and Xingcheng Zhang and Jifeng Dai and Yu Qiao and Dahua Lin and Jiaqi Wang},
      journal={arXiv preprint arXiv:2404.06512},
      year={2024}
}

@article{zhang2024tinychart,
  title={TinyChart: Efficient Chart Understanding with Visual Token Merging and Program-of-Thoughts Learning},
  author={Zhang, Liang and Hu, Anwen and Xu, Haiyang and Yan, Ming and Xu, Yichen and Jin, Qin and Zhang, Ji and Huang, Fei},
  journal={arXiv preprint arXiv:2404.16635},
  year={2024}
}

@inproceedings{liu2022convnet,
  title={A convnet for the 2020s},
  author={Liu, Zhuang and Mao, Hanzi and Wu, Chao-Yuan and Feichtenhofer, Christoph and Darrell, Trevor and Xie, Saining},
  booktitle={Proceedings of the IEEE/CVF conference on computer vision and pattern recognition},
  pages={11976--11986},
  year={2022}
}

@article{cha2023honeybee,
  title={Honeybee: Locality-enhanced projector for multimodal llm},
  author={Cha, Junbum and Kang, Wooyoung and Mun, Jonghwan and Roh, Byungseok},
  journal={arXiv preprint arXiv:2312.06742},
  year={2023}
}

@article{fang2023data,
  title={Data filtering networks},
  author={Fang, Alex and Jose, Albin Madappally and Jain, Amit and Schmidt, Ludwig and Toshev, Alexander and Shankar, Vaishaal},
  journal={arXiv preprint arXiv:2309.17425},
  year={2023}
}

@article{javaheripi2023phi,
  title={Phi-2: The surprising power of small language models},
  author={Javaheripi, Mojan and Bubeck, S{\'e}bastien and Abdin, Marah and Aneja, Jyoti and Bubeck, Sebastien and Mendes, Caio C{\'e}sar Teodoro and Chen, Weizhu and Del Giorno, Allie and Eldan, Ronen and Gopi, Sivakanth and others},
  journal={Microsoft Research Blog},
  year={2023}
}

@article{deepseek-llm,
  author = {DeepSeek-AI},
  title = {DeepSeek LLM: Scaling Open-Source Language Models with Longtermism},
  journal = {arXiv preprint arXiv:2401.02954},
  year = {2024},
  url = {https://github.com/deepseek-ai/DeepSeek-LLM}
}

@article{hu2024minicpm,
  title={MiniCPM: Unveiling the Potential of Small Language Models with Scalable Training Strategies},
  author={Hu, Shengding and Tu, Yuge and Han, Xu and He, Chaoqun and Cui, Ganqu and Long, Xiang and Zheng, Zhi and Fang, Yewei and Huang, Yuxiang and Zhao, Weilin and others},
  journal={arXiv preprint arXiv:2404.06395},
  year={2024}
}

@misc{minicpm-v,
  author = {Yao, Yuan and Yu, Tianyu and Wang, Chongyi and Cui, Junbo and Zhu, Hongji and Cai, Tianchi and Zhao, Weilin and Zhang, Kaihuo and Hong, Yixin and Li, Haoyu and Hu, Shengding and Zheng, Zhi and Zhou, Jie and Cai, Jie and Jia, Chao and Han, Xu and Zeng, Guoyang and Li, Dahai and Liu, Zhiyuan and Sun, Maosong},
  title = {MiniCPM-V 2.0: An Efficient End-side MLLM with Strong OCR and Understanding Capabilities},
  year = {2024},
  publisher = {GitHub},
  journal = {GitHub repository},
  howpublished = {\url{https://github.com/OpenBMB/MiniCPM-V}},
}

@misc{karmavlm,
  title = {KarmaVLM: A family of high efficiency and powerful visual language model},
  year = {2024},
  publisher = {GitHub},
  journal = {GitHub repository},
  howpublished = {\url{https://github.com/thomas-yanxin/KarmaVLM}},
}

@misc{moondream,
  title = {tiny vision language model},
  year = {2024},
  publisher = {GitHub},
  journal = {GitHub repository},
  howpublished = {\url{https://github.com/vikhyat/moondream}},
}

@article{kitaev2020reformer,
  title={Reformer: The efficient transformer},
  author={Kitaev, Nikita and Kaiser, {\L}ukasz and Levskaya, Anselm},
  journal={arXiv preprint arXiv:2001.04451},
  year={2020}
}

@article{hassani2021escaping,
  title={Escaping the big data paradigm with compact transformers},
  author={Hassani, Ali and Walton, Steven and Shah, Nikhil and Abuduweili, Abulikemu and Li, Jiachen and Shi, Humphrey},
  journal={arXiv preprint arXiv:2104.05704},
  year={2021}
}

@inproceedings{fayyaz2022adaptive,
  title={Adaptive token sampling for efficient vision transformers},
  author={Fayyaz, Mohsen and Koohpayegani, Soroush Abbasi and Jafari, Farnoush Rezaei and Sengupta, Sunando and Joze, Hamid Reza Vaezi and Sommerlade, Eric and Pirsiavash, Hamed and Gall, J{\"u}rgen},
  booktitle={European Conference on Computer Vision},
  pages={396--414},
  year={2022},
  organization={Springer}
}

@article{li2022sepvit,
  title={Sepvit: Separable vision transformer},
  author={Li, Wei and Wang, Xing and Xia, Xin and Wu, Jie and Xiao, Xuefeng and Zheng, Min and Wen, Shiping},
  journal={arXiv preprint arXiv:2203.15380},
  year={2022}
}

@article{renggli2022learning,
  title={Learning to merge tokens in vision transformers},
  author={Renggli, Cedric and Pinto, Andr{\'e} Susano and Houlsby, Neil and Mustafa, Basil and Puigcerver, Joan and Riquelme, Carlos},
  journal={arXiv preprint arXiv:2202.12015},
  year={2022}
}

@article{wang2023crossformer++,
  title={Crossformer++: A versatile vision transformer hinging on cross-scale attention},
  author={Wang, Wenxiao and Chen, Wei and Qiu, Qibo and Chen, Long and Wu, Boxi and Lin, Binbin and He, Xiaofei and Liu, Wei},
  journal={IEEE Transactions on Pattern Analysis and Machine Intelligence},
  year={2023},
  publisher={IEEE}
}

@article{li2022efficientformer,
  title={Efficientformer: Vision transformers at mobilenet speed},
  author={Li, Yanyu and Yuan, Geng and Wen, Yang and Hu, Ju and Evangelidis, Georgios and Tulyakov, Sergey and Wang, Yanzhi and Ren, Jian},
  journal={Advances in Neural Information Processing Systems},
  volume={35},
  pages={12934--12949},
  year={2022}
}

@inproceedings{chen2021autoformer,
  title={Autoformer: Searching transformers for visual recognition},
  author={Chen, Minghao and Peng, Houwen and Fu, Jianlong and Ling, Haibin},
  booktitle={Proceedings of the IEEE/CVF international conference on computer vision},
  pages={12270--12280},
  year={2021}
}

@article{gong2022nasvit,
  title={NASViT: Neural architecture search for efficient vision transformers with gradient conflict-aware supernet training},
  author={Gong, Chengyue and Wang, Dilin},
  journal={ICLR Proceedings 2022},
  year={2022}
}

@inproceedings{zhou2022training,
  title={Training-free transformer architecture search},
  author={Zhou, Qinqin and Sheng, Kekai and Zheng, Xiawu and Li, Ke and Sun, Xing and Tian, Yonghong and Chen, Jie and Ji, Rongrong},
  booktitle={Proceedings of the IEEE/CVF Conference on Computer Vision and Pattern Recognition},
  pages={10894--10903},
  year={2022}
}

@inproceedings{chavan2022vision,
  title={Vision transformer slimming: Multi-dimension searching in continuous optimization space},
  author={Chavan, Arnav and Shen, Zhiqiang and Liu, Zhuang and Liu, Zechun and Cheng, Kwang-Ting and Xing, Eric P},
  booktitle={Proceedings of the IEEE/CVF Conference on Computer Vision and Pattern Recognition},
  pages={4931--4941},
  year={2022}
}

@inproceedings{liu2022uninet,
  title={Uninet: Unified architecture search with convolution, transformer, and mlp},
  author={Liu, Jihao and Huang, Xin and Song, Guanglu and Li, Hongsheng and Liu, Yu},
  booktitle={European Conference on Computer Vision},
  pages={33--49},
  year={2022},
  organization={Springer}
}

@inproceedings{li2023rethinking,
  title={Rethinking vision transformers for mobilenet size and speed},
  author={Li, Yanyu and Hu, Ju and Wen, Yang and Evangelidis, Georgios and Salahi, Kamyar and Wang, Yanzhi and Tulyakov, Sergey and Ren, Jian},
  booktitle={Proceedings of the IEEE/CVF International Conference on Computer Vision},
  pages={16889--16900},
  year={2023}
}

@article{chen2021chasing,
  title={Chasing sparsity in vision transformers: An end-to-end exploration},
  author={Chen, Tianlong and Cheng, Yu and Gan, Zhe and Yuan, Lu and Zhang, Lei and Wang, Zhangyang},
  journal={Advances in Neural Information Processing Systems},
  volume={34},
  pages={19974--19988},
  year={2021}
}

@inproceedings{yu2022width,
  title={Width \& depth pruning for vision transformers},
  author={Yu, Fang and Huang, Kun and Wang, Meng and Cheng, Yuan and Chu, Wei and Cui, Li},
  booktitle={Proceedings of the AAAI Conference on Artificial Intelligence},
  volume={36},
  pages={3143--3151},
  year={2022}
}

@article{yu2022unified,
  title={Unified visual transformer compression},
  author={Yu, Shixing and Chen, Tianlong and Shen, Jiayi and Yuan, Huan and Tan, Jianchao and Yang, Sen and Liu, Ji and Wang, Zhangyang},
  journal={arXiv preprint arXiv:2203.08243},
  year={2022}
}

@inproceedings{yu2023x,
  title={X-pruner: explainable pruning for vision transformers},
  author={Yu, Lu and Xiang, Wei},
  booktitle={Proceedings of the IEEE/CVF Conference on Computer Vision and Pattern Recognition},
  pages={24355--24363},
  year={2023}
}

@inproceedings{tang2022patch,
  title={Patch slimming for efficient vision transformers},
  author={Tang, Yehui and Han, Kai and Wang, Yunhe and Xu, Chang and Guo, Jianyuan and Xu, Chao and Tao, Dacheng},
  booktitle={Proceedings of the IEEE/CVF Conference on Computer Vision and Pattern Recognition},
  pages={12165--12174},
  year={2022}
}

@article{zhu2021vision,
  title={Vision transformer pruning},
  author={Zhu, Mingjian and Tang, Yehui and Han, Kai},
  journal={arXiv preprint arXiv:2104.08500},
  year={2021}
}

@article{kuznedelev2024cap,
  title={CAP: Correlation-Aware Pruning for Highly-Accurate Sparse Vision Models},
  author={Kuznedelev, Denis and Kurti{\'c}, Eldar and Frantar, Elias and Alistarh, Dan},
  journal={Advances in Neural Information Processing Systems},
  volume={36},
  year={2024}
}

@article{wang2023cait,
  title={CAIT: Triple-Win Compression towards High Accuracy, Fast Inference, and Favorable Transferability For ViTs},
  author={Wang, Ao and Chen, Hui and Lin, Zijia and Zhao, Sicheng and Han, Jungong and Ding, Guiguang},
  journal={arXiv preprint arXiv:2309.15755},
  year={2023}
}

@inproceedings{kong2022spvit,
  title={Spvit: Enabling faster vision transformers via latency-aware soft token pruning},
  author={Kong, Zhenglun and Dong, Peiyan and Ma, Xiaolong and Meng, Xin and Niu, Wei and Sun, Mengshu and Shen, Xuan and Yuan, Geng and Ren, Bin and Tang, Hao and others},
  booktitle={European conference on computer vision},
  pages={620--640},
  year={2022},
  organization={Springer}
}

@article{kar2024brave,
        title={{BRAVE}: Broadening the visual encoding of vision-language models},
        author={Kar, O{\u{g}}uzhan Fatih and Tonioni, Alessio and Poklukar, Petra and Kulshrestha, Achin and Zamir, Amir and Tombari, Federico},
        journal={arXiv preprint arXiv:2404.07204},
        year={2024}
      }

@inproceedings{sharma2018conceptual,
  title={Conceptual captions: A cleaned, hypernymed, image alt-text dataset for automatic image captioning},
  author={Sharma, Piyush and Ding, Nan and Goodman, Sebastian and Soricut, Radu},
  booktitle={Proceedings of the 56th Annual Meeting of the Association for Computational Linguistics (Volume 1: Long Papers)},
  pages={2556--2565},
  year={2018}
}

@inproceedings{changpinyo2021conceptual,
  title={Conceptual 12m: Pushing web-scale image-text pre-training to recognize long-tail visual concepts},
  author={Changpinyo, Soravit and Sharma, Piyush and Ding, Nan and Soricut, Radu},
  booktitle={Proceedings of the IEEE/CVF conference on computer vision and pattern recognition},
  pages={3558--3568},
  year={2021}
}

@article{ordonez2011im2text,
  title={Im2text: Describing images using 1 million captioned photographs},
  author={Ordonez, Vicente and Kulkarni, Girish and Berg, Tamara},
  journal={Advances in neural information processing systems},
  volume={24},
  year={2011}
}

@article{schuhmann2022laion,
  title={Laion-5b: An open large-scale dataset for training next generation image-text models},
  author={Schuhmann, Christoph and Beaumont, Romain and Vencu, Richard and Gordon, Cade and Wightman, Ross and Cherti, Mehdi and Coombes, Theo and Katta, Aarush and Mullis, Clayton and Wortsman, Mitchell and others},
  journal={Advances in Neural Information Processing Systems},
  volume={35},
  pages={25278--25294},
  year={2022}
}

@article{schuhmann2021laion,
  title={Laion-400m: Open dataset of clip-filtered 400 million image-text pairs},
  author={Schuhmann, Christoph and Vencu, Richard and Beaumont, Romain and Kaczmarczyk, Robert and Mullis, Clayton and Katta, Aarush and Coombes, Theo and Jitsev, Jenia and Komatsuzaki, Aran},
  journal={arXiv preprint arXiv:2111.02114},
  year={2021}
}

@misc{kakaobrain2022coyo-700m,
  title         = {COYO-700M: Image-Text Pair Dataset},
  author        = {Byeon, Minwoo and Park, Beomhee and Kim, Haecheon and Lee, Sungjun and Baek, Woonhyuk and Kim, Saehoon},
  year          = {2022},
  howpublished  = {\url{https://github.com/kakaobrain/coyo-dataset}},
}

@article{chen2015microsoft,
  title={Microsoft coco captions: Data collection and evaluation server},
  author={Chen, Xinlei and Fang, Hao and Lin, Tsung-Yi and Vedantam, Ramakrishna and Gupta, Saurabh and Doll{\'a}r, Piotr and Zitnick, C Lawrence},
  journal={arXiv preprint arXiv:1504.00325},
  year={2015}
}

@inproceedings{kazemzadeh2014referitgame,
  title={Referitgame: Referring to objects in photographs of natural scenes},
  author={Kazemzadeh, Sahar and Ordonez, Vicente and Matten, Mark and Berg, Tamara},
  booktitle={Proceedings of the 2014 conference on empirical methods in natural language processing (EMNLP)},
  pages={787--798},
  year={2014}
}

@inproceedings{mathew2021docvqa,
  title={Docvqa: A dataset for vqa on document images},
  author={Mathew, Minesh and Karatzas, Dimosthenis and Jawahar, CV},
  booktitle={Proceedings of the IEEE/CVF winter conference on applications of computer vision},
  pages={2200--2209},
  year={2021}
}

@article{zhu2024multimodal,
  title={Multimodal c4: An open, billion-scale corpus of images interleaved with text},
  author={Zhu, Wanrong and Hessel, Jack and Awadalla, Anas and Gadre, Samir Yitzhak and Dodge, Jesse and Fang, Alex and Yu, Youngjae and Schmidt, Ludwig and Wang, William Yang and Choi, Yejin},
  journal={Advances in Neural Information Processing Systems},
  volume={36},
  year={2024}
}

@article{laurenccon2024obelics,
  title={Obelics: An open web-scale filtered dataset of interleaved image-text documents},
  author={Lauren{\c{c}}on, Hugo and Saulnier, Lucile and Tronchon, L{\'e}o and Bekman, Stas and Singh, Amanpreet and Lozhkov, Anton and Wang, Thomas and Karamcheti, Siddharth and Rush, Alexander and Kiela, Douwe and others},
  journal={Advances in Neural Information Processing Systems},
  volume={36},
  year={2024}
}

@inproceedings{liu2023mitigating,
  title={Mitigating hallucination in large multi-modal models via robust instruction tuning},
  author={Liu, Fuxiao and Lin, Kevin and Li, Linjie and Wang, Jianfeng and Yacoob, Yaser and Wang, Lijuan},
  booktitle={The Twelfth International Conference on Learning Representations},
  year={2023}
}

@article{wang2023see,
  title={To see is to believe: Prompting gpt-4v for better visual instruction tuning},
  author={Wang, Junke and Meng, Lingchen and Weng, Zejia and He, Bo and Wu, Zuxuan and Jiang, Yu-Gang},
  journal={arXiv preprint arXiv:2311.07574},
  year={2023}
}

@misc{laion2023gpt4v,
    title={Gpt-4v dataset},
    author={LAION},
    year={2023},
    howpublished={\url{https://huggingface.co/datasets/laion/gpt4v-dataset}},
}

@article{zhao2023svit,
  title={Svit: Scaling up visual instruction tuning},
  author={Zhao, Bo and Wu, Boya and Huang, Tiejun},
  journal={arXiv preprint arXiv:2307.04087},
  year={2023}
}

@inproceedings{lin2014microsoft,
  title={Microsoft coco: Common objects in context},
  author={Lin, Tsung-Yi and Maire, Michael and Belongie, Serge and Hays, James and Perona, Pietro and Ramanan, Deva and Doll{\'a}r, Piotr and Zitnick, C Lawrence},
  booktitle={Computer Vision--ECCV 2014: 13th European Conference, Zurich, Switzerland, September 6-12, 2014, Proceedings, Part V 13},
  pages={740--755},
  year={2014},
  organization={Springer}
}

@article{krishna2017visual,
  title={Visual genome: Connecting language and vision using crowdsourced dense image annotations},
  author={Krishna, Ranjay and Zhu, Yuke and Groth, Oliver and Johnson, Justin and Hata, Kenji and Kravitz, Joshua and Chen, Stephanie and Kalantidis, Yannis and Li, Li-Jia and Shamma, David A and others},
  journal={International journal of computer vision},
  volume={123},
  pages={32--73},
  year={2017},
  publisher={Springer}
}

@article{xu2023wizardlm,
  title={Wizardlm: Empowering large language models to follow complex instructions},
  author={Xu, Can and Sun, Qingfeng and Zheng, Kai and Geng, Xiubo and Zhao, Pu and Feng, Jiazhan and Tao, Chongyang and Jiang, Daxin},
  journal={arXiv preprint arXiv:2304.12244},
  year={2023}
}

@inproceedings{gupta2019lvis,
  title={Lvis: A dataset for large vocabulary instance segmentation},
  author={Gupta, Agrim and Dollar, Piotr and Girshick, Ross},
  booktitle={Proceedings of the IEEE/CVF conference on computer vision and pattern recognition},
  pages={5356--5364},
  year={2019}
}

@misc{ShareGPT2023,
  title = {ShareGPT},
  year = {2023},
  howpublished={\url{https://sharegpt.com/}},
}

@inproceedings{schwenk2022okvqa,
  title={A-okvqa: A benchmark for visual question answering using world knowledge},
  author={Schwenk, Dustin and Khandelwal, Apoorv and Clark, Christopher and Marino, Kenneth and Mottaghi, Roozbeh},
  booktitle={European Conference on Computer Vision},
  pages={146--162},
  year={2022},
  organization={Springer}
}

@inproceedings{sidorov2020textcaps,
  title={Textcaps: a dataset for image captioning with reading comprehension},
  author={Sidorov, Oleksii and Hu, Ronghang and Rohrbach, Marcus and Singh, Amanpreet},
  booktitle={Computer Vision--ECCV 2020: 16th European Conference, Glasgow, UK, August 23--28, 2020, Proceedings, Part II 16},
  pages={742--758},
  year={2020},
  organization={Springer}
}

@inproceedings{kirillov2023segment,
  title={Segment anything},
  author={Kirillov, Alexander and Mintun, Eric and Ravi, Nikhila and Mao, Hanzi and Rolland, Chloe and Gustafson, Laura and Xiao, Tete and Whitehead, Spencer and Berg, Alexander C and Lo, Wan-Yen and others},
  booktitle={Proceedings of the IEEE/CVF International Conference on Computer Vision},
  pages={4015--4026},
  year={2023}
}

@article{saleh2015large,
  title={Large-scale classification of fine-art paintings: Learning the right metric on the right feature},
  author={Saleh, Babak and Elgammal, Ahmed},
  journal={arXiv preprint arXiv:1505.00855},
  year={2015}
}

@inproceedings{mao2016generation,
  title={Generation and comprehension of unambiguous object descriptions},
  author={Mao, Junhua and Huang, Jonathan and Toshev, Alexander and Camburu, Oana and Yuille, Alan L and Murphy, Kevin},
  booktitle={Proceedings of the IEEE conference on computer vision and pattern recognition},
  pages={11--20},
  year={2016}
}

@inproceedings{marino2019ok,
  title={Ok-vqa: A visual question answering benchmark requiring external knowledge},
  author={Marino, Kenneth and Rastegari, Mohammad and Farhadi, Ali and Mottaghi, Roozbeh},
  booktitle={Proceedings of the IEEE/cvf conference on computer vision and pattern recognition},
  pages={3195--3204},
  year={2019}
}

@inproceedings{lee2019settransformer,
  title={Set transformer: A framework for attention-based permutation-invariant neural networks},
  author={Lee, Juho and Lee, Yoonho and Kim, Jungtaek and Kosiorek, Adam and Choi, Seungjin and Teh, Yee Whye},
  booktitle={International conference on machine learning},
  pages={3744--3753},
  year={2019},
  organization={PMLR}
}

@article{malladi2023mezo,
  title={Fine-tuning language models with just forward passes},
  author={Malladi, Sadhika and Gao, Tianyu and Nichani, Eshaan and Damian, Alex and Lee, Jason D and Chen, Danqi and Arora, Sanjeev},
  journal={Advances in Neural Information Processing Systems},
  volume={36},
  pages={53038--53075},
  year={2023}
}

@article{hinton2015distilling,
  title={Distilling the knowledge in a neural network},
  author={Hinton, Geoffrey and Vinyals, Oriol and Dean, Jeff},
  journal={arXiv preprint arXiv:1503.02531},
  year={2015}
}

@inproceedings{touvron2021training,
  title={Training data-efficient image transformers \& distillation through attention},
  author={Touvron, Hugo and Cord, Matthieu and Douze, Matthijs and Massa, Francisco and Sablayrolles, Alexandre and J{\'e}gou, Herv{\'e}},
  booktitle={International conference on machine learning},
  pages={10347--10357},
  year={2021},
  organization={PMLR}
}

@article{hao2022learning,
  title={Learning efficient vision transformers via fine-grained manifold distillation},
  author={Hao, Zhiwei and Guo, Jianyuan and Jia, Ding and Han, Kai and Tang, Yehui and Zhang, Chao and Hu, Han and Wang, Yunhe},
  journal={Advances in Neural Information Processing Systems},
  volume={35},
  pages={9164--9175},
  year={2022}
}

@article{lo2024m2mkd,
  title={m2mKD: Module-to-Module Knowledge Distillation for Modular Transformers},
  author={Lo, Ka Man and Liang, Yiming and Du, Wenyu and Fan, Yuantao and Wang, Zili and Huang, Wenhao and Ma, Lei and Fu, Jie},
  journal={arXiv preprint arXiv:2402.16918},
  year={2024}
}

@inproceedings{zhang2022minivit,
  title={Minivit: Compressing vision transformers with weight multiplexing},
  author={Zhang, Jinnian and Peng, Houwen and Wu, Kan and Liu, Mengchen and Xiao, Bin and Fu, Jianlong and Yuan, Lu},
  booktitle={Proceedings of the IEEE/CVF Conference on Computer Vision and Pattern Recognition},
  pages={12145--12154},
  year={2022}
}

@inproceedings{wu2022tinyvit,
  title={Tinyvit: Fast pretraining distillation for small vision transformers},
  author={Wu, Kan and Zhang, Jinnian and Peng, Houwen and Liu, Mengchen and Xiao, Bin and Fu, Jianlong and Yuan, Lu},
  booktitle={European Conference on Computer Vision},
  pages={68--85},
  year={2022},
  organization={Springer}
}

@inproceedings{chen2022dearkd,
  title={Dearkd: Data-efficient early knowledge distillation for vision transformers},
  author={Chen, Xianing and Cao, Qiong and Zhong, Yujie and Zhang, Jing and Gao, Shenghua and Tao, Dacheng},
  booktitle={Proceedings of the IEEE/CVF Conference on Computer Vision and Pattern Recognition},
  pages={12052--12062},
  year={2022}
}

@inproceedings{ren2022co,
  title={Co-advise: Cross inductive bias distillation},
  author={Ren, Sucheng and Gao, Zhengqi and Hua, Tianyu and Xue, Zihui and Tian, Yonglong and He, Shengfeng and Zhao, Hang},
  booktitle={Proceedings of the IEEE/CVF Conference on computer vision and pattern recognition},
  pages={16773--16782},
  year={2022}
}

@article{liu2021post,
  title={Post-training quantization for vision transformer},
  author={Liu, Zhenhua and Wang, Yunhe and Han, Kai and Zhang, Wei and Ma, Siwei and Gao, Wen},
  journal={Advances in Neural Information Processing Systems},
  volume={34},
  pages={28092--28103},
  year={2021}
}

@inproceedings{yuan2022ptq4vit,
  title={Ptq4vit: Post-training quantization for vision transformers with twin uniform quantization},
  author={Yuan, Zhihang and Xue, Chenhao and Chen, Yiqi and Wu, Qiang and Sun, Guangyu},
  booktitle={European conference on computer vision},
  pages={191--207},
  year={2022},
  organization={Springer}
}

@inproceedings{ding2022towards,
  title={Towards accurate post-training quantization for vision transformer},
  author={Ding, Yifu and Qin, Haotong and Yan, Qinghua and Chai, Zhenhua and Liu, Junjie and Wei, Xiaolin and Liu, Xianglong},
  booktitle={Proceedings of the 30th ACM International Conference on Multimedia},
  pages={5380--5388},
  year={2022}
}

@inproceedings{liu2023noisyquant,
  title={Noisyquant: Noisy bias-enhanced post-training activation quantization for vision transformers},
  author={Liu, Yijiang and Yang, Huanrui and Dong, Zhen and Keutzer, Kurt and Du, Li and Zhang, Shanghang},
  booktitle={Proceedings of the IEEE/CVF Conference on Computer Vision and Pattern Recognition},
  pages={20321--20330},
  year={2023}
}

@inproceedings{li2022auto,
  title={Auto-vit-acc: An fpga-aware automatic acceleration framework for vision transformer with mixed-scheme quantization},
  author={Li, Zhengang and Sun, Mengshu and Lu, Alec and Ma, Haoyu and Yuan, Geng and Xie, Yanyue and Tang, Hao and Li, Yanyu and Leeser, Miriam and Wang, Zhangyang and others},
  booktitle={2022 32nd International Conference on Field-Programmable Logic and Applications (FPL)},
  pages={109--116},
  year={2022},
  organization={IEEE}
}

@inproceedings{yu2023boost,
  title={Boost vision transformer with gpu-friendly sparsity and quantization},
  author={Yu, Chong and Chen, Tao and Gan, Zhongxue and Fan, Jiayuan},
  booktitle={Proceedings of the IEEE/CVF Conference on Computer Vision and Pattern Recognition},
  pages={22658--22668},
  year={2023}
}

@article{wang2022quantformer,
  title={Quantformer: Learning extremely low-precision vision transformers},
  author={Wang, Ziwei and Wang, Changyuan and Xu, Xiuwei and Zhou, Jie and Lu, Jiwen},
  journal={IEEE Transactions on Pattern Analysis and Machine Intelligence},
  year={2022},
  publisher={IEEE}
}

@article{li2022q,
  title={Q-vit: Accurate and fully quantized low-bit vision transformer},
  author={Li, Yanjing and Xu, Sheng and Zhang, Baochang and Cao, Xianbin and Gao, Peng and Guo, Guodong},
  journal={Advances in neural information processing systems},
  volume={35},
  pages={34451--34463},
  year={2022}
}

@article{xu2022tervit,
  title={TerViT: An efficient ternary vision transformer},
  author={Xu, Sheng and Li, Yanjing and Ma, Teli and Zeng, Bohan and Zhang, Baochang and Gao, Peng and Lv, Jinhu},
  journal={arXiv preprint arXiv:2201.08050},
  year={2022}
}

@inproceedings{lin2023bit,
  title={Bit-shrinking: Limiting instantaneous sharpness for improving post-training quantization},
  author={Lin, Chen and Peng, Bo and Li, Zheyang and Tan, Wenming and Ren, Ye and Xiao, Jun and Pu, Shiliang},
  booktitle={Proceedings of the IEEE/CVF Conference on Computer Vision and Pattern Recognition},
  pages={16196--16205},
  year={2023}
}

@article{dong2024packqvit,
  title={PackQViT: Faster Sub-8-bit Vision Transformers via Full and Packed Quantization on the Mobile},
  author={Dong, Peiyan and Lu, Lei and Wu, Chao and Lyu, Cheng and Yuan, Geng and Tang, Hao and Wang, Yanzhi},
  journal={Advances in Neural Information Processing Systems},
  volume={36},
  year={2024}
}

@inproceedings{he2023bivit,
  title={BiViT: Extremely Compressed Binary Vision Transformers},
  author={He, Yefei and Lou, Zhenyu and Zhang, Luoming and Liu, Jing and Wu, Weijia and Zhou, Hong and Zhuang, Bohan},
  booktitle={Proceedings of the IEEE/CVF International Conference on Computer Vision},
  pages={5651--5663},
  year={2023}
}

@article{xiao2023binaryvit,
  title={BinaryViT: Towards Efficient and Accurate Binary Vision Transformers},
  author={Xiao, Junrui and Li, Zhikai and Yang, Lianwei and Gu, Qingyi},
  journal={arXiv preprint arXiv:2305.14730},
  year={2023}
}

@inproceedings{le2023binaryvit,
  title={BinaryViT: pushing binary vision transformers towards convolutional models},
  author={Le, Phuoc-Hoan Charles and Li, Xinlin},
  booktitle={Proceedings of the IEEE/CVF Conference on Computer Vision and Pattern Recognition},
  pages={4664--4673},
  year={2023}
}

@article{du2024model,
  title={Model Quantization and Hardware Acceleration for Vision Transformers: A Comprehensive Survey},
  author={Du, Dayou and Gong, Gu and Chu, Xiaowen},
  journal={arXiv preprint arXiv:2405.00314},
  year={2024}
}

@article{papa2024survey,
  title={A survey on efficient vision transformers: algorithms, techniques, and performance benchmarking},
  author={Papa, Lorenzo and Russo, Paolo and Amerini, Irene and Zhou, Luping},
  journal={IEEE Transactions on Pattern Analysis and Machine Intelligence},
  year={2024},
  publisher={IEEE}
}

}

\newpage

\end{document}